\documentclass{article}
\usepackage[T1]{fontenc}
\usepackage[utf8]{inputenc}
\usepackage{lmodern}
\usepackage{amsfonts, amssymb, amsmath, amsthm, mathtools}
\usepackage{dsfont}
\usepackage{geometry}
\usepackage{booktabs}
\usepackage[round]{natbib}
\usepackage{enumitem}
\usepackage{listings}
\usepackage{endnotes}
\usepackage{bbm}
\usepackage{comment}
\usepackage{xtab}
\usepackage{array}
\usepackage[colorinlistoftodos]{todonotes}
\usepackage{graphicx}
\usepackage{xcolor}
\usepackage{caption}
\usepackage{hyperref}
\usepackage{cleveref}
\usepackage{algorithm}
\usepackage{algpseudocode}
\usepackage{multirow}

\graphicspath{{./}{./Figures/}{./Figures/ver3/}{./Biometrika2025/art/}}
\newcolumntype{C}{>{\centering\arraybackslash}p{1.8cm}}

\hypersetup{
colorlinks=true,
linktoc=all,
linkcolor=blue,
urlcolor=blue,
citecolor=blue,
filecolor=blue,
linktocpage=true
}

\newtheorem{definition}{Definition}
\newtheorem{theorem}{Theorem}
\newtheorem{lemma}{Lemma}
\newtheorem{corollary}{Corollary}
\newtheorem{example}{Example}
\newtheorem{assumption}{Assumption}

\makeatother
\let\footnote=\endnote

\newcommand{\mP}{\mathbb{P}}
\newcommand{\mE}{\mathbb{E}}
\newcommand{\mD}{\mathbb{D}}
\newcommand{\mV}{\mathrm{Var}}

\newcommand{\given}{\,|\,}
\newcommand{\sgiven}{\!\,|\,\!}
\newcommand{\I}{\mathbb{(I)}}
\newcommand{\II}{\mathbb{(II)}}

\newcommand{\supple}{the Appendix}
\newcommand{\main}{the main text}
\newcommand{\independent}{\mbox{${}\perp\mkern-11mu\perp{}$}}
\newcommand{\nindep}{\mathrel{\not\!\perp\mkern-11mu\perp}}
\newcommand{\iid}{ \overset{\mathrm{i.i.d.}}{\sim}}
\DeclareMathOperator*{\argmin}{arg\,min}

\definecolor{DarkBlue}{rgb}{0,0,0.55}

\newcommand{\mypara}[1]{\vspace{0.5em}\noindent\textbf{#1.}}

\begin{document}

\title{General Frameworks for Conditional Two-Sample Testing}

\author{
Seongchan Lee\textsuperscript{1}\thanks{These authors contributed equally to this work.} \hspace{1.5cm} 
Suman Cha\textsuperscript{2}\footnotemark[1] \hspace{1.5cm} 
Ilmun Kim\textsuperscript{1}
}

\maketitle
\footnotetext[1]{\textsuperscript{}Department of Mathematical Sciences, KAIST, Daejeon, South Korea.}
\footnotetext[2]{\textsuperscript{}Department of Statistics and Data Science, Yonsei University, Seoul, South Korea.}

\begin{abstract}
We study the problem of conditional two-sample testing, which aims to determine whether two populations have the same distribution after accounting for confounding factors. This problem commonly arises in various applications, such as domain adaptation and algorithmic fairness, where comparing two groups is essential while controlling for confounding variables. We begin by establishing a hardness result for conditional two-sample testing, demonstrating that no valid test can have significant power against any single alternative without proper assumptions. We then introduce two general frameworks that implicitly or explicitly target specific classes of distributions for their validity and power. Our first framework allows us to convert any conditional independence test into a conditional two-sample test in a black-box manner, while preserving the asymptotic properties of the original conditional independence test. The second framework transforms the problem into comparing marginal distributions with estimated density ratios, which enables direct application of existing methods for marginal two-sample testing. We demonstrate this idea in a concrete manner with classification and kernel-based methods. Finally, simulation studies are conducted to illustrate the proposed frameworks in finite-sample scenarios.
\end{abstract}

\tableofcontents

\section{Introduction}\label{Section : Introduction}
\subsection{Problem Setup}\label{Subsection:Problem_Setup}
This paper addresses the problem of conditional two-sample testing, which aims to test the equality of two conditional distributions. Statistical methods for this problem have important applications across diverse fields such as domain adaptation and algorithmic fairness. In domain adaptation, for instance, this methodology can serve as a formal framework to validate the covariate shift assumption. Under this assumption, the conditional distribution of $Y$ given $X$ remains unchanged, while the marginal distribution of $X$ may differ. By confirming this assumption, practitioners can effectively re-weight the training data according to the marginal density ratio with respect to $X$, which potentially leads to improved predictive performance and better adaptation to new domains~\citep{shimodaira2000improving,JMLR:v8:sugiyama07a,sugiyama2007direct}.
Moreover, in algorithmic fairness, conditional two-sample testing provides a principled tool for detecting and mitigating biases. Specifically, it helps identify whether a particular machine learning model unfairly favors or disadvantages specific groups based on demographic characteristics such as age, gender, or ethnicity~\citep{NIPS2016_9d268236,barocas2019fairness}. Conditional two-sample testing also finds applications beyond machine learning. In genomics, for example, scientists seek to identify differences in genetic distributions conditional on various factors such as disease status and environmental exposures~\citep{virolainen2022,wu2023}. This methodology aids scientists in understanding the genetic basis of diseases and in developing strategies for personalized medicine by providing a rigorous framework for comparing conditional distributions. 

With the practical motivation in mind, we now formally set up the problem. Given $n_1,n_2 \in \mathbb{N}$, suppose we observe two mutually independent samples 
\begin{align*}
\{(X_i^{(1)},Y_i^{(1)})\}_{i=1}^{n_1} \iid P_{XY}^{(1)} \quad \text{and} \quad \{(X_i^{(2)},Y_i^{(2)})\}_{i=1}^{n_2} \iid P_{XY}^{(2)}, 
\end{align*}
where $P_{XY}^{(1)}$ and $P_{XY}^{(2)}$ are joint distributions supported on some generic product space $\mathcal{X} \times \mathcal{Y}$. Let $P_{Y \sgiven X}^{(1)}$ and $P_{Y \sgiven X}^{(2)}$ denote the conditional distributions of $Y^{(1)} \given X^{(1)}$ and $Y^{(2)} \given X^{(2)}$, respectively. Similarly, let $P_{X}^{(1)}$ and $P_{X}^{(2)}$ denote the marginal distributions of $X^{(1)}$ and $X^{(2)}$, respectively. Given these two samples, our goal is to test the equality of two conditional distributions
\begin{align} \label{Eq: hypothesis}
H_0: P_{X}^{(1)} \bigl\{ P_{Y \sgiven X}^{(1)}(\cdot \given X) = P_{Y \sgiven X}^{(2)}(\cdot \given X)   \bigr\} = 1 \;\, \text{vs} \;\, H_1: P_{X}^{(1)} \bigl\{ P_{Y \sgiven X}^{(1)}(\cdot \given X) \neq  P_{Y \sgiven X}^{(2)}(\cdot \given X) \bigr\} >0, 
\end{align}
where $P^{(j)}_{Y\sgiven X}(\cdot\given x)$ denotes the conditional distribution of $Y^{(j)}$ given $X^{(j)}=x$ for $j \in \{1,2\}$. In other words, we are interested in determining whether two populations have the same distribution after controlling for potential confounding variables. Throughout this paper, we assume that $P_{X}^{(1)}$ and $P_{X}^{(2)}$ have the same support, satisfying $P_{X}^{(1)} \ll P_{X}^{(2)}$ and $P_{X}^{(2)} \ll P_{X}^{(1)}$ where the symbol $\ll$ denotes absolute continuity. Since $P_{X}^{(1)}$ and $P_{X}^{(2)}$ have the same support, the above hypotheses~\eqref{Eq: hypothesis} for conditional two-sample testing can be equivalently defined using $P_{X}^{(2)}$ instead of $P_{X}^{(1)}$. 

As pointed out by \citet{boeken2021bayesian}, conditional two-sample testing is closely connected to conditional independence testing. To illustrate this connection, we introduce a binary variable $Z \in \{1,2\}$, and see that the conditional independence between $Y$ and $Z$ given $X$ is equivalently expressed as
\begin{align} \label{Eq: CIT = C2ST}
Y \independent Z \given X ~ \Longleftrightarrow ~ (Y \given X, Z = 1) \overset{d}{=} (Y \given X, Z = 2), 
\end{align}
where the symbol $\overset{d}{=}$ denotes equality in distribution. This equivalence enables us to convert the problem of conditional two-sample testing to that of conditional independence testing based on the datasets 
$\{(X_i,Y_i) : Z_i = 1\}$ and $\{(X_i,Y_i) : Z_i = 2\}$.
Consequently, we can leverage various existing methods for conditional independence testing to tackle conditional two-sample testing. However, the relationship between conditional two-sample testing and conditional independence testing has not been rigorously established in prior work.
A key challenge arises from the deterministic nature of the group indicator $Z$ in the conditional two-sample problem.
Specifically, letting $n = n_1 + n_2$, the quantities $\sum_{i=1}^n \mathds{1}(Z_i = 1)$ and $\sum_{i=1}^n \mathds{1}(Z_i = 2)$ are fixed in advance, corresponding to the sample sizes $n_1$ and $n_2$.
This distinction prevents a direct application of existing conditional independence results and leaves a gap between the two problems. To navigate this gap while preserving readability, we primarily formulate our results using fixed distributions. However, our frameworks naturally accommodate $n$-dependent sequences, which we explicitly formalize only when theoretically necessary, as in \Cref{Section: Approach via Conditional Independence Testing}.

\subsection{An Overview of Our Results}
In this work, we make several contributions to the field of conditional two-sample testing. First, we establish that comparing conditional distributions is intrinsically more difficult than comparing marginal distributions. For marginal two-sample testing, one can design permutation tests that control type~I error, while being powerful against certain alternatives~\citep[e.g.,][]{kim2022minimax}. However, we show that this is not the case for conditional two-sample testing. \Cref{Theorem: negative result} proves that any valid conditional two-sample test has power at most equal to its size against any single alternative if the conditioning random vector is of continuous type. This is reminiscent of the negative result for conditional independence testing proved by \citet{shah2020hardness}. It is worth highlighting, however, that their negative result does not directly imply \Cref{Theorem: negative result}. The proof of \citet{shah2020hardness} relies on the assumption that the data $\{(X_i,Y_i,Z_i)\}_{i=1}^n$ are i.i.d., which does not hold in our setup as $\sum_{i=1}^n \mathds{1}(Z_i = 1)$ and $\sum_{i=1}^n \mathds{1}(Z_i = 2)$ are deterministic. We handle this distinction through a concentration argument and show that conditional two-sample testing is as difficult as conditional independence testing. This negative result naturally motivates additional assumptions that make the problem feasible.

Our next contribution is to introduce two general frameworks for conditional two-sample testing. The first framework effectively addresses the challenge arising from the deterministic nature of the group indicator in conditional two-sample testing. In particular, we develop a generic method that converts any conditional independence test into a conditional two-sample test. This general method directly transfers the asymptotic properties of a conditional independence test computed using $\{(X_i,Y_i,Z_i)\}_{i=1}^n \iid P_{XYZ}$ to the setting of conditional two-sample testing, as we show in \Cref{Theorem: Converting C2ST into CIT}. At the heart of this approach is the concentration property of a Binomial random variable around its mean, which facilitates the effective construction of i.i.d.~samples drawn from $P_{XYZ}$ (see Algorithm~\ref{Algorithm: Converting CIT into C2ST}). This development paves the way for leveraging any existing methods for conditional independence testing in the literature, thereby expanding the range of tools available to practitioners for conducting two-sample tests. 

The second framework is based on density ratio estimation. To elaborate, suppose that $P_X^{(1)}$ and $P_X^{(2)}$ admit densities $f_X^{(1)}$ and $f_X^{(2)}$, and that the conditional distributions $P_{Y\sgiven X}^{(1)}(\cdot\given x)$ and $P_{Y\sgiven X}^{(2)}(\cdot\given x)$ admit densities $f_{Y\sgiven X}^{(1)}(\cdot\given x)$ and $f_{Y\sgiven X}^{(2)}(\cdot\given x)$, all with respect to a common base measure. Then for all $(x,y) \in \mathcal{X} \times \mathcal{Y}$, we have the identity:
\begin{align} \label{Eq: C2ST = Marginal 2ST}
f_{Y \sgiven X}^{(1)}(y \given x) = f_{Y \sgiven X}^{(2)} (y \given x) ~\Longleftrightarrow~ f_{XY}^{(1)}(x,y) = \frac{f_{X}^{(1)}(x)}{f_{X}^{(2)}(x)} f_{XY}^{(2)}(x,y):= f_{XY}(x,y),
\end{align} 
where $f_{XY}^{(1)}$ is the joint density function of $(X^{(1)},Y^{(1)})$ such that $f_{XY}^{(1)}(x,y) = f_{Y \sgiven X}^{(1)}(y \given x) f_{X}^{(1)}(x)$, and $f_{XY}^{(2)}$ is similarly defined for $(X^{(2)},Y^{(2)})$. The above equivalence~\eqref{Eq: C2ST = Marginal 2ST} allows us to transform the problem of testing for conditional distributions into the one that compares marginal distributions with densities $f_{XY}^{(1)}$ and $f_{XY}$. The latter problem has been extensively studied with various methods, ranging from classical approaches such as Hotelling's test to modern methods such as kernel maximum mean discrepancy~\citep{gretton2012kernel,liu2020learning,schrab2023mmd} and machine learning-based approaches~\citep[e.g.,][]{lopez2017revisiting,kim2019,kim2021classification,hediger2022use}. The issue, however, is that we do not observe samples from $f_{XY}$ but from $f_{XY}^{(2)}$. Therefore, the success of this framework relies on how accurately one can estimate the density ratio $r_X(x):=  {f_{X}^{(1)}(x)}/{f_{X}^{(2)}(x)},$ and incorporate it into the procedure to fill the gap between $f_{XY}$ and $f_{XY}^{(2)}$. We demonstrate this methodology by focusing on a classification-based test in \Cref{Section: Classifier-based Approach} and a kernel-based test in \Cref{Section: Kernel MMD}. The theoretical results are proved in \Cref{section: Proofs}.

\subsection{Literature Review}
As mentioned earlier, conditional two-sample testing has critical applications in fields such as machine learning and genomics, where comparing groups while controlling for confounders is essential. Despite its significance, research on nonparametric comparisons of entire conditional distributions remains limited. While existing work has investigated testing for the equality of conditional moments and goodness-of-fit for pre-specified distributions, these methods target specific aspects rather than the distribution as a whole. Several recent works have begun to fill this gap by developing nonparametric methods that directly compare conditional distributions. 
\citet{hu2024two} introduced a weighted rank-sum statistic relying on both marginal and conditional density ratios, a special case of our density-ratio-based framework that we derive in Appendix~\ref{sec:importance_weighted_statistics}. \citet{chen2025biased} later extended this idea using Neyman orthogonality to reduce first-order bias. Rather than estimating density ratios, \citet{yan2025distancekernelbasedmeasuresglobal} proposed a kernel-based framework built on the conditional energy distance and conditional maximum mean discrepancy, where test statistics are constructed from conditional U-statistics with kernel smoothing and calibrated via a local bootstrap procedure. In a different direction, \citet{CHATTEJEE2024conditional} considered a paired-covariate setting where both response variables are conditioned on the same covariates, which differs from our independent-covariate setting and makes their methods not directly comparable to ours.

Our first framework builds on the connection between conditional two-sample testing and conditional independence testing described in~\eqref{Eq: CIT = C2ST}, and can be combined with a wide range of existing methods. Notable examples include the Generalized Covariance Measure \citep{shah2020hardness} and its projected variant \citep{lundborg2024projected}, as well as kernel-based approximations \citep{strobl2019approximate}, binning-based tests \citep{neykov2021minimax}, and general regression frameworks \citep{williamson2023general}.

On the other hand, our second framework relies on density ratio estimation, where directly estimating the ratio is known to be more stable than separate density estimation in high dimensions \citep{sugiyama2007direct}. A closely related contribution is that of \citet{bordino2026nonparametric}, who proposed a permutation test for the hypothesis that the ratio of two densities is proportional to a known function. Their approach connects to our framework through the reformulation in~\eqref{Eq: C2ST = Marginal 2ST}. A more comprehensive review, including classical approaches to conditional moments and goodness-of-fit, the broader conditional independence testing literature, and density ratio estimation techniques, is provided in Appendix~\ref{sec:extended_literature_technical}.

\section{Hardness Result} \label{Section: Impossibility Result of Conditional Two-Sample Testing}
Before introducing our frameworks, we present a fundamental hardness result for conditional two-sample testing. Specifically, for a continuous random vector $X$, our result demonstrates that any valid conditional two-sample test has no power against any alternative. This finding parallels the negative result established by \citet{shah2020hardness} for conditional independence testing, and our proof relies crucially on their work. Given the connection established in \eqref{Eq: CIT = C2ST}, one might argue that their negative result directly applies to the two-sample problem. However, formalizing this connection requires additional care since the sample sizes $n_1$ and $n_2$ are deterministic in our setting, which violates the i.i.d.~assumption required in \citet{shah2020hardness}.

{While the problem is formulated on a generic product space in \Cref{Section : Introduction}, we restrict our focus to Euclidean spaces to establish the hardness result. Let $\mathcal{P}$ denote the set of all pairs of distributions $P := (P_{XY}^{(1)}, P_{XY}^{(2)})$ defined on $\mathbb{R}^{d_X} \times \mathbb{R}^{d_Y}$. We write $\mathcal{P}_0 \subset \mathcal{P}$ for the subset of pairs $P$ satisfying the null hypothesis $H_0$ in \eqref{Eq: hypothesis}, and $\mathcal{P}_1 = \mathcal{P} \setminus \mathcal{P}_0$ for the corresponding alternative class. 
Let $\mathcal{P}^{\mathsf{ac}} \subseteq \mathcal{P}$ denote the subset of pairs where each joint distribution $P_{XY}^{(j)}$ ($j \in \{1,2\}$) is absolutely continuous with respect to the Lebesgue measure on $\mathbb{R}^{d_X + d_Y}$. Furthermore, for $M \in (0, \infty]$, let $\mathcal{P}_M \subseteq \mathcal{P}$ be the subset of pairs $P$ such that the supports of both $P_{XY}^{(1)}$ and $P_{XY}^{(2)}$ are contained within an $\ell_\infty$ ball of radius $M$ centered at the origin. We then define the restricted null and alternative classes for the hardness result as $\mathcal{P}_{0,M}^{\mathsf{ac}} := \mathcal{P}_0 \cap \mathcal{P}^{\mathsf{ac}} \cap \mathcal{P}_M$ and $\mathcal{P}_{1,M}^{\mathsf{ac}} := \mathcal{P}_1 \cap \mathcal{P}^{\mathsf{ac}} \cap \mathcal{P}_M$.}

\begin{theorem} \label{Theorem: negative result}
Let $n_1, n_2 \in \mathbb{N}$, $\alpha \in (0,1)$, and $M \in (0,\infty]$. Suppose we observe two independent samples $\{(X_i^{(j)}, Y_i^{(j)})\}_{i=1}^{n_j} \iid P_{XY}^{(j)}$ for $j=1,2$, where the distribution pair $P = (P_{XY}^{(1)}, P_{XY}^{(2)}) \in \mathcal{P}^{\mathsf{ac}}_M$. Let $\phi$ be any (possibly randomized) test based on these two samples, taking values in $\{0,1\}$. If $\phi$ controls the type~I error at level $\alpha$, i.e., 
\begin{align*}
    \sup_{P \in \mathcal{P}^{\mathsf{ac}}_{0,M}} \mE_P(\phi) \leq \alpha,
\end{align*}
then the power of $\phi$ cannot exceed $\alpha$ for any alternative $P \in \mathcal{P}^{\mathsf{ac}}_{1,M}$, that is, 
\begin{align*}
    \sup_{P \in \mathcal{P}^{\mathsf{ac}}_{1,M}} \mE_P(\phi) \leq \alpha.
\end{align*}
\end{theorem}

\Cref{Theorem: negative result} shows that it is necessary to impose additional assumptions (e.g., smoothness conditions on the distributions) to make the conditional two-sample problem feasible. In the next two sections, we introduce two general frameworks, which implicitly or explicitly incorporate reasonable assumptions to address this problem. The first framework converts any conditional independence test into a conditional two-sample test and is applicable whenever the chosen test is valid and consistent. In contrast, the second framework assumes that the marginal density ratio $r_X$ is well-behaved and can be estimated with high accuracy.

\section{Approach via Conditional Independence Testing} \label{Section: Approach via Conditional Independence Testing}

\subsection{Algorithm and Asymptotic Guarantees}

In this section, we introduce our first framework that converts a conditional independence test into a conditional two-sample test, while maintaining the same asymptotic guarantees. The key idea is to construct a dataset $\mathcal{D}_{\tilde{n}}$ of size $\tilde{n}$, consisting of i.i.d.~random vectors $(X,Y,Z)$, based on the given two samples $\{(X_i^{(1)}, Y_i^{(1)})\}_{i=1}^{n_1}$ and $\{(X_i^{(2)}, Y_i^{(2)})\}_{i=1}^{n_2}$. To achieve this, letting $n = n_1 + n_2$, we first draw a random variable $\tilde{n}_1$ from $\mathrm{Binomial}$($\tilde{n}$, $n_1/n$) where $\tilde{n}$ is set to be smaller than $n$ and $\tilde{n}/n \rightarrow 1$. 
A similar idea was employed by \citet{neykov2021minimax} in a different context to eliminate Poissonization for conditional independence testing.
Since a Binomial random variable is highly concentrated around its mean, we can guarantee that $\tilde{n}_1 \leq n_1$ and $\tilde{n}_2 :=  \tilde{n} - \tilde{n}_1 \leq n_2$ with high probability. In the event that $\tilde{n}_1 > n_1$ or $\tilde{n}_2 > n_2$, making the construction of $\mathcal{D}_{\tilde{n}}$ infeasible, we simply accept the null hypothesis. 
This slightly inflates the type II error in finite-sample scenarios, but it is asymptotically negligible.

Having constructed $\mathcal{D}_{\tilde{n}}$ consisting of~i.i.d.\ random samples drawn from the joint distribution of $(X,Y,Z)$, we can now implement a conditional independence test, as summarized in Algorithm~\ref{Algorithm: Converting CIT into C2ST}, while retaining the theoretical guarantees for conditional two-sample testing. 

\begin{algorithm}[t!]
\caption{Converting a Conditional Independence Test into a Conditional Two-Sample Test}
\label{Algorithm: Converting CIT into C2ST}
\begin{algorithmic}[1]
    \Require Data $\{(X_i^{(1)}, Y_i^{(1)})\}_{i=1}^{n_1}$ and $\{(X_i^{(2)}, Y_i^{(2)})\}_{i=1}^{n_2}$, a conditional independence test $\psi$ for $H_0: Y \independent Z \mid X$ of (asymptotic) size $\alpha \in (0,1)$, and an adjustment parameter $\varepsilon \in (0,1)$. \vskip .5em
    \State Let $n=n_1+n_2$. Draw $\tilde{n}_1 \sim \mathrm{Binomial}(\tilde{n}, n_1/n)$ where $\tilde{n} = \lfloor k^\ast n \rfloor$ and
    \[
    k^\ast = 1 - \frac{3\log(\varepsilon)}{2n_{\min}}
    - \sqrt{\left(1 - \frac{3\log(\varepsilon)}{2n_{\min}}\right)^2 - 1},
    \quad \text{where } n_{\min} = \min(n_1, n_2).
    \]
    Set $\tilde{n}_2 = \tilde{n} - \tilde{n}_1$. \vskip .2em
    \If{$\tilde{n}_1 > n_1$ or $\tilde{n}_2 > n_2$}
    \State Accept $H_0$.
    \Else
    \State Merge $\{(X_i^{(1)}, Y_i^{(1)}, Z_i=1)\}_{i=1}^{\tilde{n}_1}$ and $\{(X_i^{(2)}, Y_i^{(2)}, Z_i=2)\}_{i=1}^{\tilde{n}_2}$, yielding $\mathcal{D}_{\tilde{n}} \coloneqq \{(X_i, Y_i, Z_i)\}_{i=1}^{\tilde{n}}$.
    \State Apply a uniform random permutation to $\mathcal D_{\tilde n}$.
    \State Run a conditional independence test $\psi$ on $\mathcal{D}_{\tilde{n}}$ at level $\alpha$, and denote the resulting test as $\psi_{\tilde{n}}$.
    \If{$\psi_{\tilde{n}} = 1$}
    \State Reject $H_0$.
    \Else
    \State Accept $H_0$.
    \EndIf
    \EndIf
\end{algorithmic}
\end{algorithm}    	   	

To establish the asymptotic guarantees of \Cref{Algorithm: Converting CIT into C2ST}, we connect the two-sample and independence testing settings by defining a sequence of joint distributions $P_{XYZ,n}$ for a given pair $P = (P_{XY}^{(1)}, P_{XY}^{(2)}) \in \mathcal{P}$:
\begin{align}\label{eq : definition of P_XYZ}
P_{XYZ,n} = \lambda_n \big(P_{XY}^{(1)} \otimes \delta_{\{Z=1\}}\big) + (1-\lambda_n) \big(P_{XY}^{(2)} \otimes \delta_{\{Z=2\}}\big),
\end{align}
where $\lambda_n = n_1/n$ and $\delta_{\{Z=j\}}$ is the Dirac measure at $Z=j$. As noted in \Cref{Subsection:Problem_Setup}, this triangular array rigorously accommodates the deterministic sample sizes $n_1$ and $n_2$. To prevent either group from vanishing asymptotically, we assume $\lambda_n \in [c, 1-c]$ for some constant $c \in (0, 1/2)$ and all sufficiently large $n$. Hereafter, we write $P_{XYZ}$ for simplicity, keeping its implicit dependence on $n$ in mind. Let $\mathcal{P}_0^{\mathsf{cit}} \subseteq \mathcal{P}_0$ and $\mathcal{P}_1^{\mathsf{cit}} \subseteq \mathcal{P}_1$ be the subclasses of null and alternative distributions where a given conditional independence test $\psi$ is asymptotically valid and consistent under $P_{XYZ}$. The following theorem shows our procedure preserves both properties, with non-asymptotic bounds deferred to \Cref{Section: Proof of Converting C2ST into CIT}.

\begin{theorem} \label{Theorem: Converting C2ST into CIT}
{Let $P = (P_{XY}^{(1)}, P_{XY}^{(2)}) \in \mathcal{P}$ and let $P_{XYZ}$ be the joint distribution induced by $P$ as in \eqref{eq : definition of P_XYZ}. Assume there exists a constant $c \in (0, 1/2)$ such that $c \leq n_1/n \leq 1-c$ for all sufficiently large $n$, which implies $n_{\min} = \min(n_1, n_2) \to \infty$. Suppose $\psi$ is a conditional independence test of nominal level $\alpha \in (0,1)$ for i.i.d.\ data from $P_{XYZ}$, satisfying
    \begin{align*}
        \limsup_{n \to \infty} \sup_{P \in \mathcal{P}_0^{\mathsf{cit}}} \mE_P(\psi) \leq \alpha \quad \text{and} \quad \lim_{n \to \infty} \sup_{P \in \mathcal{P}_1^{\mathsf{cit}}} \mE_P(1-\psi) = 0.
    \end{align*}
    Let $\phi$ denote the output of Algorithm~\ref{Algorithm: Converting CIT into C2ST} with $\psi$ and $\varepsilon = o(1)$ satisfying $\lim_{n \to \infty} \frac{\log(1/\varepsilon)}{n_{\min}} = 0$, applied to independent samples $\{(X_i^{(j)}, Y_i^{(j)})\}_{i=1}^{n_j} \iid P_{XY}^{(j)}\,$ for $j=1,2$. Then
    \begin{align*}
        \limsup_{n \to \infty} \sup_{P \in \mathcal{P}_0^{\mathsf{cit}}} \mE_P(\phi) \leq \alpha \quad \text{and} \quad \lim_{n \to \infty} \sup_{P \in \mathcal{P}_1^{\mathsf{cit}}} \mE_P(1-\phi) = 0.
\end{align*}}
\end{theorem}

\vspace{-0.5em}
In \Cref{Theorem: Converting C2ST into CIT}, $o(1)$ denotes a term vanishing as $n \to \infty$. The operator $\mE_P(\cdot)$ denotes the expectation over all randomness induced by $P$. This expectation is taken over the $P_{XYZ}$ samples for $\psi$, and encompasses both original data generation and algorithmic subsampling for $\phi$.

When $X$ is degenerate (e.g., $X = 0$ with probability one), the conditional two-sample problem reduces to its unconditional counterpart, and Algorithm~\ref{Algorithm: Converting CIT into C2ST} provides a generic method to convert unconditional independence tests into unconditional two-sample tests. Additionally, the specific form of $k^\ast$ in Algorithm~\ref{Algorithm: Converting CIT into C2ST} is derived from the multiplicative Chernoff bound for a Binomial random variable, which can be refined by numerically computing the exact Binomial tail probability, as detailed in Appendix~\ref{Appendix: refined k star}.

\subsection{Coupling Stability}

Despite its generality, one drawback of Algorithm~\ref{Algorithm: Converting CIT into C2ST} is that it discards the observations $\{(X_i^{(1)}, Y_i^{(1)})\}_{i=\tilde{n}_1+1}^{n_1}$ and $\{(X_i^{(2)}, Y_i^{(2)})\}_{i=\tilde{n}_2+1}^{n_2}$ whenever $\tilde{n}_1 < n_1$ and $\tilde{n}_2 < n_2$. Provided that $n_1$ and $n_2$ grow proportionally with $n$, the number of discarded samples $n - \tilde{n}$ is $O\bigl(\sqrt{n\log(1/\varepsilon)}\bigr)$. While this loss may degrade performance in small-sample regimes, it becomes negligible when $n$ is large and $\varepsilon$ decreases slowly, as supported empirically in Appendix~\ref{Appendix: With_Without_Algorithm_1}. Nevertheless, for a restricted class of test statistics that are \emph{coupling-stable}, defined in \Cref{def:stability}, the conclusion of \Cref{Theorem: Converting C2ST into CIT} may continue to hold without discarding samples.

To formalize this idea, we introduce a coupling construction adapted from \citet{chung2013}. Let $\bar{n}_1 \sim \mathrm{Binomial}(n,n_1/n)$ and $\bar{n}_2 = n - \bar{n}_1$. If $\bar{n}_1 > n_1$, we draw $\bar{n}_1 - n_1$ additional samples 
$\{(X_i^{(1)}, Y_i^{(1)})\}_{i=n_1+1}^{\bar{n}_1}$ from $P_{XY}^{(1)}$. Otherwise, we draw $\bar{n}_2 - n_2$ additional samples $\{(X_i^{(2)}, Y_i^{(2)})\}_{i=n_2+1}^{\bar{n}_2}$ from $P_{XY}^{(2)}$. In either case, we define the coupled dataset 
$\{(\tilde{X}_i,\tilde{Y}_i,\tilde{Z}_i)\}_{i=1}^n = \{(X_i^{(1)},Y_i^{(1)}, 1)\}_{i=1}^{\bar{n}_1} \cup \{(X_i^{(2)},Y_i^{(2)}, 2)\}_{i=1}^{\bar{n}_2}$, which, after randomly permuting indices, can be viewed as i.i.d.~draws from $P_{XYZ}$.
When this newly constructed dataset is compared with the original dataset $\{(X_i,Y_i,Z_i)\}_{i=1}^n = \{(X_i^{(1)},Y_i^{(1)}, 1)\}_{i=1}^{n_1} \cup \{(X_i^{(2)},Y_i^{(2)}, 2)\}_{i=1}^{n_2}$, there are $|\bar{n}_1 - n_1|$ distinct data points satisfying $\mE[|\bar{n}_1 - n_1|] \leq \sqrt{(n/4)}$. This motivates the following definition.
\begin{definition}[Coupling stability]\label{def:stability}
    Let $T_n$ be a test statistic computed on the original dataset $\{(X_i,Y_i,Z_i)\}_{i=1}^n$, and let $\tilde T_n$ denote its analogue computed on the coupled dataset $\{(\tilde X_i,\tilde Y_i,\tilde Z_i)\}_{i=1}^n$ constructed above. We say that $T_n$ is \emph{asymptotically coupling-stable} if $T_n-\tilde T_n$ converges in probability to zero as $n \to \infty$.
\end{definition}

{This definition captures the idea that, when a test statistic $T_n$ is asymptotically insensitive to $\sqrt{n}$-scale perturbations of the sample, its asymptotic behavior is preserved across both the original and the coupled datasets. Whether a given statistic is coupling-stable, however, depends on its specific form, as the following examples illustrate.}

\begin{example}[Stable case] \label{Example : Stable case}\normalfont
To simplify our presentation, consider the case $Y \in \mathbb{R}$, and assume $f(x) :=  \mE(Y \given X = x)$ and $g(x) :=  \mE(Z \given X =x)$ are known. Letting $R_i :=  \{Y_i - f(X_i)\} \{Z_i - g(X_i)\}$, the generalized covariance measure introduced by \citet{shah2020hardness} is 
\begin{align*}
    T_n = \frac{n^{-1/2}\sum_{i=1}^n R_i}{\sqrt{\frac{1}{n} \sum_{i=1}^n R_i^2 - \bigl(\frac{1}{n} \sum_{r=1}^n R_r\bigr)^2}},
\end{align*}
and let $\tilde{T}_n$ be the analogue of $T_n$ computed on $\{(\tilde{X}_i,\tilde{Y}_i,\tilde{Z}_i)\}_{i=1}^n$. Focusing on the numerators of $T_n$ and $\tilde{T}_n$, we see that their difference is 
\begin{align*}
    \frac{1}{\sqrt{n}}\sum_{i=1}^n (R_i - \tilde{R}_i) =
    \frac{1}{\sqrt{n}}\sum_{i= n_1 + 1}^{\bar{n}_1} (R_i - \tilde{R}_i) \cdot \mathds{1}(\bar{n}_1 > n_1) + \frac{1}{\sqrt{n}}\sum_{i= \bar{n}_1 + 1}^{n_1} (R_i - \tilde{R}_i) \cdot \mathds{1}(\bar{n}_1 \leq n_1).
\end{align*}
Under the null hypothesis, the expectation of the difference is zero and the variance is bounded above by $O(n^{-1/2})$, provided $\mE(Y_i^2) < \infty$. Therefore, the difference of the numerators is asymptotically negligible. A similar argument shows that the difference of the denominators is also asymptotically negligible as detailed in Appendix~\ref{Appendix : Proof of Example : Unstable case}. Putting things together, we conclude that $T_{n}$ is coupling-stable in the sense of \Cref{def:stability}.
\end{example}

In practice, $f$ and $g$ are estimated from the data, which can introduce coupling-instability when these estimators are sensitive to sample perturbations. However, as the next example shows, coupling-instability is not solely an artifact of estimation error; it can also arise intrinsically from the structure of the test statistic itself.
{\begin{example}[Unstable case] 
    \label{Example : Unstable case}\normalfont
    In the same univariate setting as \Cref{Example : Stable case}, suppose $f(x) = \mE(Y \mid X=x)$ is difficult to estimate nonparametrically, whereas the labeling mechanism $g(x) = \mE(Z \mid X=x)$ is known. One might then consider the statistic $T_n = n^{-1/2} \sum_{i=1}^n \zeta_i$, where $\zeta_i := Y_i\{Z_i - g(X_i)\}$, which does not require knowledge of $f$. Let $\tilde{T}_n$ denote the same statistic computed on the coupled sample. Under the null hypothesis $Y \independent Z \mid X$, both $T_n$ and $\tilde{T}_n$ converge to centered normal limits with different variances, provided that $\mE\{f(X)g(X)(1 - g(X))\} \neq 0$ (see Appendix~\ref{Appendix : Proof of Example : Unstable case}). Because their asymptotic distributions differ under this condition, $T_n$ is not coupling-stable in the sense of \Cref{def:stability}. Consequently, bypassing Algorithm~\ref{Algorithm: Converting CIT into C2ST} and applying such a test directly to the original data may fail to control the type~I error, underscoring the necessity of coupling stability.
\end{example}}

The previous examples highlight the need for caution when applying a conditional independence test directly to the two-sample data and justify our generic conversion approach in Algorithm~\ref{Algorithm: Converting CIT into C2ST}. The next section introduces an alternative framework for conditional two-sample testing based on density ratio estimation.

\section{Approach via Density Ratio Estimation} 
\label{Section: Approach via Density Ratio Estimation}

\subsection{General Principle and Examples}
\label{Subsection : General principle and examples}

In this subsection, we present our second framework, which transforms conditional two-sample testing into a marginal comparison via density ratio estimation. Recall from \eqref{Eq: C2ST = Marginal 2ST} that the null hypothesis holds if and only if $f_{XY}^{(1)} = f_{XY} := r_X \cdot f_{XY}^{(2)}$, where $r_X(x) = {f_{X}^{(1)}(x)}/{f_{X}^{(2)}(x)}$. Since we only observe samples from $f_{XY}^{(2)}$ rather than $f_{XY}$, we correct this sampling bias using importance weighting~\citep[][for a survey]{kimura2024short}. Specifically, estimating an expectation under a target density $p$ using samples $X_1,\ldots,X_n$ from a source density $q$ is achieved by reweighting each observation by $p(X_i)/q(X_i)$, since $\mE_Q\bigl[\frac{p(X)}{q(X)}\varphi(X)\bigr] = \mE_P[\varphi(X)]$.

Assuming $r_X$ is known (or accurately estimated), this principle naturally extends to marginal two-sample testing. Let $V^{(j)} := (X^{(j)},Y^{(j)})$ for $j \in \{1, 2\}$. For basic test statistics, such as mean comparison or rank sum statistics, one can simply reweight the transformed samples from the second distribution by $r_X(X^{(2)})$. Notably, the recent approach by \citet{hu2024two} can be interpreted through this lens of importance-weighted rank sum statistics. Such importance-weighted statistics admit standard Gaussian calibration and are detailed in Appendix~\ref{sec:importance_weighted_statistics}. Below, we turn to two more involved examples.

\begin{example}[Classifier-based approach] \normalfont 
Let $\mathcal{H}$ be a class of binary classifiers $h: \mathcal{X} \times \mathcal{Y} \to \{1,2\}$ and let $\ell: \{1,2\} \times \{1,2\} \to \mathbb{R}$ be a loss function that penalizes incorrect predictions. The core idea behind classifier-based two-sample tests~\citep{lopez2017revisiting,kim2021classification,hediger2022use} is that when the null hypothesis of equality of distributions is true, any classifier will return a random guess. On the other hand, when two distributions are significantly different, the accuracy of a reasonable classifier exceeds chance level. Therefore, empirical classification accuracy serves as an effective test statistic. However, since we do not observe a sample from $f_{XY}$ but a sample from $f_{XY}^{(2)}$, we must account for this discrepancy when training the classifier. Specifically, we compute a classifier
\begin{align} \label{Eq: classifier}
    \widehat{h} = \argmin_{h \in \mathcal{H}}\biggl\{ \frac{1}{n_1} \sum_{i=1}^{n_1} \ell\bigl(h\bigl(V_i^{(1)}\bigr), 1\bigr) + \frac{1}{n_2} \sum_{i=1}^{n_2} r_X\bigl(X_i^{(2)}\bigr) \ell\bigl(h\bigl(V_i^{(2)}\bigr), 2\bigr) \biggr\},
\end{align}
and use the empirical classification accuracy of $\widehat{h}$, similarly weighted by the density ratio, as our test statistic. We analyze this in detail in \Cref{Section: Classifier-based Approach}.
\end{example}

\begin{example}[Kernel MMD] \normalfont \label{Example : Quadratic time MMD}
The second example is a kernel MMD statistic~\citep{gretton2012kernel}. Given a kernel $k$, the population MMD compares the kernel mean embeddings of two distributions with density functions $f_{XY}^{(1)}$ and $f_{XY}$, respectively. In kernel form, the squared MMD can be written as 
\begin{align*}
    \mathrm{MMD}^2 = \mE[k(V_1^{(1)}, V_2^{(1)})] + \mE[r_X\bigl(X_1^{(2)}\bigr)r_X\bigl(X_2^{(2)}\bigr)k(V_1^{(2)}, V_2^{(2)})] - 2 \mE[r_X\bigl(X_2^{(2)}\bigr)k(V_1^{(1)}, V_2^{(2)})],
\end{align*}
where the bias is corrected via importance weighting. The natural U-statistic estimator of $\mathrm{MMD}^2$ is degenerate under the null hypothesis, which makes its null distribution analytically intractable in general. In \Cref{Section: Kernel MMD}, we develop calibration procedures for two distinct regimes, namely a Gaussian regime via block constructions \citep{zaremba2013b}, and the fully quadratic regime via multiplier bootstrap.
\end{example}

In practice, the marginal density ratio $r_X$ must be estimated. While various techniques exist~\citep{sugiyama2010density, sugiyama2012density}, using the same dataset for estimation and testing causes plug-in bias. We thus recommend sample splitting, as in \Cref{Section: Classifier-based Approach} and \Cref{Section: Kernel MMD}. However, highly irregular or unbounded density ratios can destabilize the test statistic. While clipping or shrinkage~\citep{shimodaira2000improving} mitigates this, such settings should be handled with caution, as testing performance may suffer. This issue is empirically demonstrated in \Cref{Section: Numerical Experiments}. Although we focus on fixed subclasses of distributions for simplicity below, our framework naturally extends to sequences of null classes under uniform regularity conditions.

\subsection{Classifier-based Approach} 
\label{Section: Classifier-based Approach}

This subsection develops a classifier-based test for conditional two-sample testing. {To simplify the presentation, we assume equal sample sizes $n_1=n_2=n/2$ and split the dataset into two equal halves, $D_{\mathsf{infer}} := \{V_i^{(1)}\}_{i=1}^{\lfloor n/4\rfloor} \cup \{V_i^{(2)}\}_{i=1}^{\lfloor n/4\rfloor}$ and $D_{\mathsf{clf}} := \{V_i^{(1)}\}_{i=\lfloor n/4\rfloor+1}^{\lfloor n/2\rfloor} \cup \{V_i^{(2)}\}_{i=\lfloor n/4\rfloor+1}^{\lfloor n/2\rfloor}$. The second half $D_{\mathsf{clf}}$ is used to train the classifier $\widehat{h}$ as in \eqref{Eq: classifier}. For some positive integer $m < \lfloor n/4\rfloor$, we further divide the first half $D_{\mathsf{infer}}$ into a test set $D_{\mathsf{test}} := \{V_i^{(1)}\}_{i=1}^{m} \cup \{V_i^{(2)}\}_{i=1}^{m}$ of size $2m$, and a set $D_{\mathsf{ratio}} := D_{\mathsf{infer}} \setminus D_{\mathsf{test}}$ of size $2(\lfloor n/4\rfloor-m)$. Let $\widehat{r}_X$ denote an estimator of $r_X$ formed on $D_{\mathsf{ratio}}$.}

For $i \in \{1,\ldots,m\}$, define $\widehat{A}_{1,i} :=   \mathds{1}\{\widehat{h}(V_i^{(1)})=1\}$ and $\widehat{A}_{2,i} :=  \widehat{r}_X(X_i^{(2)}) \mathds{1}\{\widehat{h}(V_i^{(2)})=2\}$, and let $\overline{A}_1 :=  m^{-1} \sum_{i=1}^m \widehat{A}_{1,i}$ and $\overline{A}_2 :=  m^{-1} \sum_{i=1}^m \widehat{A}_{2,i}$. The population-level classification accuracy of $\widehat{h}$ is $\mP\{\widehat{h}(V^{(1)})=1\}/2 + \mE[r_X(X^{(2)}) \mathds{1}\{\widehat{h}(V^{(2)})=2\}]/2$, which equals $1/2$ for any classifier $\widehat{h}$ under the null hypothesis. This observation motivates the following test statistic
\begin{align} \label{Eq: Acc statistic}
\widehat{\mathrm{Acc}} :=  \frac{\sqrt{m}(\overline{A}_1 + \overline{A}_2 - 1)}{\sqrt{{\widehat{\sigma}_1^2 + \widehat{\sigma}_2^2}}},
\end{align}
where $\widehat{\sigma}_1^2 :=  (m-1)^{-1} \sum_{i=1}^m (\widehat{A}_{1,i} - \overline{A}_1)^2$ and $\widehat{\sigma}_2^2 :=  (m-1)^{-1} \sum_{i=1}^m (\widehat{A}_{2,i} - \overline{A}_2)^2$. To establish the limiting distribution of $\widehat{\mathrm{Acc}}$, we consider the following assumptions. 

\begin{assumption} \label{Assumption: classifier} 
Let $m_n := m$ be an increasing sequence of positive integers with $\lim_{n\to\infty} m_n = \infty$. 
Let $\mathcal{P}_0^{\mathsf{clf}} \subseteq \mathcal{P}_0$ be a subclass of null distributions such that
\begin{enumerate}[label=(\alph*)]
    \item For all sufficiently large $n$, there exist constants $0<\underline{c}_j<\overline{c}_j<1$ for $j=1,2$, and constants $C,\delta>0$ such that $\sup_{P \in \mathcal{P}_0^{\mathsf{clf}}} \mE_P[\{r_X(X^{(2)})\}^{2+\delta}] \leq C$, and with probability one,
    \begin{align*}
        \underline{c}_1 &\leq \inf_{P\in\mathcal P_0^{\mathsf{clf}}} \mP_P\{\widehat h(V^{(1)})=1\mid \widehat h\} \leq \sup_{P\in\mathcal P_0^{\mathsf{clf}}} \mP_P\{\widehat h(V^{(1)})=1\mid \widehat h\} \leq \overline{c}_1, \\
        \underline{c}_2 &\leq \inf_{P\in\mathcal P_0^{\mathsf{clf}}} \mE_P\big[r_{X}(X^{(2)})\mathds{1}\{\widehat h(V^{(2)})=2\}\mid \widehat h\big] \leq \sup_{P\in\mathcal P_0^{\mathsf{clf}}} \mE_P\big[r_{X}(X^{(2)})\mathds{1}\{\widehat h(V^{(2)})=2\}\mid \widehat h\big] \leq \overline{c}_2.
    \end{align*}
    \item For any $\epsilon > 0$, the density ratio estimator satisfies
    \begin{align*}
        \lim_{n \to \infty} \sup_{P \in \mathcal{P}_0^{\mathsf{clf}}} \mP_P \bigl( (n -m) \mE_P[\{\widehat{r}_X(X^{(2)}) - r_X(X^{(2)})\}^2 \given \widehat{r}_X] \geq \epsilon \bigr) = 0.
    \end{align*}
\end{enumerate}
\end{assumption}

Assumption~\ref{Assumption: classifier}(a) is imposed to establish the conditional central limit theorem for the test statistic with the true density ratio. It excludes classifiers that, under the null hypothesis, return the same prediction value regardless of inputs. Assumption~\ref{Assumption: classifier}(b) ensures that the approximation error from $\widehat{r}_X$ is asymptotically negligible, which closely parallels the assumption in \citet{hu2024two}. Although this requires a faster convergence rate than standard parametric cases (see Appendix~\ref{sec:convergence_rates}), we mitigate this by strategically allocating a larger sample to train $\widehat{r}_X$. This approach is theoretically supported by choosing a small $m$. Furthermore, in practice, we observe in \Cref{Section: Numerical Experiments} that the empirical results closely align with the expected normal approximation under an 8:2 split.

Under Assumption~\ref{Assumption: classifier}, the test statistic \eqref{Eq: Acc statistic} converges to $N(0,1)$ uniformly over $\mathcal{P}_0^{\mathsf{clf}}$. 
\begin{theorem} \label{Theorem: Asymptotic Normality of Classification Accuracy}
For the class of null distributions $\mathcal{P}_0^{\mathsf{clf}}$ in Assumption~\ref{Assumption: classifier}, $\widehat{\mathrm{Acc}}$ in \eqref{Eq: Acc statistic} satisfies
\begin{align*}
    \lim_{n \rightarrow \infty}\sup_{P \in \mathcal{P}_0^{\mathsf{clf}}} \sup_{t \in \mathbb{R}} \big| \mP_P(\widehat{\mathrm{Acc}} \leq t) - \Phi(t) \big| = 0.
\end{align*}
\end{theorem}

By \Cref{Theorem: Asymptotic Normality of Classification Accuracy}, rejecting the null hypothesis when $\widehat{\mathsf{Acc}} > \Phi^{-1}(1-\alpha)$ has asymptotic validity over $\mathcal{P}^{\mathsf{clf}}_0$ satisfying Assumption~\ref{Assumption: classifier}. We next improve the efficiency via $K$-fold cross-fitting on $D_{\mathsf{infer}}$, keeping the classifier $\widehat{h}$ fixed. We begin by considering $K$ disjoint subsets $D_{\textsf{infer},1},\ldots,D_{\textsf{infer},K}$ of $D_{\textsf{infer}}$, each of which contains $m_{\mathsf{cf}} := \lfloor n/(4K) \rfloor$ observations from each of the two samples. For $j \in \{1,\ldots,K\}$, let $\overline{A}_{1,j} + \overline{A}_{2,j} - 1$ and $\widehat{\sigma}_{1,j}^{2} + \widehat{\sigma}_{2,j}^{2}$ denote quantities analogous to $\overline{A}_{1} + \overline{A}_{2} - 1$ and $\widehat{\sigma}_{1}^{2} + \widehat{\sigma}_{2}^{2}$, respectively, by letting $D_{\textsf{test}} = D_{\textsf{infer},j}$ and $D_{\textsf{ratio}} = D_{\textsf{infer}} \setminus \! D_{\textsf{infer},j}$. We define the cross-fitted classification accuracy statistic as
\begin{align} \label{Eq: cv_Acc statistic}
{}^{\dagger}\widehat{\mathrm{Acc}} :=  \frac{1}{K^{1/2}}\sum_{j = 1}^K \frac{m_{\mathsf{cf}}^{1/2}(\overline{A}_{1,j} + \overline{A}_{2,j} - 1)}{\sqrt{{\widehat{\sigma}_{1,j}^2 + \widehat{\sigma}_{2,j}^2}}}.
\end{align}
Although the overlapping $D_{\mathsf{ratio}}$ sets induce dependence across folds, this plug-in effect is asymptotically negligible under Assumption~\ref{Assumption: classifier}(b).
The following corollary establishes that the cross-fitted accuracy statistic is asymptotically normally distributed under the null hypothesis. 

\begin{corollary} \label{Corollary: Asymptotic Normality of CV Accuracy}
Consider the same setting as in \Cref{Theorem: Asymptotic Normality of Classification Accuracy}. For any fixed $K \geq 2$, it holds that 
\begin{align*}
    \lim_{n \rightarrow \infty}\sup_{P \in \mathcal{P}_0^{\mathsf{clf}}} \sup_{t \in \mathbb{R}} \big| \mP_P\big({}^{\dagger}\widehat{\mathrm{Acc}} \leq t\big) - \Phi(t) \big| = 0.
\end{align*}
\end{corollary}

By \Cref{Corollary: Asymptotic Normality of CV Accuracy}, the test that rejects the null when ${}^{\dagger}\widehat{\mathrm{Acc}} > \Phi^{-1}(1-\alpha)$ controls the asymptotic type I error.

Many practical classifiers attempt to mimic the Bayes optimal classifier. For the balanced-sample setting, the Bayes optimal classifier is $h^\star(x,y) :=  1+\mathds{1}\bigl(f_{XY}^{(1)}(x,y) / \{f_{XY}^{(1)}(x,y) + f_{XY}(x,y)\} \leq 1/2\bigr)$. The classification accuracy of $h^\star$ can be explicitly computed in terms of the total variation (TV) distance. Specifically, twice the classification accuracy equals 
\begin{align*}
\mP\{h^\star(V^{(1)})=1\} + \mE[r_X(X^{(2)}) \mathds{1}\{h^\star(V^{(2)})=2\}] = 1 + \mathrm{TV}(f_{XY}^{(1)}, f_{XY}),
\end{align*}
where $\mathrm{TV}(f_{XY}^{(1)}, f_{XY})$ denotes the TV distance between two distributions with densities $f_{XY}^{(1)}$ and $f_{XY}$, respectively. Since the TV distance equals zero if and only if two distributions are identical, our classifier-based test can be powerful against general alternatives when the employed classifier approximates the Bayes classifier. 

We now turn to a unified kernel-based MMD framework in the next subsection, developing parallel results that bridge the block-wise and quadratic-time statistics.

\subsection{Kernel MMD}
\label{Section: Kernel MMD}

{In this subsection, we introduce a unified block-wise MMD framework governed by a parameter $\gamma \in [0,1]$, encompassing both the block-wise estimator \citep{zaremba2013b} and the quadratic-time MMD \citep[Theorem~12]{gretton2012kernel}. As in \Cref{Section: Classifier-based Approach}, we assume equal sample sizes $n_1 = n_2 = n/2$ for simplicity. Since no auxiliary classifier needs to be trained in this setting, we allocate the entire sample to inference, so that $\tilde{D}_{\mathsf{infer}}= \{V_i^{(1)}\}_{i=1}^{n/2} \cup \{V_i^{(2)}\}_{i=1}^{n/2}$ coincides with the full dataset. We partition $\tilde{D}_{\mathsf{infer}}$ via sample splitting to decouple the estimation of the density ratio from the evaluation of the test statistic and thereby prevent plug-in bias. For some positive integer $m < n$, we define an evaluation set $\tilde{D}_{\mathsf{test}} := \{V_i^{(1)}\}_{i=1}^{m} \cup \{V_i^{(2)}\}_{i=1}^{m}$ of size $2m$ to compute the MMD statistic, and an estimation set $\tilde{D}_{\mathsf{ratio}} := \tilde{D}_{\mathsf{infer}} \setminus \tilde{D}_{\mathsf{test}}$ of size $n-2m$ to construct the density ratio estimator $\widehat{r}_X$.}

{Let $\mathcal{P}_{0}^{\mathsf{mmd}} \subseteq \mathcal{P}_0$ be a subclass of null distributions satisfying $H_0$ in \eqref{Eq: hypothesis}. For each $i\in\{1,\ldots,m\}$, write $W_i := (V_i^{(1)}, V_i^{(2)})$. We define the pairwise kernel contribution as
\begin{align*}
    \widehat{H}(W_i,W_j)
    &:= k(V_i^{(1)},V_j^{(1)}) + \widehat{r}_X(X_i^{(2)})\widehat{r}_X(X_j^{(2)})k(V_i^{(2)},V_j^{(2)}) \\
    &\qquad - \widehat r_X(X_i^{(2)})k(V_i^{(2)},V_j^{(1)}) - \widehat r_X(X_j^{(2)})k(V_i^{(1)},V_j^{(2)}),
\end{align*}
and abbreviate $\widehat H_{ij}:=\widehat H(W_i,W_j)$. Setting the block size $B = \max\{2,\lfloor m^\gamma \rfloor\}$, the number of blocks $S = \lfloor m/B \rfloor$, and the index sets $\mathcal{I}_b := \{(b-1)B+1,\dots,bB\}$ for $b=1,\dots,S$, we then consider the generalized block-wise $U$-statistic
\begin{align}
    \label{equation : unified_statistic}
    \widehat{\mathrm{GB}}_\gamma = \frac{1}{S} \sum_{b=1}^{S} \bigg(\frac{1}{B-1}\sum_{\substack{i,j\in\mathcal I_b\\ i\neq j}}\widehat H_{ij}\bigg).
\end{align}
The parameter $\gamma$ controls the number of blocks $S$, which drives the asymptotic behavior. For $0 \leq \gamma < 1$, the growing number of blocks ($S\to\infty$) resolves the kernel degeneracy under $H_0$, yielding asymptotic normality. When $\gamma = 1$, the data form a single block ($S=1$), recovering the standard $U$-statistic with a chi-square-type limit. To unify these two regimes, we define the empirical statistic $\widehat{\mathrm{MMD}}^2_{\gamma}$ and its population counterpart $\mathrm{MMD}^{2}_{\gamma}$ as
\begin{align*}
    \widehat{\mathrm{MMD}}_\gamma^2 := 
    \begin{cases}
        \sqrt{S}\,\widehat{\mathrm{GB}}_\gamma/\widehat{\sigma}_\gamma, & 0 \leq \gamma < 1, \\[0.4em]\sqrt{\frac{m-1}{m}}\,\widehat{\mathrm{GB}}_1, & \gamma = 1,
    \end{cases}
    \qquad
    \mathrm{MMD}_\gamma^2:=
    \begin{cases}
        \sqrt{S}\,\mathrm{GB}_\gamma/\sigma_\gamma, & 0 \leq \gamma < 1, \\[0.4em]\sqrt{\frac{m-1}{m}}\,\mathrm{GB}_1, & \gamma = 1.
    \end{cases}
\end{align*}
Here, the oracle quantities $H_{ij}$ and $\mathrm{GB}_\gamma$ are obtained by replacing $\widehat{r}_X$ with the true $r_X$ in $\widehat{H}_{ij}$ and $\widehat{\mathrm{GB}}_\gamma$, respectively. The variance $\sigma_\gamma^2$ denotes the population variance of the oracle block statistics $\frac{1}{B-1}\sum_{i \neq j \in \mathcal{I}_b}H_{ij}$ under $P\in\mathcal{P}_{0}^{\mathsf{mmd}}$, and $\widehat{\sigma}_\gamma^2$ is its sample counterpart. Explicit forms are provided in Appendix~\ref{Proof of theorem: MMD Asymptotic Distribution}. To establish the asymptotic behavior of $\widehat{\mathrm{MMD}}^2_\gamma$ for all $\gamma \in [0,1]$, we impose the following assumptions.}
\vspace{-0.2em}
{\begin{assumption}\label{Assumption : Unified MMD}
    Let $m_n:=m$ be an increasing sequence of positive integers with $\lim_{n\to\infty}m_n=\infty$. Consider a class of null distributions $\mathcal{P}^{\mathsf{mmd}}_0$ such that
    \begin{enumerate}[label=(\alph*), itemsep=0.1em]
        \item There exist constants $c_1,c_2>0$ and some $\delta \in (0,1]$ such that
        \begin{align*}
            \inf _{P \in \mathcal{P}_{0}^{\mathsf{mmd}}} \mE_P\big(H^{2}_{12}\big) \geq c_1 \quad \text{and} \quad \sup _{P \in \mathcal{P}_{0}^{\mathsf{mmd}}} \mE_P\big(|H_{12}|^{2+\delta}\big) \leq c_2.
        \end{align*}
        \item The population density ratio satisfies $\sup_{P\in\mathcal{P}_{0}^{\mathsf{mmd}}}\mE_P[r_X(X^{(2)})^4]< c_3$ for some constant $c_3>0$, and the corresponding estimator approximates $r_X$ at the rate
        \begin{align*}
            \sup _{P \in \mathcal{P}_{0}^{\mathsf{mmd}}}\mE_P\big[\{\widehat{r}_X(X^{(2)})-r_X(X^{(2)})\}^2\big]=o\big((n-m)^{-1/2}\big).
        \end{align*}
        \item The kernel is uniformly bounded, i.e., $\|k\|_{\infty} \leq c_4$ for some constant $c_4>0$.
        \item The subsample sizes grow such that $m^{1+\gamma}/(n-m) \to c_5 \in [0,\infty)$ for $0 \leq \gamma \leq 1$.
    \end{enumerate}
\end{assumption}}

{Assumption~\ref{Assumption : Unified MMD}(a) and (c) impose standard moment and boundedness conditions on the kernel to guarantee uniform distributional convergence. The boundedness condition~(c) is readily satisfied by practical choices such as the Gaussian kernel. Assumption~\ref{Assumption : Unified MMD}(b) requires the plug-in error from estimating $r_X$ to be asymptotically negligible. Compared to Assumption~\ref{Assumption: classifier}(b), this requirement is relatively mild and attainable under broader nonparametric smoothness conditions. While a second-moment condition on the true density ratio $r_X$ identical to that of \cite{hu2024two} suffices for distributional convergence, we additionally impose a uniform fourth-moment bound on $r_X$ to establish bootstrap consistency. Nevertheless, our overall requirement for density-ratio estimation remains strictly weaker than theirs (we provide a detailed discussion of these convergence rates in Appendix~\ref{sec:convergence_rates}). Finally, Assumption~\ref{Assumption : Unified MMD}(d) regulates the relative growth of $m$ and $n-m$. Because a larger block parameter $\gamma$ aggregates more kernel interactions and thereby magnifies the plug-in error, $n-m$ must grow sufficiently fast to preserve the limiting distribution. This naturally motivates allowing distinct splitting proportions for test-statistic construction and density-ratio estimation \citep[see][Example~1]{bordino2026nonparametric}.}

{We now characterize the asymptotic null distribution of $\widehat{\mathrm{MMD}}^2_{\gamma}$ for $\gamma \in [0,1]$.}

\begin{theorem}\label{theorem: MMD Asymptotic Distribution}
{For the class of null distributions $\mathcal{P}_{0}^{\mathsf{mmd}}$ satisfying Assumption~\ref{Assumption : Unified MMD}, the fixed $\gamma\in[0,1]$ determines the asymptotic behavior of the test statistic:}
\begin{enumerate}[itemsep=0.1em]
    \item {\textbf{\normalfont{Regime I} ($0 \leq \gamma < 1$):} The studentized statistic converges to $N(0,1)$ as}
    \begin{align*}
        {\lim _{n \rightarrow \infty}
            \sup _{P \in \mathcal{P}_{0}^{\mathsf{mmd}}}
            \sup _{t \in \mathbb{R}}
            \Big|
            \mP_{P}\big(\widehat{\mathrm{MMD}}^2_\gamma \leq t\big)-\Phi(t)
            \Big|
            =0.}
    \end{align*}
    
    \item {\textbf{\normalfont{Regime II} ($\gamma = 1$):} The MMD statistic with $\gamma=1$ satisfies}
    \begin{align*}
        {\lim_{n \to \infty}
            \sup_{P \in \mathcal{P}_{0}^{\mathsf{mmd}}}
            \sup_{t \in \mathbb{R}}
            \Big|
            \mP_{P} \big(\widehat{\mathrm{MMD}}^2_1 \leq t\big) - \mP_{P}(G \leq t)
            \Big|
            = 0,}
    \end{align*}
    {where $G = \sum_{\ell=1}^{\infty} \lambda_{\ell}(\chi^2_{1,\ell}-1)$, $\{\chi^{2}_{1,\ell}\}_{\ell\geq1}$ are independent chi-square random variables with one degree of freedom, and $\{\lambda_{\ell}\}_{\ell\geq1}$ are the eigenvalues of the integral operator $g \mapsto \int H(w, \cdot)g(w)dP(w)$.}
\end{enumerate}
\end{theorem}

{\Cref{theorem: MMD Asymptotic Distribution} shows that the asymptotic regime is determined entirely by $\gamma$: Regime I ($0\leq \gamma<1$) is asymptotically standard Normal after studentization, whereas Regime II ($\gamma=1$) yields a non-Gaussian quadratic-form limit. Meanwhile, Assumption~\ref{Assumption : Unified MMD}(d) requires the estimation sample $\tilde{D}_{\mathsf{ratio}}$ to be sufficiently large relative to the testing sample $\tilde{D}_{\mathsf{test}}$. Under a single split, this creates an efficiency trade-off: improving density-ratio estimation leaves fewer observations for computing the MMD statistic. We therefore consider $K$-fold cross-fitting, which preserves the decoupling between estimation and evaluation while using the data more efficiently.}

Fix an integer $K \geq 1$ such that $K \leq \lfloor n/(2m) \rfloor$. In particular, $K=1$ recovers the single-split construction above. For each fold $j\in\{1,\dots,K\}$, let $\tilde{D}_{\mathsf{test},j}:=\{V_i^{(1)}\}_{i=(j-1)m+1}^{jm}\cup\{V_i^{(2)}\}_{i=(j-1)m+1}^{jm}$ and $\tilde{D}_{\mathsf{ratio},-j}:=\tilde{D}_{\mathsf{infer}}\setminus \tilde{D}_{\mathsf{test},j}$. Let $\widehat{\mathrm{GB}}_{\gamma,j}$ denote the statistic in \eqref{equation : unified_statistic} computed on $\tilde{D}_{\mathsf{test},j}$ using the density-ratio estimator $\widehat r_{X,-j}$ constructed from $\tilde{D}_{\mathsf{ratio},-j}$. We define
\begin{align*}
    {}^{\dagger}\widehat{\mathrm{MMD}}_{\gamma}^2 :=
    \begin{cases}
        \frac{1}{\sqrt{K}}\sum_{j=1}^K \sqrt{S}\,\widehat{\mathrm{GB}}_{\gamma,j}/\widehat{\sigma}_{\gamma,j}, & 0 \leq \gamma < 1, \\[0.4em]
        \tfrac{1}{K}\sum_{j=1}^K \sqrt{\frac{m-1}{m}}\,\widehat{\mathrm{GB}}_{1,j}, & \gamma = 1,
    \end{cases}
\end{align*}
where $\widehat{\sigma}_{\gamma,j}^2$ is the fold-$j$ analogue of $\widehat{\sigma}_\gamma^2$, computed on $\tilde{D}_{\mathsf{test},j}$ using $\widehat{r}_{X,-j}$. 
Replacing $\widehat{r}_{X,-j}$ with the true $r_X$ yields the oracle counterpart ${}^{\dagger}\mathrm{MMD}_\gamma^2$, defined by the same expression with $H_{ij}$ and $\sigma_{\gamma,j}$ in place of $\widehat{H}_{ij}$ and $\widehat{\sigma}_{\gamma,j}$. In Regime~II, we abbreviate the fold-level oracle statistic as 
$\mathrm{MMD}_{1,j}^2 := \sqrt{\frac{m-1}{m}}\mathrm{GB}_{1,j}$,so that 
${}^{\dagger}\mathrm{MMD}_1^2 = K^{-1}\sum_{j=1}^K\mathrm{MMD}_{1,j}^2$. When $K=1$, both statistics reduce to their single-split counterparts. As with the classifier-based approach, while overlapping estimation sets $\tilde{D}_{\mathsf{ratio},-j}$ induce dependence across folds for the plug-in statistics, this effect is asymptotically negligible under \Cref{Assumption : Unified MMD}(b).

{The following corollary extends \Cref{theorem: MMD Asymptotic Distribution} to the cross-fitted statistic.}

\begin{corollary}\label{cor: Cross-fitted MMD Asymptotic Distribution}
{Suppose Assumption~\ref{Assumption : Unified MMD} holds, and fix $K \geq 1$.}
\begin{enumerate}[itemsep=0.1em]
    \item {\textbf{\normalfont{Regime I} ($0 \leq \gamma < 1$):}}
    \begin{align*}
        {\lim_{n\to\infty}
            \sup_{P\in\mathcal P_{0}^{\mathsf{mmd}}}
            \sup_{t\in\mathbb R}
            \Big|
            \mP_P\big({}^{\dagger}\widehat{\mathrm{MMD}}_\gamma^2 \leq t\big)-\Phi(t)
            \Big|
            =0.}
    \end{align*}
    
    \item {\textbf{\normalfont{Regime II} ($\gamma = 1$):} Let $G_1,\dots,G_K$ be independent copies of $G=\sum_{\ell=1}^{\infty}\lambda_\ell(\chi_{1,\ell}^2-1)$. Then}
    \begin{align*}
        {\lim_{n\to\infty}
            \sup_{P\in\mathcal P_{0}^{\mathsf{mmd}}}
            \sup_{t\in\mathbb R}
            \Big|
            \mP_P\big({}^{\dagger}\widehat{\mathrm{MMD}}_1^2 \leq t\big)
            -
            \mP_P\Big(\frac{1}{K}\sum_{j=1}^K G_j \leq t\Big)
            \Big|
            =0.}
    \end{align*}
\end{enumerate}
\end{corollary}

{For Regime~I ($0 \leq \gamma < 1$), \Cref{cor: Cross-fitted MMD Asymptotic Distribution} shows that rejecting $H_0$ when ${}^{\dagger}\widehat{\mathrm{MMD}}_\gamma^2 > \Phi^{-1}(1-\alpha)$ controls the type~I error at level $\alpha$ uniformly over $\mathcal{P}_0^{\mathsf{mmd}}$.
For Regime~II ($\gamma = 1$), the oracle statistics $\mathrm{MMD}_{1,j}^2$ are computed on disjoint evaluation sets and hence independent, but the limiting distribution $K^{-1}\sum_{j=1}^K G_j$ has no closed-form quantile. 
We therefore calibrate via multiplier bootstrap.}

{For each fold $j \in \{1,\ldots,K\}$, let $\{\xi_{i,j} : 1 \leq i \leq m\}$ be independent standard Gaussian multipliers, and let $\widehat{H}_{ii'}^{(j)}$ denote the plug-in kernel computed on $\tilde{D}_{\mathsf{test},j}$ using $\widehat{r}_{X,-j}$. Define
\begin{align*}
    \widehat{\mathrm{MMD}}_{1,\mathrm{Boot},j}^2 := \frac{1}{\sqrt{m(m-1)}}\sum_{1 \leq i \neq i' \leq m}\xi_{i,j}\xi_{i',j}\widehat{H}_{ii'}^{(j)}, \quad {}^{\dagger}\widehat{\mathrm{MMD}}_{1,\mathrm{Boot}}^2 := \frac{1}{K}\sum_{j=1}^K \widehat{\mathrm{MMD}}_{1,\mathrm{Boot},j}^2.
\end{align*}
For $R$ multiplier draws $\Xi_R=\big\{\{\xi_{i,j}^{(r)}:1\leq i\leq m,\ 1\leq j\leq K\}\big\}_{r=1}^R$, let $\widehat q_{1-\alpha}^{R}$ denote the empirical $(1-\alpha)$-quantile of $\{{}^{\dagger}\widehat{\mathrm{MMD}}_{1,\mathrm{Boot}}^{2,(r)}\}_{r=1}^R$. Define the test ${}^{\dagger}\phi_\alpha^{R}(\tilde{D}_{\mathsf{infer}},\Xi_R):=\mathds{1}\{{}^{\dagger}\widehat{\mathrm{MMD}}_1^2>\widehat q_{1-\alpha}^{R}\}$.}
\medskip

{The next theorem establishes that this bootstrap calibration consistently approximates the null distribution in Regime II and therefore yields an asymptotically valid test.}
{\begin{theorem}\label{thm: Cross-fitted Bootstrap Validity}
    Suppose Assumption~\ref{Assumption : Unified MMD} holds with $\gamma=1$, and fix $K\geq1$.
    \begin{enumerate}[label=(\roman*), itemsep=0.1em]
        \item Writing $\mP_P^*(\cdot):=\mP_P(\cdot\mid \tilde{D}_{\mathsf{infer}})$, for any $\varepsilon>0$,
        \begin{align*}
            \lim_{n\to\infty}
            \sup_{P\in\mathcal P_{0}^{\mathsf{mmd}}}
            \mP_P\bigg\{
            \sup_{t\in\mathbb R}
            \Big|
            \mP_P^*\big({}^{\dagger}\widehat{\mathrm{MMD}}_{1,\mathrm{Boot}}^2 \leq t\big)
            -
            \mP_P\big({}^{\dagger}\mathrm{MMD}_1^2 \leq t\big)
            \Big|
            >
            \varepsilon
            \bigg\}
            =0.
        \end{align*}
        \item Let ${}^{\dagger}F_{0,\scriptscriptstyle P}$ denote the limiting null distribution of ${}^{\dagger}\mathrm{MMD}_1^2$ under $P$, and define ${}^{\dagger}q_{1-\alpha,\scriptscriptstyle P} := \inf\{t \in \mathbb{R} : {}^{\dagger}F_{0,\scriptscriptstyle P}(t) \geq 1-\alpha\}$.  Suppose that for every $\tau>0$,
        \begin{align*}
        \inf_{P\in\mathcal{P}_{0}^{\mathsf{mmd}}}\inf_{|t - {}^{\dagger}q_{1-\alpha,\scriptscriptstyle P}| \geq \tau}\big|{}^{\dagger}F_{0,\scriptscriptstyle P}(t)-(1-\alpha)\big|> 0.
        \end{align*}
        Then, for any fixed $\alpha\in(0,1)$,
        \begin{align*}
            \lim_{n,R\to\infty}
            \sup_{P\in\mathcal P_{0}^{\mathsf{mmd}}}
            \Big|
            \mP_P\big(
            {}^{\dagger}\phi_\alpha^R(
            \tilde{D}_{\mathsf{infer}},\Xi_R)=1
            \big)-\alpha
            \Big|
            =0.
        \end{align*}
    \end{enumerate}
\end{theorem}
\Cref{thm: Cross-fitted Bootstrap Validity} demonstrates that the conditional distribution of the bootstrap statistic uniformly approximates the oracle null distribution. Consequently, the bootstrap critical value yields an asymptotically valid test with uniform level $\alpha$ over the null class $\mathcal{P}_0^{\mathsf{mmd}}$, thereby obviating the need for an explicit closed-form limiting distribution.}

When the kernel $k$ is a characteristic kernel, the population MMD equals zero if and only if two distributions coincide \citep{fukumizu2007kernel}. Hence, as with the classifier-based tests in \Cref{Section: Classifier-based Approach}, the MMD-based tests can be powerful against general alternatives, provided that the density ratio $r_X$ can be accurately estimated.

The next section illustrates the numerical performance of the proposed tests in comparison to existing methods.

\section{Numerical Experiments} 
\label{Section: Numerical Experiments}

\subsection{Experimental Setup}
\label{subsec: Experimental Setup}
In this section, we evaluate the numerical performance of several conditional two-sample tests on both synthetic and real datasets. The methods include conditional independence tests adapted via the framework of Section~\ref{Section: Approach via Conditional Independence Testing}, density-ratio-based tests, and conditional kernel-based tests of \citet{yan2025distancekernelbasedmeasuresglobal}. An overview of all methods is provided in Section~\ref{Section: Overview of Testing Methods}, with implementation details deferred to Appendix~\ref{Section: Experimental Details}. All empirical rejection rates reported below are computed from $500$ Monte Carlo replications at nominal level $\alpha = 0.05$, and code reproducing every experiment is available at \url{https://github.com/suman-cha/Cond2ST}.

Synthetic experiments are presented in Section~\ref{Section: Synthetic Data Examples} and evaluate three scenarios under a fixed covariate dimension $p = 10$, with sample sizes $n \in \{200, 500, 1000, 2000\}$. Each scenario admits two variants, corresponding to \emph{unbounded}~(U) and \emph{bounded}~(B) marginal density ratios $r_X$, which together span a range of difficulty for density-ratio estimation. For every DRT method, $r_X$ is estimated by linear logistic regression \citep[LLR, Section~3]{sugiyama2010density}. For CIT methods that rely on nuisance regressions, we use random forests in the generalized covariance measure \citep{shah2020hardness} and the projected covariance measure \citep{lundborg2024projected}, and XGBoost in the WGSC procedure \citep{williamson2023general}. Every CIT method is converted to a conditional two-sample test through Algorithm~\ref{Algorithm: Converting CIT into C2ST} with adjustment parameter $\varepsilon = \{\log(n)\}^{-1/2}$.

Real-data experiments are presented in Section~\ref{Section: Real Data Analysis} and use two datasets, namely the diamonds dataset~($p=6$) available in the \textsf{R} package \texttt{ggplot2} and the superconductivity dataset~($p=81$) compiled by \citet{Hamidieh2018}. Following \citet{kim2024conditional}, each dataset is treated as a finite population from which we draw samples, enabling controlled experiments with a known ground truth. Both $X$ and $Y$ are standardized to zero mean and unit variance prior to sampling. A covariate shift between the two groups is induced by drawing $X^{(1)}$ uniformly from the population and $X^{(2)}$ with probability proportional to $\exp(-x_1^2)$, where $x_1$ denotes the first coordinate of $X$. Under the null, both $Y^{(1)}$ and $Y^{(2)}$ are drawn uniformly, which removes any dependence on $X$ and yields identical conditional distributions. Under the alternative, $Y^{(1)}$ remains uniform while $Y^{(2)}$ is drawn with probability proportional to $\exp(-y)$, which produces a detectable discrepancy between the two conditional distributions. The sample size is varied over $n \in \{200, 400, 800, 1200, 1600, 2000\}$. In contrast to the synthetic setting, we estimate $r_X$ using both LLR and kernel logistic regression~(KLR) on the DRT methods.

\subsection{Overview of Testing Methods}
\label{Section: Overview of Testing Methods}

\mypara{DRT methods}
The cross-fitted classifier-based test ${}^{\dagger}\mathrm{CLF}$ of Section~\ref{Section: Classifier-based Approach} and three cross-fitted instances of the kernel MMD framework of Section~\ref{Section: Kernel MMD} constitute DRT methods. The MMD instances correspond to three choices of the block-size parameter $\gamma \in [0, 1]$: ${}^{\dagger}\mathrm{MMD}$-$\ell$ ($\gamma = 0$, linear-time), ${}^{\dagger}\mathrm{MMD}$-$b$ ($\gamma = 0.5$, block-wise), and ${}^{\dagger}\mathrm{MMD}$-$q$ ($\gamma = 1$, quadratic-time calibrated by the multiplier bootstrap of Theorem~\ref{thm: Cross-fitted Bootstrap Validity}). We also include the conformal prediction test \citep[$\mathrm{CP}$,][]{hu2024two} and its debiased variant \citep[$\mathrm{DCP}$,][]{chen2025biased}. For these DRT methods, we employ an 8:2 sample split (80$\%$ for density-ratio estimation) to ensure stable estimation and rigorous type~I error control, accepting a potential power loss as discussed by \citet{hu2024two}.

\vspace{0.3em}
\mypara{CIT methods}
Four conditional independence tests are considered, each paired with Algorithm~\ref{Algorithm: Converting CIT into C2ST}, namely $\mathrm{RCIT}$ \citep{strobl2019approximate}, $\mathrm{GCM}$ \citep{shah2020hardness}, $\mathrm{PCM}$ \citep{lundborg2024projected}, and $\mathrm{WGSC}$ \citep{williamson2023general}. Empirical sensitivity of each test to the subsampling step of Algorithm~\ref{Algorithm: Converting CIT into C2ST} is examined in Appendix~\ref{Appendix: With_Without_Algorithm_1}.

\vspace{0.3em}
\mypara{Conditional kernel methods} Finally, we include the conditional maximum mean discrepancy test ($\mathrm{CMMD}$) and the conditional generalized energy distance test ($\mathrm{CGED}$) of \citet{yan2025distancekernelbasedmeasuresglobal}, evaluated via the \textsf{R} package \texttt{KDist}. Neither test requires density-ratio estimation.

\begin{figure}[t]
\centering
\includegraphics[width=1\textwidth]{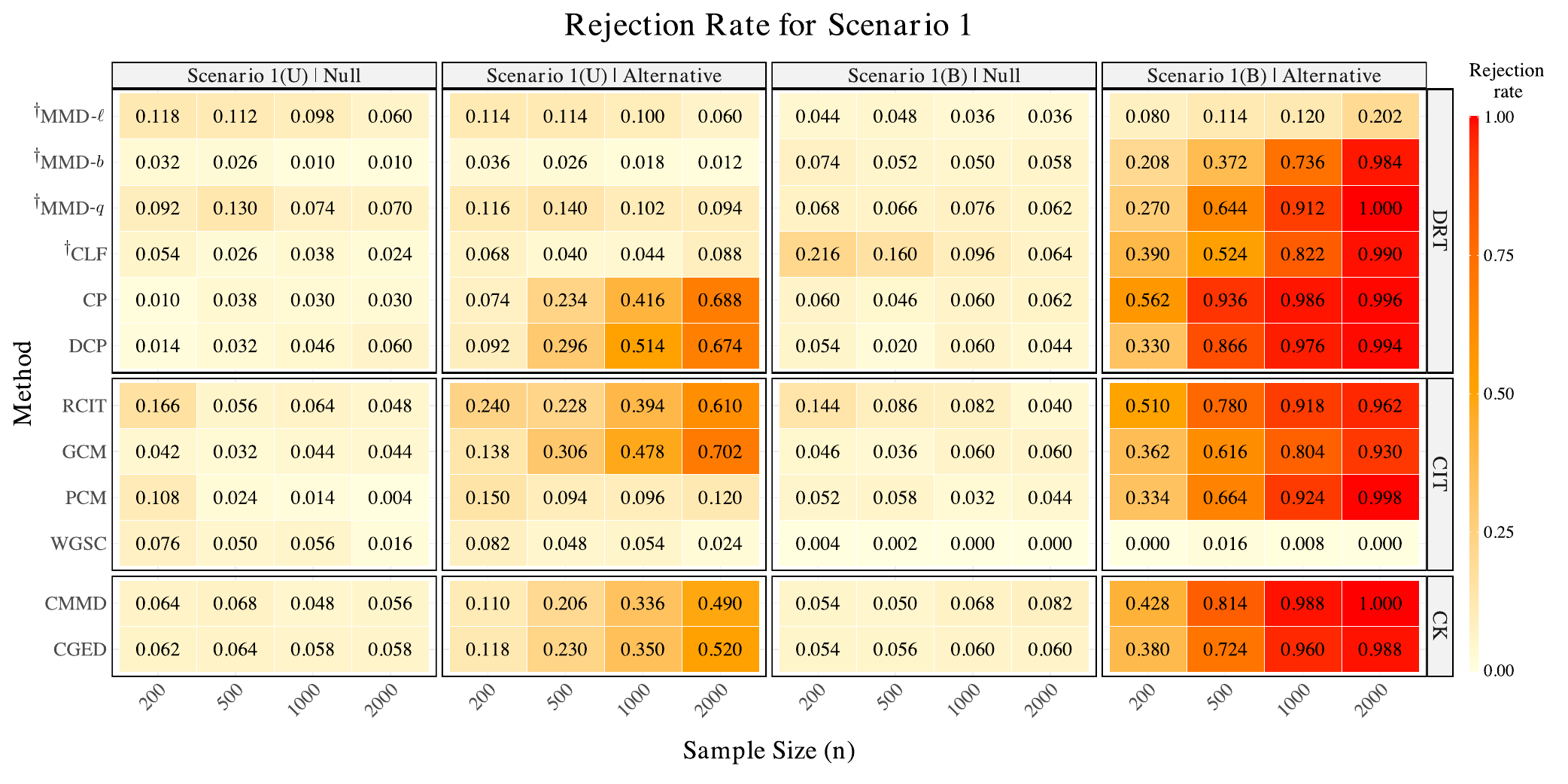}
\captionsetup{skip=0pt}  
\caption{Rejection rates for Scenario~1 under null and alternative hypotheses, shown for both unbounded~(U) and bounded~(B) settings. Results are averaged over 500 repetitions with $\alpha = 0.05$.}
\label{fig:Scenario 1}
\end{figure}

\subsection{Synthetic Data Examples} 
\label{Section: Synthetic Data Examples}

Across all three scenarios, $X$ shares the same marginal distributions. Setting~(U) uses $X^{(1)} \sim N(0, I_p)$ and $X^{(2)} \sim N(\mu, I_p)$ with $\mu = (1, 1, -1, -1, 0, \ldots, 0)^{\top}$, under which the resulting density ratio $r_X$ is unbounded. Setting~(B) truncates both distributions to $[-0.5, 0.5]^p$, which enforces boundedness of $r_X$.

\vspace{0.3em}
\mypara{Scenario 1: Linear Model with Mean Shift}
Following \citet{hu2024two}, this scenario investigates the efficacy of testing methods in detecting the mean difference between two linear models. For each $j \in \{1, 2\}$, we set $y^{(j)} \given x^{(j)} = \delta^{(j)} + x^{(j) \top}\beta + \epsilon^{(j)}$, where $\epsilon^{(j)}$ follows a $t$-distribution with 2 degrees of freedom. The regression coefficient $\beta$ is set to $(1, -1, -1, 1, 0, \ldots, 0)^{\top}$. Under the null hypothesis, we set $\delta^{(1)} = \delta^{(2)} = 0$, while for the alternative hypothesis, we introduce a mean shift by setting $\delta^{(1)} = 0$ and $\delta^{(2)} = 0.5$, thereby inducing a difference in the two conditional distributions.

\vspace{0.3em}
\mypara{Scenario 2: High Variability in Conditional Distribution}
We also investigate the effect of high variability in the conditional distribution, adapting the example outlined in \citet[][Section 6.2]{CHATTEJEE2024conditional}. Under the null hypothesis, we model the conditional distributions as $y^{(j)} \given x^{(j)} \sim N\left(x^{(j) \top}\beta^{(j)}, (\sigma^{(j)})^2 \right)$, where $\beta^{(j)} = \mathbf{1}_{p}$, a $p$-dimensional vector of ones, and $(\sigma^{(j)})^2 = 10^2$ for both $j \in \{1, 2\}$. For the alternative hypothesis, we modify $\beta^{(2)}$ to $(1, \ldots, 1, 0)^\top$ and introduce heteroscedasticity by varying the variance for $j = 2$ as $(\sigma^{(2)})^2 = 10\left(1 + \exp\left(-\lVert x^{(2)} - 0.5 \mathbf{1}_{p} \rVert^2_{2}/64 \right)\right)$. 

\begin{figure}[t]
\centering
\includegraphics[width=1\textwidth]{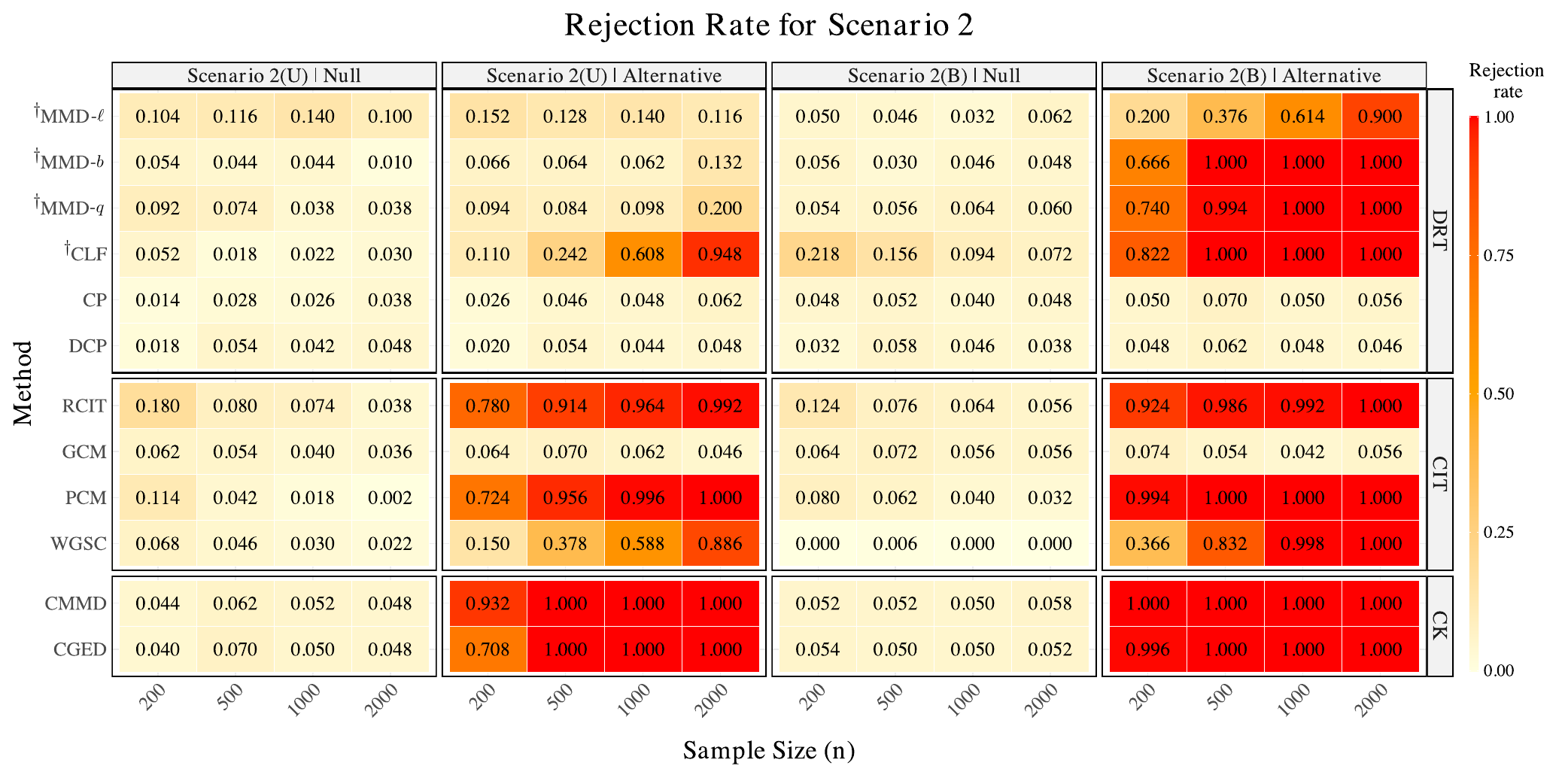}
\captionsetup{skip=0pt}  
\caption{Rejection rates for Scenario 2 under null and alternative hypotheses, shown for both unbounded (U) and bounded (B) settings. Results are averaged over 500 repetitions with $\alpha = 0.05$.}
\label{fig:Scenario 2}
\end{figure}

\vspace{0.3em}
\mypara{Scenario 3: Post-Nonlinear Model}
Our final scenario considers a post-nonlinear (PNL) model, which is widely used in causal predictive inference~\citep{zhang2017causal,Li2023knn}. This scenario tests the ability of the methods to detect differences in nonlinear relationships between variables. We model the conditional distributions as $y^{(j)} \given x^{(j)} = f^{(j)}(x^{(j)\top}\mathbf{1}_{p} + 2\epsilon^{(j)})$, where $\epsilon^{(j)} \sim N(0,1)$ for $j \in \{1, 2\}$. Under the null hypothesis, we set $f^{(j)}(x) = \cos(x)$ for both $j \in \{1, 2\}$, while for the alternative hypothesis, $f^{(2)}(x)$ is randomly sampled from the set $\{x, x^2, x^3, \sin(x), \tanh(x)\}$.

\begin{figure}[t]
\centering
\includegraphics[width=1\textwidth]{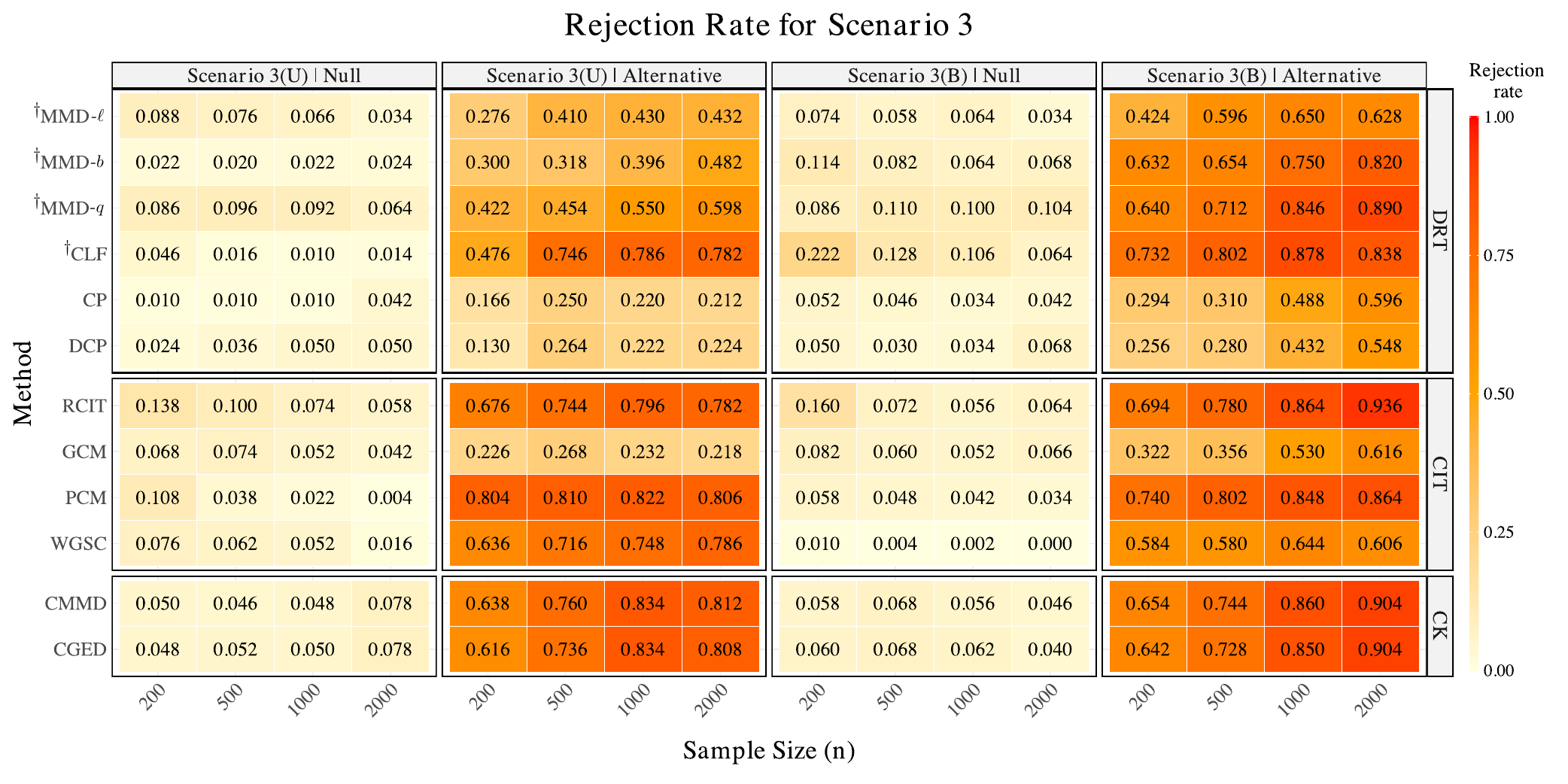}
\captionsetup{skip=0pt}  
\caption{Rejection rates for Scenario 3 under null and alternative hypotheses, shown for both unbounded (U) and bounded (B) settings. Results are averaged over 500 repetitions with $\alpha = 0.05$.}
\label{fig:Scenario 3}
\end{figure}

\Cref{fig:Scenario 1,fig:Scenario 2,fig:Scenario 3}
collectively reveal four patterns, which we summarize before discussing the scenarios. First, DRT methods are markedly more powerful under setting~(B), where the density ratio is bounded and therefore easier to estimate, whereas CIT and conditional-kernel methods (CMMD, CGED) are comparatively insensitive to the difficulty of density ratio estimation. Second, within the DRT family, methods that require only the marginal density ratio ($^{\dagger}$MMD-$\ell$, $^{\dagger}$MMD-$b$, $^{\dagger}$MMD-$q$, $^{\dagger}\text{CLF}$) dominate those requiring the conditional ratio (CP, DCP) in Scenario~2, where high conditional variability makes the latter quantity difficult to estimate. Third, no single method dominates uniformly. For instance, GCM performs well in Scenario~1 but loses power in Scenario~2, WGSC exhibits the reverse pattern, and the influence of the regression backbone (random forests versus XGBoost) is documented in \supple. Fourth, RCIT rejects at a rate exceeding~$\alpha$ at small sample sizes in every scenario, warranting caution in small-sample applications.

Scenario~1 is favorable to mean-based contrasts. The tests $^{\dagger}$CLF, CP, DCP, $^{\dagger}$MMD-$b$, and $^{\dagger}$MMD-$q$ reach near-unit power, with $^{\dagger}$MMD-$\ell$ trailing by design and PCM leading the CIT family under~(B). Scenario~2 is diagnostic for the marginal-versus-conditional divide identified above, with PCM and RCIT being the strongest CIT methods. Scenario~3 is where $^{\dagger}$MMD-$q$ and $^{\dagger}$MMD-$b$ achieve the highest DRT power, RCIT and PCM lead among CIT methods, and CMMD and CGED match the top CIT methods. As documented in \supple, CMMD and CGED incur substantially higher computational cost, which limits their scalability.

\subsection{Real Data Analysis} \label{Section: Real Data Analysis}
\begin{figure}[t]
\centering
\includegraphics[width=1\textwidth]{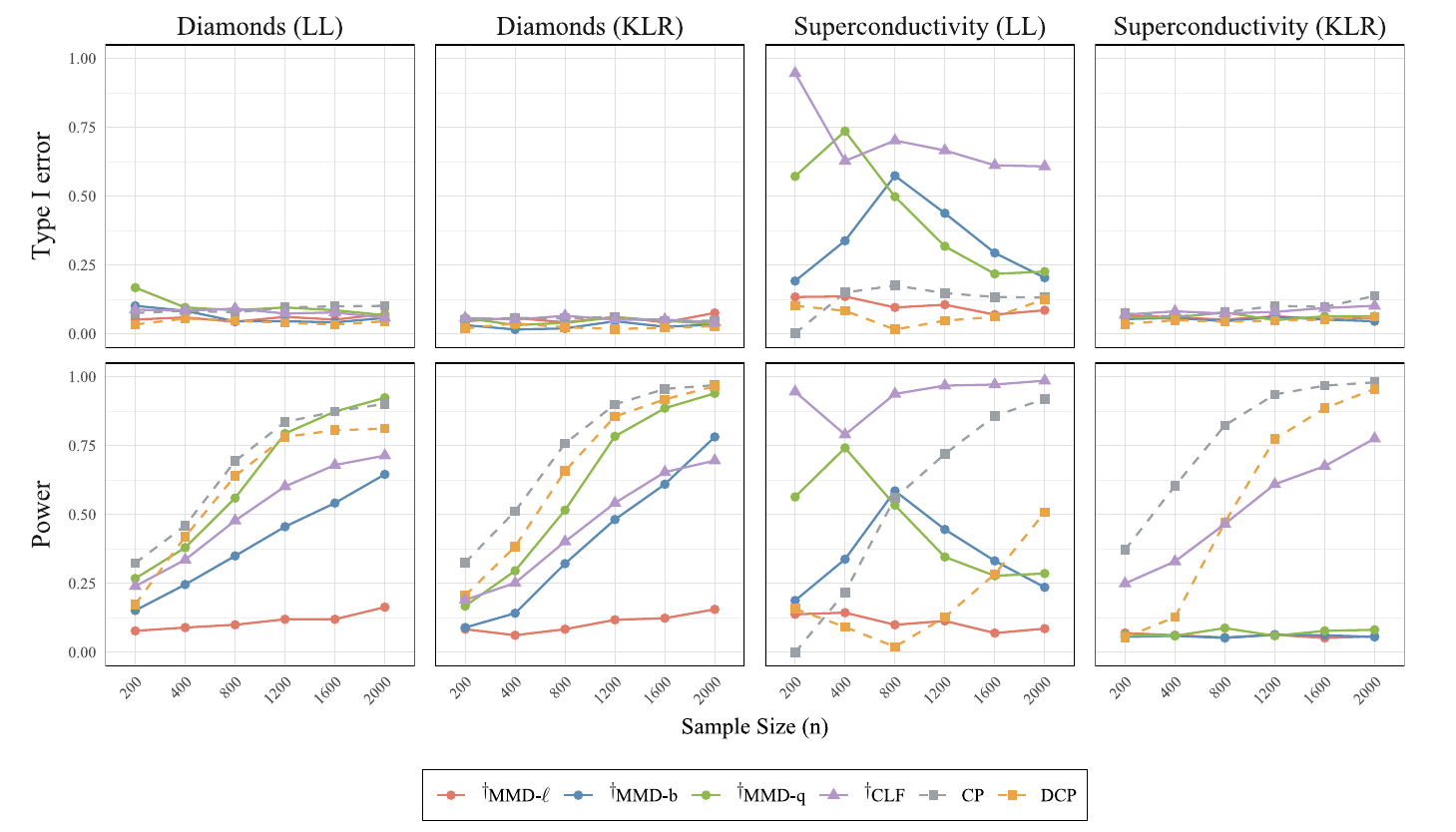}
\caption{Performance comparison of DRT methods on diamonds and superconductivity datasets using LLR and KLR for density ratio estimation. Rejection rates are averaged over 500 repetitions with $\alpha = 0.05$, under null (\emph{top}) and alternative (\emph{bottom}) hypotheses.}
\label{fig:real DRT}
\end{figure}

We evaluate the proposed methods on two real-world datasets, using the resampling procedure described in \Cref{subsec: Experimental Setup}. Figure~\ref{fig:real DRT} summarizes the performance of DRT methods under both LLR and KLR for density ratio estimation. 

\vspace{0.3em}
\mypara{Diamonds dataset} The diamonds dataset, available in the \textsf{R} package \texttt{ggplot2}, consists of $53{,}490$ observations and 10 features, including price, carat, clarity and color. In our analysis, we set price as the response~$Y$, and use the 6 numerical features \texttt{(carat, depth, table, x, y, z)} as covariates~$X$. Most DRT methods exhibit good type~I error control under both LLR and KLR, with rejection rates close to the nominal level~$\alpha$. Under the alternative, power increases steadily with sample size for all methods.

\vspace{0.3em}
\mypara{Superconductivity dataset}
The superconductivity dataset \citep{Hamidieh2018}, obtained from the UCI Machine Learning Repository, comprises $21{,}263$ superconductors described by $81$ features, with critical temperature as the response~$Y$. This high-dimensional setting poses a more challenging density ratio estimation problem. The results reveal a significant contrast between LLR and KLR. Under LLR, several DRT methods, particularly the classifier-based tests, fail to control the type~I error, with rejection rates exceeding~$\alpha$. Under KLR, the type~I error control improves considerably, illustrating the sensitivity of DRT methods to the choice of estimation procedure in high dimensions. 

These findings illustrate that the choice of density ratio estimator affects the validity of DRT methods in high-dimensional settings, with the selection of LLR or KLR depending on data dimensionality and the reliability of density ratio estimation. Experimental results for CIT methods on these datasets are reported in Appendix~\ref{Appendix: CIT Real Data Results}, and a more detailed empirical analysis of the trade-offs among computational cost, power, and type~I error control across all method variants is provided in Appendix~\ref{Appendix: practitioners_guide}. 

\section{Conclusion} \label{Section: Conclusion}
In this paper, we shed new light on the relatively underexplored problem of conditional two-sample testing. We begin by characterizing the fundamental difficulty of the problem and highlighting the importance of assumptions to make it feasible. We then introduce two general frameworks: (1) converting conditional independence tests into conditional two-sample tests and (2) transforming the problem of comparing conditional distributions into a problem of comparing marginal distributions based on density ratio estimation. Both approaches offer significant flexibility, allowing one to leverage well-developed tools to effectively tackle the problem. 

Our work opens up several interesting directions for future research. One promising avenue is to extend our framework to conditional 
$K$-sample testing with a general $
K \geq 2$. Such an extension would expand the applicability of our framework beyond the comparison of just two groups. This setting is related to conditional independence testing where $Z$ is a categorical random variable taking values in $\{1,2,\ldots,K\}$. We expect our results established in \Cref{Section: Approach via Conditional Independence Testing} to serve as a cornerstone for this extension. Another direction worth exploring is establishing a framework for conditional two-sample testing based on resampling methods. One promising approach is the Sampling Importance Resampling (SIR) algorithm \citep[][Chapter 6.3]{GiveHoet12}, which allows us to obtain an approximate sample from the distribution with density 
$f_{XY}$. Future work can focus on methods that compare the sample from $P_{XY}^{(1)}$ with the approximate sample obtained from the SIR algorithm. We leave these interesting topics for future investigation.

\mypara{Acknowledgement}
We thank the editor, associate editor, and anonymous reviewers for their constructive feedback, which has significantly improved the quality of this paper. We acknowledge support from the Basic Science Research Program through the National Research Foundation of Korea (NRF) funded by the Ministry of Education (2022R1A4A1033384), and the Korea government (MSIT) RS-2023-00211073. We are grateful to Gyumin Lee for the careful proofreading.

\bibliographystyle{apalike}
\bibliography{reference}

\newpage
\appendix
\mypara{Overview}
\Cref{section: Proofs} presents the proofs of our main theoretical results.
\Cref{Section: Supporting Lemmas} collects the supporting lemmas invoked in these proofs.
\Cref{sec:extended_literature_technical} provides an extended literature review together with the detailed formulations of importance-weighted statistics that are summarized in \main.
Finally, \Cref{section: Additional Experiments} reports implementation details and additional simulation results.

\mypara{Notation} 
For real sequences $(a_n)_{n \geq 1}$ and $(b_n)_{n \geq 1}$ with $b_n > 0$, we write $a_n \lesssim b_n$ if there exists a constant $C > 0$ such that $|a_n| \leq C b_n$ for all $n \in \mathbb{N}$. 
Given a family $\mathcal{P}$ of distributions, let $(X_{P,n})_{n \in \mathbb{N},\, P \in \mathcal{P}}$ denote a family of sequences of random variables indexed by $P \in \mathcal{P}$. 
We write $X_{P,n} = o_{\mathcal{P}}(n^{-a})$ if for every $\epsilon > 0$,
\begin{align*}
\sup_{P \in \mathcal{P}} 
\mP_P \bigl(n^a|X_{P,n}| > \epsilon\bigr) \to 0 
\quad \text{as } n \to \infty.
\end{align*}
For a positive integer $n$, we write $[n] := \{1,\ldots,n\}$.

\section{Proofs}\label{section: Proofs}
\subsection{\texorpdfstring{Proof of Theorem~\ref{Theorem: negative result}}{Proof of Theorem 1}} \label{Section: Proof of Hardness Result}

\begin{proof}
To formalize the connection between the conditional two-sample problem and conditional independence testing (CIT), we first construct an auxiliary joint distribution. For any given distribution pair $P = (P_{XY}^{(1)}, P_{XY}^{(2)}) \in \mathcal{P}_{M}^{\mathsf{ac}}$, we define a joint distribution $P_{XYZ}$ on $\mathbb{R}^{d_X} \times \mathbb{R}^{d_Y} \times \{1, 2\}$ as follows
\begin{align}\label{eq: joint pdf auxiliary}
    P_{XYZ} \coloneqq \frac{n_1}{n} \big(P_{XY}^{(1)} \otimes \delta_{\{Z=1\}}\big) + \frac{n_2}{n} \big(P_{XY}^{(2)} \otimes \delta_{\{Z=2\}}\big),
\end{align}
where $n \coloneqq n_1+n_2$ and $\delta_{\{Z=j\}}$ denotes the Dirac measure at $Z=j$ for $j \in \{1,2\}$.

Let $\mathcal{Q}_{M}^{\mathsf{ac}}$ denote the class of all such joint distributions induced by \eqref{eq: joint pdf auxiliary} for any $P \in \mathcal{P}_{M}^{\mathsf{ac}}$. Within this induced space, we define the CIT null class as $\mathcal{Q}_{0,M}^{\mathsf{ac}} \coloneqq \{P_{XYZ} \in \mathcal{Q}_{M}^{\mathsf{ac}} : Y \independent Z \mid X\}$ and the alternative class as $\mathcal{Q}_{1,M}^{\mathsf{ac}} \coloneqq \{P_{XYZ} \in \mathcal{Q}_{M}^{\mathsf{ac}} : Y \nindep Z \mid X\}$.

Under the distribution \eqref{eq: joint pdf auxiliary}, the condition $P_{Y|X}^{(1)} = P_{Y|X}^{(2)}$ 
holds almost surely if and only if $P(Y \mid X, Z=1) = P(Y \mid X, Z=2)$ holds almost surely, which corresponds to $Y \independent Z \mid X$. Therefore, the two-sample null class $\mathcal{P}_{0,M}^{\mathsf{ac}}$ maps to the CIT null class $\mathcal{Q}_{0,M}^{\mathsf{ac}}$, and similarly, the alternative class $\mathcal{P}_{1,M}^{\mathsf{ac}}$ maps to $\mathcal{Q}_{1,M}^{\mathsf{ac}}$.

Because the sample sizes $n_1$ and $n_2$ are deterministic in our setting, we cannot directly apply the i.i.d.~hardness result of \citet{shah2020hardness}. To bridge this gap, we embed the test $\phi$ from Theorem~\ref{Theorem: negative result} (denoted as $\phi_{\text{two}}$ hereafter) into an i.i.d.~sampling scheme. For an arbitrary integer $N \ge 2n$, suppose we draw $\{(X_i,Y_i,Z_i)\}_{i=1}^N \iid P_{XYZ}$. Let $N_1' \coloneqq \sum_{i=1}^N \mathds{1}(Z_i = 1)$ and $N_2' \coloneqq \sum_{i=1}^N \mathds{1}(Z_i = 2)$, and define the event $\mathcal{A}\coloneqq \{N_1' \geq n_1,\ N_2' \geq n_2\}$.

Writing $\lambda_n \coloneqq n_1/n$, we have $N_1' \sim \mathrm{Binomial}(N,\lambda_n)$ and $N_2' \sim \mathrm{Binomial}(N,1-\lambda_n)$.
Since $N \geq 2n$ implies $N\lambda_n \geq 2n_1$, we have $n_1 - N\lambda_n \leq -\tfrac{1}{2}N\lambda_n$, and applying Chebyshev's inequality yields:
\begin{align*}
    \mP(N_1' < n_1) & =  \mP(N_1' - N\lambda_n < n_1 - N\lambda_n)  \leq \mP\Bigl( N_1' - N\lambda_n \le - \frac{1}{2} N \lambda_n \Bigr) \\
    & \leq \mP\Bigl( \big|N_1' - N\lambda_n \big| \geq  \frac{1}{2} N \lambda_n \Bigr) \leq \frac{4(1 - \lambda_n)}{N\lambda_n}.
\end{align*}
Similarly, $\mP(N_2' < n_2) \leq \tfrac{4\lambda_n}{N(1-\lambda_n)}$. By the union bound,
\begin{align}
    \label{Eq: lower bound for the prob A}
    \mP(\mathcal{A}) \geq 1 - \frac{4(1-\lambda_n)^2 + 4\lambda_n^2}{N\lambda_n(1-\lambda_n)} \overset{\mathrm{set}}{=} 1 -  \varepsilon_N,
\end{align}
where $\varepsilon_N \rightarrow 0$ as $N \rightarrow \infty$.

We now construct a conditional independence test $\phi_{\text{indep}}^{(N)}$ based on these $N$ i.i.d.~samples. Whenever $\mathds{1}(\mathcal{A})=1$, this test extracts $\mathcal{D}_1$ as exactly the first $n_1$ observations from the subset $\{(X_i, Y_i): Z_i=1\}$, and $\mathcal{D}_2$ as the first $n_2$ observations from $\{(X_i, Y_i): Z_i=2\}$. It then sets $\phi_{\text{indep}}^{(N)} \coloneqq \phi_{\text{two}}(\mathcal{D}_1, \mathcal{D}_2)$. If $\mathds{1}(\mathcal{A})=0$, the test simply returns 0 (i.e., accepts $H_0$).

Conditional on $\mathcal{A}$, selecting the first $n_1$ and $n_2$ observations from each respective group preserves the exact joint distribution of the original two independent samples. Thus, $\sup _{P \in \mathcal{P}^{\mathsf{ac}}_{0,M}} \mE_P(\phi_{\text{indep}}^{(N)} \mid \mathcal{A}) = \sup _{P \in \mathcal{P}^{\mathsf{ac}}_{0,M}} \mE_P(\phi_{\text{two}}) \leq \alpha$.
The unconditional type~I error of $\phi_{\text{indep}}^{(N)}$ then satisfies
\begin{align*}
    \sup_{P \in \mathcal{P}^{\mathsf{ac}}_{0,M}} \mE_P(\phi_{\text{indep}}^{(N)}) = \sup_{P \in \mathcal{P}^{\mathsf{ac}}_{0,M}} \mE_P(\phi_{\text{indep}}^{(N)} \mathds{1}(\mathcal{A})) \leq \sup_{P \in \mathcal{P}^{\mathsf{ac}}_{0,M}} \mE_P(\phi_{\text{indep}}^{(N)} \mid \mathcal{A}) \leq \alpha,
\end{align*}
making $\phi_{\text{indep}}^{(N)}$ a valid test for conditional independence with size $\alpha$. 

Furthermore, for any $P \in \mathcal{P}^{\mathsf{ac}}_{1,M}$, we can bound the expected power as follows
\begin{align*}
    & \mE_P(\phi_{\text{indep}}^{(N)} \mid \mathcal{A})(1-\varepsilon_N) \stackrel{(\mathrm{i})}{\leq} \mE_P(\phi_{\text{indep}}^{(N)} \mid \mathcal{A}) \mP(\mathcal{A}) = \mE_P(\phi_{\text{indep}}^{(N)}) \stackrel{(\mathrm{ii})}{\leq} \alpha \\
    \Longleftrightarrow \quad & \mE_P(\phi_{\text{two}}) = \mE_P(\phi_{\text{indep}}^{(N)} \mid \mathcal{A}) \leq \frac{\alpha}{1-\varepsilon_N},
\end{align*}
where step~(i) uses the inequality \eqref{Eq: lower bound for the prob A}, step~(ii) holds by the hardness of CIT \citep[Theorem~2 and Remark~4]{shah2020hardness}, and the equivalence $\mE_P(\phi_{\text{two}}) = \mE_P(\phi_{\text{indep}}^{(N)} \mid \mathcal{A})$ holds due to the distributional equivalence of the subsampled data conditional on $\mathcal{A}$.

Since $N$ is an arbitrary integer satisfying $N \geq 2n$ and $\varepsilon_N \rightarrow 0$ as $N \rightarrow \infty$, taking the limit yields $\mE_P(\phi_{\text{two}}) \leq \alpha$ for any $P \in \mathcal{P}^{\mathsf{ac}}_{1,M}$.
\end{proof}

\subsection{\texorpdfstring{Proof of Theorem~\ref{Theorem: Converting C2ST into CIT}}{Proof of Theorem 2}} \label{Section: Proof of Converting C2ST into CIT}

\begin{proof}
We may write $\phi = \mathds{1}(\tilde{n}_1 \leq n_1)\mathds{1}(\tilde{n}_2 \leq n_2)\psi_{\tilde{n}}$, where $\psi_{\tilde{n}}$ denotes the output of $\psi$ on the dataset $\mathcal{D}_{\tilde n} = \{(X_i,Y_i,Z_i)\}_{i=1}^{\tilde{n}}$ constructed in Algorithm~\ref{Algorithm: Converting CIT into C2ST}. Now generate new i.i.d.\ samples $\tilde{\mathcal{D}}_{\tilde n} = \{(\tilde X_i, \tilde Y_i, \tilde Z_i)\}_{i=1}^{\tilde n} \iid P_{XYZ}$ independent of $\mathcal{D}_{\tilde{n}}$ and define
\begin{align*}
    \tilde{\psi}^\dagger = \mathds{1}\biggl\{\sum_{i=1}^{\tilde{n}} \mathds{1}(\tilde{Z}_i=1) \leq n_1 \biggr\} \mathds{1}\biggl\{\sum_{i=1}^{\tilde{n}} \mathds{1}(\tilde{Z}_i=2) \leq n_2 \biggr\} \psi_{\tilde{n}}^\dagger,
\end{align*}
where $\psi_{\tilde{n}}^\dagger$ is defined as $\psi$ but based on $\tilde{\mathcal{D}}_{\tilde{n}}$. Since Algorithm~\ref{Algorithm: Converting CIT into C2ST} draws $\tilde n_1 \sim \mathrm{Binomial}(\tilde n, n_1/n)$, uniformly subsamples $\tilde n_1$ and $\tilde n_2$ observations from the two groups, and applies a random permutation, the joint distribution of $(\tilde{n}_1, \mathcal{D}_{\tilde{n}})$ is identical to that of $(\tilde{N}_1, \tilde{\mathcal{D}}_{\tilde{n}})$, where $\tilde{N}_j = \sum_{i=1}^{\tilde{n}} \mathds{1}(\tilde{Z}_i = j)$. Thus $\phi \overset{d}{=} \tilde{\psi}^\dagger$, and it follows that
\begin{align*}
    \mE_P(\phi) = \mE_P(\tilde{\psi}^\dagger) \leq \mE_P(\psi_{\tilde n}^\dagger)
\end{align*}
for all $P$. Therefore, the first claim on type~I error control follows.

Moving to the type~II error, observe that 
\begin{align*}
    \phi = \mathds{1}(\tilde{n}_1 \leq n_1) \mathds{1}(\tilde{n}_2 \leq n_2)\psi_{\tilde{n}} 
    &= \psi_{\tilde{n}} - \mathds{1}(\tilde{n}_1 > n_1 \text{ or } \tilde{n}_2 > n_2)\psi_{\tilde{n}} \\&\geq \psi_{\tilde{n}} - \mathds{1}(\tilde{n}_1 > n_1) - \mathds{1}(\tilde{n}_2 > n_2),
\end{align*}
by the union bound, which yields
\begin{align} \label{Eq: type II error bound}
    \mE_P(1 - \phi) = \mE_{P}(1-\tilde{\psi}^\dagger) \leq \mE_P(1 - \psi_{\tilde{n}}^\dagger)  + \mP(\tilde{N}_1 > n_1) + \mP(\tilde{N}_2 > n_2).
\end{align}
Since $\tilde{N}_1 \sim \mathrm{Binomial}(\tilde{n}, n_1/n)$, applying \Cref{Lemma: Concentration inequality} with $(1+\delta)\tilde{n} \cdot n_1/n = n_1$ gives $\delta = k^{-1} -1$. 
Thus, by \Cref{Lemma: Concentration inequality}, we have
\begin{align*}
    \mP(\tilde{N}_1 > n_1) \leq \exp\biggl( - \frac{n_1 k(k^{-1}-1)^2}{3} \biggr) \stackrel{(\mathrm{i})}{\leq} \exp\biggl( - \frac{n_{\min} k(k^{-1}-1)^2}{3} \biggr),
\end{align*}
where $(\mathrm{i})$ follows from $n_{\min} \leq n_1$. Letting the right-hand side equal $\varepsilon$ and solving for $k\in(0,1)$, we derive the form of $k^\ast$ as presented in Algorithm~\ref{Algorithm: Converting CIT into C2ST}. To verify $\delta \in [0,1]$, note that $n_{\min} \geq 6\log(1/\varepsilon)$ implies $k^\ast \geq 1/2$. By symmetry, $\mP(\tilde{N}_2 > n_2) \leq \varepsilon$. As a result, continuing from inequality~\eqref{Eq: type II error bound},
\begin{align*}
    \mE_P(1 - \phi) \leq \mE_P(1 - \psi_{\tilde{n}}^\dagger) + 2\varepsilon,
\end{align*}	
for all $n$ satisfying $n_{\min} \geq 6\log(1/\varepsilon)$.

Recall from the theorem condition that $\lambda_n = n_1/n$ is bounded away from $0$ and $1$, meaning $c \leq \lambda_n \leq 1-c$ for some $c \in (0, 1/2)$. This condition ensures that $n_{\min} \geq cn$, which guarantees $n_{\min} \to \infty$ as $n \to \infty$. Coupled with the condition $\lim_{n \to \infty} \frac{\log(1/\varepsilon)}{n_{\min}} = 0$, this implies that $k^\ast \to 1$, and consequently, $\tilde{n} = \lfloor k^\ast n \rfloor \to \infty$ as $n \to \infty$. Since $\varepsilon = o(1)$ and $\tilde{n} \to \infty$, the asymptotic assumptions on $\psi$ give 
\begin{align*}
    \limsup_{n\to\infty} \sup_{P\in\mathcal{P}_0^{\mathsf{cit}}} \mE_P(\psi_{\tilde n}^\dagger) \leq \alpha 
    \quad \text{and} \quad 
    \lim_{n\to\infty} \sup_{P\in\mathcal{P}_1^{\mathsf{cit}}} \mE_P(1-\psi_{\tilde n}^\dagger) = 0.
\end{align*}
Combining with the above results and $\varepsilon = o(1)$,
\begin{align*}
    \limsup_{n\to\infty}\sup_{P\in\mathcal{P}_0^{\mathsf{cit}}}\mE_P(\phi) \le \alpha, \qquad
    \lim_{n\to\infty}\sup_{P\in\mathcal{P}_1^{\mathsf{cit}}}\mE_P(1-\phi) = 0.
\end{align*}
This completes the proof of Theorem~\ref{Theorem: Converting C2ST into CIT}.
\end{proof}

\subsection{\texorpdfstring{Proof of Theorem~\ref{Theorem: Asymptotic Normality of Classification Accuracy}}{Proof of Theorem 3}} 
\label{Section: Proof: Asymptotic Normality of Classification Accuracy}
\begin{proof}
We analyze the numerator and the denominator of $\widehat{\mathrm{Acc}}$ separately. We first show that the numerator converges to a Gaussian distribution and then establish that the denominator is a ratio-consistent estimator of the population-level standard deviation under Assumption~\ref{Assumption: classifier}.  

\mypara{$\mathrm{(i)}$ Analysis of the numerator} 
Let us rewrite 
\begin{align*}
    \overline{A}_1 + \overline{A}_2 - 1  & = \frac{1}{m} \sum_{i=1}^m \bigl[ \mathds{1}\{\widehat{h}(V_i^{(1)})=1\} + r_X(X_i^{(2)})\mathds{1}\{\widehat{h}(V_i^{(2)})=2\} - 1\bigr] \\
    & \hskip 5em + \frac{1}{m} \sum_{i=1}^m  \bigl\{ \widehat{r}_X(X_i^{(2)}) - r_X(X_i^{(2)}) \bigr\} \mathds{1}\{\widehat{h}(V_i^{(2)})=2\} \\
    & = \frac{1}{m} \sum_{i=1}^m \underbrace{\bigl[ \mathds{1}\{\widehat{h}(V_i^{(1)})=1\} + r_X(X_i^{(2)})\mathds{1}\{\widehat{h}(V_i^{(2)})=2\} - 1\bigr]}_{\coloneqq L_i(\widehat{h})} + o_{\mathcal{P}^{\mathsf{clf}}_0}(m^{-1/2}),
\end{align*}
where the last approximation holds since 
\begin{align*}
    \bigg|\frac{1}{m} \sum_{i=1}^m \bigl\{ \widehat{r}_X(X_i^{(2)}) - r_X(X_i^{(2)}) \bigr\} \mathds{1}\{\widehat{h}(V_i^{(2)})=2\} \bigg| 
    \leq \sqrt{\frac{1}{m} \sum_{i=1}^m \bigl\{ \widehat{r}_X(X_i^{(2)}) - r_X(X_i^{(2)}) \bigr\}^2},
\end{align*}
and the upper bound is 
$o_{\mathcal{P}_0^{\mathsf{clf}}}(m^{-1/2})$
due to Assumption~\ref{Assumption: classifier}(b). Thus $\overline{A}_1 + \overline{A}_2 - 1$ is dominated by the average of $L_i(\widehat{h})$. 
It therefore suffices to study the limiting distribution of 
this average.
Under Assumption~\ref{Assumption: classifier}(a), the conditional central limit theorem (\Cref{Lemma: conditional clt}) yields that 
\begin{align*}
    \lim_{n \rightarrow \infty}\sup_{P \in \mathcal{P}_0^{\mathsf{clf}}} \sup_{t \in \mathbb{R}} \Bigg| \mP_P\Biggl( \frac{m^{-1/2}\sum_{i=1}^m L_i(\widehat{h})}{\sqrt{\{\mV[L(\widehat{h}) \given \widehat{h}]\}}} \leq t \Biggr) - \Phi(t) \Bigg| = 0.
\end{align*}

\mypara{$\mathrm{(ii)}$ Consistency of the variance estimate} We next show the ratio consistency of the variance estimator. Observe that
\begin{align*}
    \mV[L(\widehat{h}) \given \widehat{h}] = \underbrace{\mV[ \mathds{1}\{\widehat{h}(V_i^{(1)})=1\} \given \widehat{h}]}_{\coloneqq  \sigma_1^2} + \underbrace{\mV\bigl[ r_X(X_i^{(2)})\mathds{1}\{\widehat{h}(V_i^{(2)})=2\} \given \widehat{h}\bigr]}_{\coloneqq  \sigma_2^2},
\end{align*}
and 
\begin{align*}
    \bigg|\frac{\widehat{\sigma}_1^2+ \widehat{\sigma}_2^2}{\sigma_1^2 + \sigma_2^2} - 1 \bigg| \leq \bigg|\frac{\widehat{\sigma}_1^2 - \sigma_1^2}{\sigma_1^2} \bigg| + \bigg|\frac{\widehat{\sigma}_2^2 - \sigma_2^2}{\sigma_2^2} \bigg|. 
\end{align*}
Therefore, the ratio consistency of $\widehat{\sigma}_1^2 + \widehat{\sigma}_2^2$ reduces to that of $\widehat{\sigma}_1^2$ and $\widehat{\sigma}_2^2$ individually.
For $\widehat{\sigma}_1^2$, conditional Chebyshev's inequality gives
\begin{align*}
    \sup_{P \in \mathcal{P}_0^{\mathsf{clf}}}\mP_P\bigl( |\widehat{\sigma}_1^2/\sigma_1^2 - 1| \geq t \given \widehat{h} \bigr) \leq \frac{1}{t^2} \sup_{P \in \mathcal{P}_0^{\mathsf{clf}}}\mV_P(\widehat{\sigma}_1^2/\sigma_1^2 \given \widehat{h}) \leq \frac{1}{t^2\underline{c}_1(1-\overline{c}_{1})m},
\end{align*}
for sufficiently large $n$ and for all $t > 0$, under Assumption~\ref{Assumption: classifier}(a). Hence $\widehat{\sigma}_1^2/\sigma_1^2$ converges to one in probability uniformly over 
$\mathcal{P}_0^{\mathsf{clf}}$. 

For $\widehat{\sigma}_2^2$, the analysis requires a finer decomposition. Letting $A_{2,i} := r_X(X_i^{(2)}) \mathds{1} \{\widehat{h}(V_i^{(2)}) = 2\}$, we write
\begin{align*}
    \widehat{\sigma}_2^2 & = \frac{1}{m-1}\sum_{i=1}^m \biggl[ A_{2,i} - \frac{1}{m} \sum_{j=1}^m A_{2,j} \biggr]^2 \\
    & + \frac{1}{m-1}\sum_{i=1}^m \biggl[ \bigl(\widehat{A}_{2,i} - A_{2,i}\bigr) - \frac{1}{m} \sum_{j=1}^m \bigl( \widehat{A}_{2,j} -  A_{2,j}\big) \biggr]^2\\
    & + \frac{2}{m-1}\sum_{i=1}^m\biggl[ A_{2,i} - \frac{1}{m} \sum_{j=1}^m A_{2,j} \biggr] \cdot \biggl[ \bigl(\widehat{A}_{2,i} - A_{2,i}\bigr) - \frac{1}{m} \sum_{j=1}^m \bigl( \widehat{A}_{2,j} -  A_{2,j}\big) \biggr] \\
    & \eqqcolon  (\mathbb{I}) + (\mathbb{II}) + (\mathbb{III}). 
\end{align*}
As before, the term $(\mathbb{I})/\sigma_2^2$ converges to one in probability uniformly over $\mathcal{P}_0^{\mathsf{clf}}$ under Assumption~\ref{Assumption: classifier}(a). 
The term $(\mathbb{II})/\sigma_2^2$ is $o_{\mathcal{P}_0^{\mathsf{clf}}}(1)$ by Markov's inequality combined with Assumption~\ref{Assumption: classifier}(b). Lastly, 
$|(\mathbb{III})| \leq 2\sqrt{(\mathbb{I}) \times (\mathbb{II})}$, 
which is again $o_{\mathcal{P}_0^{\mathsf{clf}}}(1)$. Therefore, $\widehat{\sigma}_2^2/\sigma_2^2$ converges to one in probability uniformly over $\mathcal{P}_0^{\mathsf{clf}}$,
which
proves that $\sqrt{\big\{(\sigma_1^2 + \sigma_2^2)/(\widehat{\sigma}_1^2 + \widehat{\sigma}_2^2)\big\}} = 1 + o_{\mathcal{P}^{\mathsf{clf}}_0}(1)$ by \citet[][Lemma S7]{lundborg2024projected}.

Putting all pieces together with \Cref{Lemma: uniform Slutsky}(b) proves the claim. 
\end{proof}

\subsection{\texorpdfstring{Proof of Corollary~\ref{Corollary: Asymptotic Normality of CV Accuracy}}{Proof of Corollary 1}}
\begin{proof}
For each $j \in \{1,\ldots,K\}$, the variance analysis in the
proof of Theorem~\ref{Theorem: Asymptotic Normality of Classification Accuracy} shows that 
$\{(\sigma_1^2 + \sigma_2^2)/
(\widehat{\sigma}_{1,j}^2 + 
\widehat{\sigma}_{2,j}^2)\}^{1/2} = 
1 + o_{\mathcal{P}_0^{\mathsf{clf}}}(1)$. 
Thus, by \Cref{Lemma: uniform Slutsky}(b), it is enough to show the asymptotic normality of  
\begin{align*}
    \frac{1}{\sqrt{K}}\sum_{j = 1}^K \frac{\sqrt{m}(\overline{A}_{1,j} + \overline{A}_{2,j} - 1)}{\sqrt{(\sigma_{1}^2 + \sigma_2^2)}}.
\end{align*}
Without loss of generality, denote the index sets of the folds
$D_{\mathsf{infer},1}, \ldots, D_{\mathsf{infer},K}$ as
\begin{align*}
    I_1 = \bigl\{1,\ldots,m\bigr\}, \ I_2 = \bigl\{m+1,\ldots,\ 2m \bigr\},\ldots, I_K =\bigl\{ m(K-1) + 1,\ldots, mK\bigr\}.
\end{align*}
Then the proof of Theorem~\ref{Theorem: Asymptotic Normality of Classification Accuracy} establishes that 
\begin{align*}
    \frac{1}{\sqrt{K}}\sum_{j = 1}^K \frac{\sqrt{m}(\overline{A}_{1,j} + \overline{A}_{2,j} - 1)}{\sqrt{(\sigma_{1}^2 + \sigma_2^2)}} = {\frac{1}{\sqrt{(mK)}}} \sum_{j=1}^K \underbrace{\Biggl\{\sum_{i \in I_j} \frac{A_{1,i} + A_{2,i} - 1}{\sqrt{(\sigma_{1}^2 + \sigma_2^2)}} \Biggr\}}_{\coloneqq  B_j} + o_{\mathcal{P}^{\mathsf{clf}}_0}(1),
\end{align*}
where $A_{1,i} + A_{2,i} - 1 \coloneqq  \mathds{1}\{\widehat{h}(V_i^{(1)}) = 1\} + r_X(X_i^{(2)}) \mathds{1}\{\widehat{h}(V_i^{(2)})= 2\} - 1$. Note that $B_1,\ldots,B_K$ are mutually independent conditional on $\widehat{h}$. As in the proof of Theorem~\ref{Theorem: Asymptotic Normality of Classification Accuracy}, we apply the conditional central limit theorem (\Cref{Lemma: conditional clt}) to the average of $B_1,\ldots,B_K$ conditional on $\widehat{h}$ under Assumption~\ref{Assumption: classifier}, which completes the proof. 
\end{proof}

\subsection{\texorpdfstring{Proof of Theorem~\ref{theorem: MMD Asymptotic Distribution}}{Proof of Theorem 4}}\label{Proof of theorem: MMD Asymptotic Distribution}
\begin{proof}
The proof proceeds in four steps. First, we derive a block-level decomposition of the statistic into four components. Second, we show that three remainder terms are asymptotically negligible, uniformly over $0 \leq \gamma \leq 1$. Third, we analyze the leading term, distinguishing between the regimes $\gamma=1$ and $0 \le \gamma < 1$. Finally, for $0 \le \gamma < 1$, we establish the consistency of the variance estimator and conclude by an application of Lemma~\ref{Lemma: uniform Slutsky}(b).

The remainder terms arising from density ratio estimation are controlled by Assumption~\ref{Assumption : Unified MMD}(b), which requires $\mE(\{\widehat r_X(X)-r_X(X)\}^2)
=o((n-m)^{-1/2})$, an $o((n-m)^{-1/4})$ root-mean-square error rate. To see how this compares with existing work, note that \citet{hu2024two} impose the condition
\begin{align*}
    \mE\big(\{\widehat G_{11}-G_{11}\}^2 \mid D_{\mathsf{ratio}}\big)=o_P(m^{-1}),
\end{align*}
where $\widehat G_{11}$ and $G_{11}$ denote the plug-in and population kernel matrices in their framework. In squared-error scale, this effectively requires $(n-m)^{-1}$ accuracy, whereas our condition requires only $(n-m)^{-1/2}$.

\mypara{$\mathrm{(i)}$ Block-Level Decomposition}
All inner products and norms below are taken 
in
the reproducing kernel Hilbert space $\mathcal{H}_k$, but we omit explicit reference to $\mathcal{H}_k$ for notational simplicity. Using the feature map representation $k(x,y)=\langle\psi(x),\psi(y)\rangle$, the block-averaged statistic $\widehat{\mathrm{GB}}_\gamma$ can be written as
\begin{align*}
    \widehat{\mathrm{GB}}_\gamma
    = \frac{1}{S} \sum_{b=1}^{S} \underbrace{\frac{B}{\binom{B}{2}}
        \sum_{\substack{i,j\in\mathcal I_b\\ i< j}}
        \big\langle \psi(V_i^{(1)}) - \widehat r_X(X_i^{(2)})\,\psi(V_i^{(2)}),\;
        \psi(V_j^{(1)}) - \widehat r_X(X_j^{(2)})\,\psi(V_j^{(2)}) \big\rangle}_{=:\hat{\eta}_b}.
\end{align*}
Letting $V_{B,(b)}$ denote the Hilbert-space norm of the difference between the block-wise mean embeddings of the two samples,
\begin{align*}
    V_{B,(b)}
    := \bigg\|
    \frac{1}{B}\sum_{i\in \mathcal I_b}\psi(V_i^{(1)})
    -
    \frac{1}{B}\sum_{i\in \mathcal I_b}
    \widehat r_X(X_i^{(2)})\,\psi(V_i^{(2)})
    \bigg\|,
\end{align*}
each block contribution $\hat{\eta}_b$ can be expressed as
\begin{align*}
    \hat{\eta}_{b} = \bigg(\underbrace{B\,V_{B,(b)}^2 - \frac{1}{B}\sum_{i\in\mathcal{I}_b}\widehat{H}_{ii}}_{=:T_b}\bigg)\cdot\frac{B}{B-1},
\end{align*}
so that $\widehat{\mathrm{GB}}_\gamma = S^{-1}\sum_{b=1}^{S}\hat{\eta}_{b}$. We decompose each block term $T_b$ into a population part and an estimation-error part by writing $\sqrt{B}\,V_{B,(b)} = \|\I_b + \II_b\|$, where $\I_b$ and $\II_b$ denote the population and estimation-error components, respectively:
\begin{align*}
    \I_{b} &:= \frac{1}{\sqrt{B}}\sum_{i\in\mathcal{I}_b}\bigl\{\psi(V_i^{(1)}) - r_X(X_i^{(2)})\,\psi(V_i^{(2)})\bigr\}, \\
    \II_{b} &:= \frac{1}{\sqrt{B}}\sum_{i\in\mathcal{I}_b}\psi(V_i^{(2)})\{r_X(X_i^{(2)})-\widehat r_X(X_i^{(2)})\}.
\end{align*}

Expanding the squared norm yields
$B\,V_{B,(b)}^2 = \|\I_{b}\|^2 + \|\II_{b}\|^2
+ 2\langle \I_{b}, \II_{b}\rangle$.
By construction of $\widehat H$ and $H$, the 
diagonal correction can be written as 
\begin{align*}
    \frac{1}{B}\sum_{i\in\mathcal{I}_b}\widehat H_{ii}
    &= \frac{1}{B}\sum_{i\in\mathcal{I}_b} H_{ii} \\
    &\quad + \frac{2}{B}\sum_{i\in\mathcal{I}_b}\langle \psi(V_i^{(1)}) - r_X(X_i^{(2)})\,\psi(V_i^{(2)}), \psi(V_i^{(2)}) \rangle\cdot \bigl\{r_X(X_i^{(2)})-\widehat{r}_X(X_i^{(2)})\bigr\} \\
    &\quad + \frac{1}{B}\sum_{i\in\mathcal{I}_b}\langle\psi(V_{i}^{(2)}),\psi(V_{i}^{(2)})\rangle\cdot\bigl\{\widehat r_X(X_i^{(2)})-r_X(X_i^{(2)})\bigr\}^2 \, .
\end{align*}
Combining these expressions, the block-level statistic $T_b$ admits the four-term decomposition
\begin{align*}
    T_b
    &= T_{1,b} + T_{2,b} + T_{3,b} + T_{4,b} 
\end{align*}
where
\begin{align*}
    T_{1,b}
    &:= \|\I_b\|^2
    - \frac{1}{B}\sum_{i\in \mathcal I_b} H_{ii}, 
    \\[0.3em]
    T_{2,b}
    &:= \|\II_b\|^2, \\[0.3em]
    T_{3,b}
    &:= 2\langle \I_b,\II_b\rangle
    - \frac{2}{B}\sum_{i\in\mathcal{I}_b}
    \langle \psi(V_i^{(1)}) 
    - r_X(X_i^{(2)})\,\psi(V_i^{(2)}), 
    \psi(V_i^{(2)}) \rangle
    \cdot \bigl\{r_X(X_i^{(2)})
    -\widehat{r}_X(X_i^{(2)})\bigr\}, \\[0.3em]
    T_{4,b}
    &:= - \frac{1}{B}\sum_{i\in \mathcal I_b}
    \bigl\{\widehat{r}_X(X_i^{(2)})
    -r_X(X_i^{(2)})\bigr\}^2
    k(V_i^{(2)},V_i^{(2)}).
\end{align*}
Defining $\mathcal{T}_j := S^{-1/2}\sum_{b=1}^{S} T_{j,b}$ for $j=1,2,3,4$, the scaled statistic becomes
\begin{align*}
    \sqrt{S}\,\widehat{\mathrm{GB}}_\gamma=\frac{B}{B-1}\sum_{j=1}^{4} \mathcal{T}_j.  
\end{align*}
Since $B/(B-1)$ is a bounded deterministic factor, it does not affect asymptotic negligibility. We therefore first analyze the remainder terms $\mathcal{T}_2, \mathcal{T}_3, \mathcal{T}_4$ before returning to $\mathcal{T}_1$.

\vspace{0.5em}
\mypara{$\mathrm{(ii)}$ Control of Negligible Terms} We now show that the remainder terms
are asymptotically negligible. Specifically, we establish that, uniformly over $0 \leq \gamma \leq 1$,
\begin{align*}
    \mathcal{T}_2 = o_{\mathcal{P}_{0}^{\mathsf{mmd}}}(1), 
    \qquad 
    \mathcal{T}_3 = o_{\mathcal{P}_{0}^{\mathsf{mmd}}}(1), 
    \qquad 
    \mathcal{T}_4 = o_{\mathcal{P}_{0}^{\mathsf{mmd}}}(1).    
\end{align*}

\begin{enumerate}[label=(\alph*)]
    \item \textit{Term $\mathcal{T}_{2}$.}
    Recall that $\mathcal{T}_{2} = S^{-1/2}\sum_{b=1}^{S}\|\II_b\|^2$. For a fixed block $b\in\{1,\ldots,S\}$, writing $\Delta_i := r_X(X_i^{(2)})-\widehat r_X(X_i^{(2)})$, we have
    \begin{align*}
        \|\II_b\|^2=\frac{1}{B}\sum_{i\in\mathcal{I}_b}\sum_{j\in\mathcal{I}_b}k(V_i^{(2)},V_j^{(2)})\Delta_i\Delta_j\leq\frac{c_4}{B}
        \bigg(\sum_{i\in \mathcal{I}_b}|\Delta_i|\bigg)^2\leq c_4\sum_{i\in \mathcal{I}_b}\Delta_i^2,
    \end{align*}
    where the first inequality uses 
    $|k(x,y)| \leq c_4$ and the second follows from 
    the Cauchy--Schwarz inequality. Summing over $b=1,\dots,S$ yields
    \begin{align*}
        \mathcal{T}_2
        \leq
        \frac{c_4}{\sqrt{S}}
        \sum_{i=1}^{SB}
        \Delta_i^2.
    \end{align*}
    By Markov's inequality and Assumption~\ref{Assumption : Unified MMD}(b),
    \begin{align*}
        \mathcal{T}_2=o_{\mathcal{P}_{0}^{\mathsf{mmd}}}\big(B\,\sqrt{(S/(n-m))}\big)=o_{\mathcal{P}_{0}^{\mathsf{mmd}}}\big(\sqrt{(m^{1+\gamma}/(n-m))}\big),
    \end{align*}
    where the second equality uses $S=\lfloor m/B\rfloor$ and $B=\max\{2, \lfloor m^{\gamma} \rfloor\}$.
    Under Assumption~\ref{Assumption : Unified MMD}(d), 
    $m^{1+\gamma}/(n-m) \to c_5 \in [0,\infty)$, and thus $\mathcal{T}_2 = o_{\mathcal{P}_{0}^{\mathsf{mmd}}}(1).$
    
    \vspace{0.5em}
    \item \textit{Term $\mathcal{T}_{3}$.}  
    Subtracting the diagonal terms from the inner product $\langle \I_b, \II_b \rangle$ and writing $h_{ij} := \langle \psi(V_i^{(1)}) - r_X(X_i^{(2)})\,\psi(V_i^{(2)}), \psi(V_j^{(2)}) \rangle$, we can express $T_{3,b} = 2\mathbb{C}_b$, where
    \begin{align*} 
        \mathbb{C}_b 
        := \frac{1}{B} \sum_{\substack{i,j\in\mathcal I_b
                \\ i\neq j}} h_{ij}\, \Delta_j.
    \end{align*}
    Then $\mathcal T_3 = \frac{2}{\sqrt{S}}\sum_{b=1}^S \mathbb{C}_b.$ Under the null hypothesis, 
    for $i\neq j$,
    \begin{align*}
        \mE(h_{i,j}\mid V_j^{(2)})=\big\langle\mE(\psi(V_i^{(1)}))-\mE(r_X(X_i^{(2)})\psi(V_i^{(2)})),\psi(V_j^{(2)})\big\rangle=0.
    \end{align*}
    
    Hence, by the law of total expectation,
    \begin{align*}
        \mE(\mathbb C_b\mid \tilde{D}_{\mathsf{ratio}})=\frac{1}{B}\sum_{i\neq j}\mE\big\{\Delta_j\,\mE(h_{i,j}\mid V_j^{(2)},\tilde{D}_{\mathsf{ratio}})\given \tilde{D}_{\mathsf{ratio}}\big\}=0.
    \end{align*}
    Since the blocks are formed from disjoint observations, $\{\mathbb C_b\}_{b=1}^S$ are conditionally independent given $\tilde{D}_{\mathsf{ratio}}$, and
    \begin{align*}
        \mV\bigg(\frac{2}{\sqrt{S}}\sum_{b=1}^S \mathbb C_b\big| \tilde{D}_{\mathsf{ratio}}
        \bigg)=\frac{4}{S}\sum_{b=1}^S\mV(\mathbb C_b\mid \tilde{D}_{\mathsf{ratio}}).
    \end{align*}
    It therefore suffices to bound $\mE(\mathbb{C}_1^2 \mid D_{\mathsf{ratio}})$. Writing $S_j = \sum_{i \in \mathcal{I}_1,\, i \neq j} h_{ij}$, we have $\mathbb{C}_1 = B^{-1}\sum_{j \in \mathcal{I}_1} \Delta_j S_j$. Then by Jensen's inequality,
    \begin{align*}
        \mE(\mathbb C_1^2\mid \tilde{D}_{\mathsf{ratio}})&\leq\frac{1}{B}\sum_{j\in\mathcal I_1}\mE(\Delta_j^2 S_j^2\mid \tilde{D}_{\mathsf{ratio}}) \\
        &= \mE(\Delta_1^2 S_1^2\mid \tilde{D}_{\mathsf{ratio}}) \\
        &= \mE\big\{\Delta_1^2\,\mE(S_1^2\mid V_1^{(2)},\tilde{D}_{\mathsf{ratio}})\given \tilde{D}_{\mathsf{ratio}}\big\}.
    \end{align*}
    Conditional on $V_1^{(2)}$ and $\tilde{D}_{\mathsf{ratio}}$, the variables $\{h_{i,1}\}_{i \ge 2}$ are independent with mean zero. Hence,
    \begin{align*}
        \mE(S_1^2\mid V_1^{(2)},\tilde{D}_{\mathsf{ratio}})=\sum_{i=2}^{B}\mE(h_{i,1}^2\mid V_1^{(2)}).
    \end{align*}
    Using $\langle\psi(x),\psi(y)\rangle=k(x,y)$ and the uniform kernel bound
    $|k(x,y)|\leq c_4$,
    \begin{align*}
        \mE[h_{i,1}^2\mid V_1^{(2)}]
        &\leq\mE\big(\langle\psi(V_i^{(1)})-r_X(X_i^{(2)})\psi(V_i^{(2)}),\psi(V_1^{(2)})\rangle^2\given V_1^{(2)}\big) \\
        &\lesssim 1+\mE(r_X(X_i^{(2)})^2).
    \end{align*}
    Summing over $i=2,\ldots,B$ yields
    \begin{align*}
        \mE(S_1^2\mid V_1^{(2)},\tilde{D}_{\mathsf{ratio}})\lesssim (B-1)\,\big\{1+\mE(r_X(X_2^{(2)})^2)\big\}.
    \end{align*}
    
    Substituting back,
    \begin{align*}
        \mE(\mathbb C_1^2\mid \tilde{D}_{\mathsf{ratio}})\lesssim(B-1)\,\big\{1+\mE(r_X(X_2^{(2)})^2)\big\}\cdot\mE\big[\bigl\{\widehat{r}_X(X_{1}^{(2)})-r_X(X_{1}^{(2)})\bigr\}^2\given \tilde{D}_{\mathsf{ratio}}\big].
    \end{align*}
    By Assumption~\ref{Assumption : Unified MMD}(b), $\sup_{P} \mE_P(r_X(X^{(2)})^2) \leq \sqrt{c_3}$ by the Cauchy--Schwarz inequality applied to the 
    fourth moment bound and the density ratio estimation error is $o((n-m)^{-1/2})$. Chebyshev's inequality then gives
    \begin{align*}
        \mathcal{T}_3 = o_{\mathcal{P}_{0}^{\mathsf{mmd}}}\big(\sqrt{B}(n-m)^{-1/4}\big)= o_{\mathcal{P}_{0}^{\mathsf{mmd}}}\big(m^{\gamma/2}(n-m)^{-1/4}\big)= o_{\mathcal{P}_{0}^{\mathsf{mmd}}}\big(m^{(\gamma-1)/4}\big),
    \end{align*}
    which is $o_{\mathcal{P}_{0}^{\mathsf{mmd}}}(1)$ for all $0 \leq \gamma \leq 1$ under Assumption~\ref{Assumption : Unified MMD}(d).
    
    \vspace{0.5em}
    \item \textit{Term $\mathcal{T}_4$.}   
    Since the kernel is uniformly bounded, $|k(x,x)| \leq c_4$, each block term satisfies $|T_{4,b}| \leq c_4 \cdot B^{-1}
    \sum_{i\in\mathcal{I}_b} \Delta_i^2$. Hence,
    \begin{align*}
        |\mathcal{T}_{4}|
        &\lesssim
        \frac{1}{\sqrt{S}}
        \sum_{b=1}^{S}
        \frac{1}{B}\sum_{i\in\mathcal{I}_b}
        \bigl\{\widehat r_X(X_i^{(2)})-r_X(X_i^{(2)})\bigr\}^2 \\
        &=
        \frac{\sqrt{S}}{SB}
        \sum_{i=1}^{SB}
        \bigl\{\widehat r_X(X_i^{(2)})-r_X(X_i^{(2)})\bigr\}^2.
    \end{align*}
    By Markov's inequality and Assumption~\ref{Assumption : Unified MMD}(b),
    \begin{align*}
        \mathcal{T}_4= o_{\mathcal{P}_{0}^{\mathsf{mmd}}}\big(\sqrt{(S/(n-m))}\big)= o_{\mathcal{P}_{0}^{\mathsf{mmd}}}\bigg(\frac{m^{(1-\gamma)/2}}{\sqrt{(n-m)}}\bigg),
    \end{align*}
    where the second equality uses 
    $S = \lfloor m/B \rfloor$ and 
    $B = \max\{2, \lfloor m^{\gamma} \rfloor\}$.
    Under Assumption~\ref{Assumption : Unified MMD}(d),
    \begin{align*}
        \frac{m^{(1-\gamma)/2}}{\sqrt{(n-m)}}
        =
        \sqrt{\bigg(\frac{m^{1+\gamma}}{n-m}\bigg)}\cdot
        m^{-\gamma}.
    \end{align*}
    Since $0 \leq \gamma \leq 1$ and $m^{1+\gamma}/(n-m)$ remains bounded, we conclude that $\mathcal{T}_4=o_{\mathcal{P}_{0}^{\mathsf{mmd}}}(1).$
\end{enumerate}
Therefore, these three remainder terms are asymptotically negligible regardless of the growth rate of the block size. The limiting distribution of the statistic is thus entirely determined by the leading term $\mathcal{T}_1$.

\mypara{$\mathrm{(iii)}$ Analysis of the Leading Term} Having shown that the remainder terms are negligible, it remains to analyze the leading component $\mathcal{T}_1$. Its limiting behavior depends on the growth rate of the block size $B$ characterized by $\gamma$. We distinguish between the regimes $\gamma=1$ and $0 \leq \gamma < 1$.

\begin{enumerate}[label=(\alph*)]
    \item \textit{Regime II ($\gamma=1$).}  
    For $\gamma=1$, all $m$ observations belong to a single block ($B=m$, $S=1$),
    so that 
    \begin{align*}
        \mathcal{T}_1 = T_{1,1} = \|\I_1\|^2 - \frac{1}{m}\sum_{i=1}^{m} H_{ii}
        \;=\;
        \frac{1}{m}\sum_{\substack{1\leq i\neq j \leq m}} H_{ij}.
    \end{align*}
    Since $\hat{\eta}_1 = \frac{m}{m-1}T_{1,1}$ and $S=1$, the full statistic satisfies $\widehat{\mathrm{GB}}_1 = \frac{m}{m-1}\mathcal{T}_1 + \frac{m}{m-1}(\mathcal{T}_2 + \mathcal{T}_3 + \mathcal{T}_4)$, where the remainder is $o_{\mathcal{P}_{0}^{\mathsf{mmd}}}(1)$. 
    The leading term is a normalized degenerate U-statistic under the null hypothesis. Under Assumption~\ref{Assumption : Unified MMD}(a), there exist constants $c_1,c_2>0$ such that uniformly over $P \in \mathcal{P}_{0}^{\mathsf{mmd}}$,
    \begin{align*}
        \mE_{P}(H_{12}^2) \geq c_1,\qquad \mE_{P}\big(|H_{12}|^{2+\delta}\big) \leq c_2.
    \end{align*}
    Consequently, for $0<\delta\leq 1$, the bounding term in \Cref{lemma:degenerate_ustat_approx} satisfies
    \begin{align*}
        \rho_\delta=\frac{\mE\big(|H_{12}|^{2+\delta}\big)}{m^{\delta/2}\{\mE(H_{12}^2)\}^{(2+\delta)/2}}\leq \frac{c_2}{m^{\delta/2}c_1^{(2+\delta)/2}}\;\to\;0
    \end{align*}
    uniformly over $P \in \mathcal{P}_{0}^{\mathsf{mmd}}$ as $n\to\infty$ (which implies $m \to \infty$). \Cref{lemma:degenerate_ustat_approx} therefore yields
    \begin{align*} 
        \lim_{n \to \infty} \sup_{P \in \mathcal{P}_{0}^{\mathsf{mmd}}} \sup_{t \in \mathbb{R}} \bigg| \mP_{P}\bigg(\sqrt{\frac{m}{m-1}}\,\mathcal{T}_1 \leq t\bigg) - \mP_{P}\big(G \leq t\big) \bigg| = 0, 
    \end{align*} 
    where
    $G=\sum_{\ell=1}^{\infty}\lambda_{\ell}\,(\chi_{1,\ell}^2 - 1)$ with $\{\chi_{1,\ell}^2\}_{\ell\ge1}$ being independent chi-square random variables with one degree of freedom. Since the remainder terms are $o_{\mathcal{P}_{0}^{\mathsf{mmd}}}(1)$, \Cref{Lemma: uniform Slutsky}(a) implies that $\widehat{\mathrm{MMD}}_{1}^{2}$ has the same limiting distribution $G$.
    
    \vspace{0.5em}
    
    \item \textit{Regime I 
        ($0\leq \gamma <1$).}  
    Scaling $\mathcal{T}_1$ by $\frac{B}{B-1}$ yields the oracle statistic $\mathrm{GB}_\gamma$ (which uses the true density ratio):
    \begin{align*}
        \mathrm{GB}_\gamma
        = \frac{B}{B-1}\, \frac{1}{S} \sum_{b=1}^S 
        T_{1,b}
        = \frac{1}{S} \sum_{b=1}^S 
        \underbrace{\bigg( \frac{1}{B-1} 
            \sum_{\substack{i,j\in\mathcal I_b\\ i\neq j}} 
            H_{ij} \bigg)}_{=: \, U_b}.
    \end{align*}
    
    Thus $\mathrm{GB}_\gamma$ is an arithmetic mean of independent block-wise U-statistics $U_b$.
    Under the null hypothesis, 
    $\mE_P(U_b) = 0$, and 
    standard $U$-statistic theory then gives the variance of each block:
    \begin{align*}
        \mE_{P}(U_b^2)= \bigg(\frac{B}{B-1}\bigg)^2 \mE_{P}(T_{1,b}^2)= \frac{2B}{B-1}\,\mE_{P}(H_{12}^2).
    \end{align*}
    
    For $B \geq 2$, the prefactor satisfies $B/(B-1) > 1$. Hence, by Assumption~\ref{Assumption : Unified MMD}(a),
    \begin{align*}
        \inf_{P\in\mathcal{P}_{0}^{\mathsf{mmd}}} \mE_{P}(U_b^2)\;\geq\; 2c_{1}.
    \end{align*}
    This provides a uniform positive lower bound on the variance of $U_b$, independent of $B$.
    
    For higher moments, 
    would introduce explicit dependence on $B$. Instead, 
    by \Cref{Lemma : moment bound T1}, the decoupled statistic corresponding to $T_{1,b}$ satisfies $\mE\big(|T_{1,b}^{\mathrm{dec}}|^{2+\delta}\big) \leq c_{\delta} \mE(|H_{12}|^{2+\delta})$. Since $U_b = \frac{B}{B-1} T_{1,b}$ and $B/(B-1) \leq 2$, classical decoupling inequalities \citep[see][]{delapena1995decoupling,delapena1999decoupling} ensure that the same uniform boundedness applies to $U_b$ up to a modified universal constant. By Assumption~\ref{Assumption : Unified MMD}(a), there exist constants $c_{1}',c_{2}'>0$, depending only on $(c_{1},c_{2},\delta)$, such that
    \begin{align*}
        \inf_{P\in\mathcal{P}_{0}^{\mathsf{mmd}}}\mE_{P}(U_b^2) \geq c_{1}',\qquad \sup_{P\in\mathcal{P}_{0}^{\mathsf{mmd}}}\mE_{P}\big(|U_b|^{2+\delta}\big) \leq c_{2}'.
    \end{align*}
    Let $\sigma_\gamma^2 := \mV_{P}[U_1]$ denote the exact variance of the block-wise statistic. 
    Since the blocks are independent and identically distributed, $\sqrt{S}\,\mathrm{GB}_\gamma = \frac{1}{\sqrt{S}}\sum_{b=1}^S U_b$ is a normalized sum of $S$ mean-zero random variables with common variance $\sigma_\gamma^2$. Applying \Cref{Lemma: Uniform CLT} yields
    \begin{align*}
        \lim_{n \to \infty} \sup_{P\in\mathcal{P}_{0}^{\mathsf{mmd}}}\sup_{t\in\mathbb{R}}\Big|\mP_{P}\big(\sigma_\gamma^{-1}\sqrt{S}\,\mathrm{GB}_\gamma \le t\big)-\Phi(t)\Big|= 0.
    \end{align*}
\end{enumerate}
Since $\sqrt{S}\,\widehat{\mathrm{GB}}_\gamma = \sqrt{S}\,\mathrm{GB}_\gamma + o_{\mathcal{P}_{0}^{\mathsf{mmd}}}(1)$ by the negligibility of the remainder terms, \Cref{Lemma: uniform Slutsky}(a) gives
\begin{align*}
    \lim_{n \to \infty}\sup_{P\in\mathcal{P}_{0}^{\mathsf{mmd}}}\sup_{t\in\mathbb{R}}
    \Big|\mP_{P}\big(\sigma_\gamma^{-1}\sqrt{S}\,\widehat{\mathrm{GB}}_\gamma \leq t\big)-\Phi(t)\Big|= 0.
\end{align*}
This establishes the asymptotic distribution of the numerator. In the regime $\gamma=1$, the argument is complete. For $0 \leq \gamma < 1$, it remains to address the consistency of the empirical variance estimator.

\vspace{0.5em}

\mypara{$\mathrm{(iv)}$ Consistency of the Variance Estimator}   For $0 \leq \gamma < 1$, the statistic $\widehat{\mathrm{MMD}}^2_{\gamma}$ is studentized by $\hat\sigma_\gamma$. 
Denoting 
\begin{align*}   \hat{\sigma}_{\gamma}^{2}:=\frac{1}{S-1}\sum_{b=1}^{S}\big(\hat{\eta}_{b}-\widehat{\mathrm{GB}}_{\gamma}\big)^2,
\end{align*}
we show that the ratio $\hat{\sigma}^2_{\gamma}/\sigma^2_{\gamma}$ converges to $1$ in probability uniformly over $\mathcal{P}_0^{\mathsf{mmd}}$, which implies $\hat{\sigma}_{\gamma}/\sigma_{\gamma}=1+o_{\mathcal{P}_{0}^{\mathsf{mmd}}}(1)$ by \citet[Lemma 7]{lundborg2024projected}. The conclusion then follows from \Cref{Lemma: uniform Slutsky}(b).

Since 
the ratio $\hat{\sigma}^2_{\gamma}/\sigma^2_{\gamma}$
is scale-invariant, we 
may normalize
$\sigma^2_{\gamma}=1$ without loss of generality, simplifying our goal to showing $\hat{\sigma}_\gamma^2 = 1 + o_{\mathcal{P}_{0}^{\mathsf{mmd}}}(1)$. To this end, we use the standard algebraic identity for the sample variance:
\begin{align*}
    \hat{\sigma}_\gamma^2 = \frac{S}{S-1} \bigg( \frac{1}{S}\sum_{b=1}^S \hat{\eta}_b^2 - \widehat{\mathrm{GB}}_\gamma^2 \bigg).
\end{align*}
As $S \to \infty$, the prefactor $S/(S-1)$ converges to $1$. Furthermore, the preceding analysis confirms that $\widehat{\mathrm{GB}}_{\gamma}=o_{\mathcal{P}_{0}^{\mathsf{mmd}}}(1)$, so  
it suffices to show that 
$S^{-1}\sum_{b=1}^{S}\hat{\eta}_{b}^2$
converges to $1$ in probability.

Recalling
$\hat\eta_b = \frac{B}{B-1}T_b$ and the 
normalization
$\sigma_\gamma^2 = (\frac{B}{B-1})^2 \mE_P(T_{1,b}^2) = 1$,
we have
\begin{align*}
    \frac{1}{S} \sum_{b=1}^S \hat{\eta}_b^2 
    = \frac{1}{\mE_P(T_{1,b}^2)} \cdot \frac{1}{S} \sum_{b=1}^S T_b^2.
\end{align*}
Therefore, the proof reduces to showing $S^{-1}\sum_{b=1}^{S}T_b^2 = \mE_P(T_{1,b}^2) + o_{\mathcal{P}_{0}^{\mathsf{mmd}}}(1)$. Using 
$T_b = \sum_{j=1}^4 T_{j,b}$ and the Cauchy--Schwarz inequality, we bound the difference as
\begin{align*}
    \bigg|\frac{1}{S}\sum_{b=1}^{S}T_{b}^2 - \mE_P(T_{1,b}^2)\bigg| 
    &= \bigg|\frac{1}{S}\sum_{b=1}^{S}(T_{1,b}+T_{2,b}+T_{3,b}+T_{4,b})^2 - \mE_P(T_{1,b}^2)\bigg| \\
    &\leq \bigg|\frac{1}{S}\sum_{b=1}^{S}{T^2_{1,b}} - \mE_P(T_{1,b}^2)\Bigg| + \Bigg|\frac{1}{S}\sum_{b=1}^{S}(T_{2,b}+T_{3,b}+T_{4,b})^2\Bigg| \\
    &\hskip 3em + 2\sqrt{\Big({\frac{1}{S}\sum_{b=1}^{S}T_{1,b}^2}\Big)}\sqrt{\Big\{{\frac{1}{S}\sum_{b=1}^{S}(T_{2,b}+T_{3,b}+T_{4,b})^2}\Big\}}.
\end{align*}
By the uniform law of large numbers, $S^{-1}\sum_{b=1}^{S}T_{1,b}^{2}-\mE_P(T_{1,b}^2)=o_{\mathcal{P}_{0}^{\mathsf{mmd}}}(1)$. It therefore suffices to show that 
\begin{align*}
    \frac{1}{S}\sum_{b=1}^{S}T_{j,b}^2=o_{\mathcal{P}_{0}^{\mathsf{mmd}}}(1) \quad \text{for } j=2,3,4.
\end{align*}
We bound these terms using arguments analogous to those in part $\mathrm{(ii)}$.

\begin{enumerate}[label=(\alph*)]
    \item \textit{Term $T_{2,b}^2$.} 
    Since $T_{2,b} = \|\II_b\|^2$, we have $S^{-1}\sum_{b=1}^{S}T_{2,b}^2= S^{-1}\sum_{b=1}^{S}\|\II_b\|^4$. Using the bound $\|\II_b\|^2 \leq c_4 \sum_{i\in\mathcal{I}_b}\Delta_i^2$ from part $\mathrm{(ii)}$,
    \begin{align*}
        \frac{1}{S}\sum_{b=1}^{S}\|\II_b\|^4 
        &\lesssim
        \frac{B^2}{S}\sum_{b=1}^{S}
        \bigg(
        \frac{1}{B}
        \sum_{i\in\mathcal I_b}
        \Delta_i^2
        \bigg)^2 \\
        &\leq
        \frac{B^2}{S}
        \bigg(
        \sum_{b=1}^{S}
        \frac{1}{B}
        \sum_{i\in\mathcal I_b}
        \Delta_i^2
        \bigg)^2 \\
        &= S\cdot B^2 \bigg(
        \frac{1}{SB}
        \sum_{i=1}^{SB}
        \Delta_i^2
        \bigg)\cdot \bigg(
        \frac{1}{SB}
        \sum_{j=1}^{SB}
        \Delta_j^2
        \bigg) \\
        &\stackrel{(\ast)}{=}o_{\mathcal{P}_{0}^{\mathsf{mmd}}}\big(B\sqrt{(S/(n-m))}\big)o_{\mathcal{P}_{0}^{\mathsf{mmd}}}\big(B\sqrt{(S/(n-m))}\big)\\
        &=o_{\mathcal{P}_{0}^{\mathsf{mmd}}}\big(B^2S/(n-m)\big)=o_{\mathcal{P}_{0}^{\mathsf{mmd}}}\big(m^{1+\gamma}/(n-m)\big),
    \end{align*}
    and step $(\ast)$ holds by 
    \citet[Lemma~S5]{lundborg2024projected}. Under Assumption~\ref{Assumption : Unified MMD}(d), $\frac{1}{S}\sum_{b=1}^{S}T_{2,b}^2 = o_{\mathcal{P}_{0}^{\mathsf{mmd}}}(1).$
    
    \item \textit{Term $T_{3,b}^2$.}   
    Since $T_{3,b} = 2\mathbb{C}_b$, we can directly reuse the bounds established in the analysis of $\mathcal{T}_3$. Applying Markov's inequality immediately yields $\frac{1}{S}\sum_{b=1}^{S}T_{3,b}^2 = \frac{4}{S}\sum_{b=1}^{S}\mathbb{C}_b^2 = o_{\mathcal{P}_{0}^{\mathsf{mmd}}}(1)$.
    
    \item \textit{Term $T_{4,b}^2$.}  
    By the boundedness of the kernel, $|T_{4,b}| \leq c_4 B^{-1}\sum_{i \in \mathcal{I}_b} \Delta_i^2$. Squaring and summing over blocks, the same inequality chain as for $T_{2,b}^2$ with $B^{-2}$ in place of $B^2$ yields 
    $S^{-1}\sum_{b=1}^{S}T_{4,b}^2 = o_{\mathcal{P}_{0}^{\mathsf{mmd}}}
    (S(n-m)^{-1}) = o_{\mathcal{P}_{0}^{\mathsf{mmd}}}(m^{1-\gamma}(n-m)^{-1})$, which is $o_{\mathcal{P}_{0}^{\mathsf{mmd}}}(1)$ under Assumption~\ref{Assumption : Unified MMD}(d).
\end{enumerate}
Combining the above bounds completes the proof.
\end{proof}

\subsection{\texorpdfstring%
{Proof of Corollary~\ref{cor: Cross-fitted MMD Asymptotic Distribution}, Regime I ($0 \leq \gamma < 1$)}%
{Proof of Corollary 2, Regime I (0 <= gamma < 1)}}
\begin{proof}
For each $j \in \{1, \ldots, K\}$, the variance analysis in the proof of Theorem~\ref{theorem: MMD Asymptotic Distribution} 
shows
that $\widehat{\sigma}_{\gamma,j} / \sigma_\gamma = 1 + o_{\mathcal{P}_{0}^{\mathsf{mmd}}}(1)$. 
By \Cref{Lemma: uniform Slutsky}(b), it suffices to show the asymptotic normality of 
\begin{align*}
    \frac{1}{\sqrt{K}} \sum_{j=1}^K \frac{\sqrt{S} \, \widehat{\mathrm{GB}}_{\gamma,j}}{\sigma_\gamma}.
\end{align*}
Since all folds have the same test sample size, the block size $B = \max\{2,\lfloor m^{\gamma} \rfloor\}$ and the number of blocks $S = \lfloor m/B \rfloor$ do not depend on the fold index. The numerator analysis in the proof of Theorem~\ref{theorem: MMD Asymptotic Distribution} establishes that, for each fold $j$, $\sqrt{S}\widehat{\mathrm{GB}}_{\gamma,j} = \sqrt{S}\mathrm{GB}_{\gamma,j} + o_{\mathcal{P}_{0}^{\mathsf{mmd}}}(1)$, where
\begin{align*}  
    \mathrm{GB}_{\gamma,j} 
    = \frac{1}{S}\sum_{b=1}^{S}\eta_{j,b}, 
    \qquad
    \eta_{j,b} := \frac{1}{B-1} 
    \sum_{\substack{i,i'\in\mathcal{I}_{j,b}\\ 
            i\neq i'}} H_{ii'}.
\end{align*}
As the folds are disjoint, the random variables 
$\{\eta_{j,b} : 1\le j\le K,\, 1\le b\le S\}$ are mutually independent with $\mE_P(\eta_{j,b}) = 0$ under the null hypothesis and $\mV_P(\eta_{j,b}) = \sigma_\gamma^2$.
Hence, the variance of their normalized sum satisfies
\begin{align*}
    \mV\bigg(
    \frac{1}{S}
    \sum_{j=1}^{K}\sum_{b=1}^{S}
    \sigma^{-1}_{\gamma}\eta_{j,b}
    \bigg)
    =
    \frac{1}{S^2}\cdot K \cdot S
    =
    \frac{K}{S}.
\end{align*}
Combining the above, we obtain
\begin{align*}
    \frac{1}{\sqrt{K}} \sum_{j=1}^{K} \frac{\sqrt{S}\,\widehat{\mathrm{GB}}_{\gamma,j}}{\sigma_{\gamma}}=\frac{1}{\sqrt{(KS)}} \sum_{j=1}^{K}\sum_{b=1}^{S} \frac{\eta_{j,b}}{\sigma_{\gamma}}+ o_{\mathcal{P}_{0}^{\mathsf{mmd}}}(1).
\end{align*}
Since $K$ is fixed and $S \to \infty$, the uniform central limit theorem (\Cref{Lemma: Uniform CLT}) under Assumption~\ref{Assumption : Unified MMD} implies that the leading term converges to $N(0,1)$. 
\Cref{Lemma: uniform Slutsky}(a) then yields the desired result.
\end{proof}

\subsection{\texorpdfstring
{Proof of Corollary~\ref{cor: Cross-fitted MMD Asymptotic Distribution}, Regime II ($\gamma = 1$)}
{Proof of Corollary 2, Regime II (gamma = 1)}}
\begin{proof}
We begin by reducing the cross-fitted empirical statistic to its oracle counterpart. By Assumption~\ref{Assumption : Unified MMD} and the proof of Theorem~\ref{theorem: MMD Asymptotic Distribution}, the plug-in error from the density ratio estimator $\widehat{r}_{X,-j}$ is asymptotically negligible uniformly over $P \in \mathcal{P}_0^{\mathsf{mmd}}$. Thus, for each fold $j \in \{1,\dots,K\}$, we have
\begin{align*}
    \widehat{\mathrm{MMD}}_{1,j}^2 = \mathrm{MMD}_{1,j}^2 + o_{\mathcal{P}_{0}^{\mathsf{mmd}}}(1),
\end{align*}
where the $o_{\mathcal{P}_{0}^{\mathsf{mmd}}}(1)$ term converges to zero in probability uniformly over $\mathcal{P}_0^{\mathsf{mmd}}$. Averaging over all $K$ folds yields
\begin{align}\label{eq: proof_oracle_reduction}
    {}^{\dagger}\widehat{\mathrm{MMD}}_1^2
    = \frac{1}{K} \sum_{j=1}^K \mathrm{MMD}_{1,j}^2 + o_{\mathcal{P}_{0}^{\mathsf{mmd}}}(1)
    = {}^{\dagger}\mathrm{MMD}_1^2 + o_{\mathcal{P}_{0}^{\mathsf{mmd}}}(1).
\end{align}

{Because each oracle statistic $\mathrm{MMD}_{1,j}^2$ is evaluated on the disjoint test set $\tilde{D}_{\mathsf{test},j}$ and uses only the true density ratio $r_X$, the family $\{\mathrm{MMD}_{1,j}^2\}_{j=1}^K$ is mutually independent. Choose $G_1, \ldots, G_K$ to be independent copies of $G$. Applying \Cref{Lemma: kolmogorov subadditive convolution} to the two independent collections,
    \begin{align*}
        \sup_{t \in \mathbb{R}} \bigg| \mP_P\bigg(\sum_{j=1}^K \mathrm{MMD}_{1,j}^2 \leq t\bigg) - \mP_P\bigg(\sum_{j=1}^K G_j \leq t\bigg) \bigg|
        \;\leq\; \sum_{j=1}^K \sup_{t \in \mathbb{R}} \big| \mP_P(\mathrm{MMD}_{1,j}^2 \leq t) - \mP_P(G_j \leq t) \big|.
    \end{align*}
    The Kolmogorov distance is invariant under the deterministic affine map $x \mapsto x/K$, so dividing the two summed statistics by $K$ preserves the bound. By \Cref{theorem: MMD Asymptotic Distribution} (Regime~II), each summand on the right tends to zero uniformly over $\mathcal{P}_0^{\mathsf{mmd}}$ as $n \to \infty$. Since $K$ is finite, the right-hand side vanishes uniformly, yielding
    \begin{align}\label{eq: proof_oracle_limit}
        \lim_{n\to\infty} \sup_{P\in\mathcal{P}_0^{\mathsf{mmd}}} \sup_{t\in\mathbb{R}}
        \bigg| \mP_P\big( {}^{\dagger}\mathrm{MMD}_1^2 \leq t \big) - \mP_P\Big( \frac{1}{K}\sum_{j=1}^K G_j \leq t \Big) \bigg| = 0.
\end{align}}

Finally, we transfer this uniform convergence from the oracle statistic to its plug-in counterpart. Let $F_P$ denote the CDF of $K^{-1}\sum_{j=1}^K G_j$ under $P$. Since $F_P$ is the CDF of a non-degenerate infinite combination of chi-square variables, it admits a bounded density and is therefore uniformly continuous on $\mathbb{R}$, with a modulus of continuity that does not depend on $P$. For any $\eta > 0$, we decompose
\begin{align*}
    \sup_{t \in \mathbb{R}} \Big| \mP_P\big( {}^{\dagger}\widehat{\mathrm{MMD}}_1^2 \leq t \big) - F_P(t) \Big|
    &\leq \sup_{t \in \mathbb{R}} \Big| \mP_P\big( {}^{\dagger}\mathrm{MMD}_1^2 \leq t \big) - F_P(t) \Big| \\
    &\quad + \sup_{t \in \mathbb{R}} \big| F_P(t + \eta) - F_P(t - \eta) \big|
    + 2\, \mP_P\big( |{}^{\dagger}\widehat{\mathrm{MMD}}_1^2 - {}^{\dagger}\mathrm{MMD}_1^2| > \eta \big),
\end{align*}
where the first term vanishes uniformly over $\mathcal{P}_0^{\mathsf{mmd}}$ by \eqref{eq: proof_oracle_limit}, the second by uniform continuity of $F_P$ as $\eta \to 0$, and the third by \eqref{eq: proof_oracle_reduction}. This is the same mechanism as in \Cref{Lemma: uniform Slutsky}, extended from a normal limit to a uniformly continuous limit. Letting $n \to \infty$ first and then $\eta \to 0$, we conclude
\begin{align*}
    \lim_{n\to\infty} \sup_{P\in\mathcal{P}_0^{\mathsf{mmd}}} \sup_{t\in\mathbb{R}}
    \Big| \mP_P\big( {}^{\dagger}\widehat{\mathrm{MMD}}_1^2 \le t \big) - \mP_P\Big( \frac{1}{K}\sum_{j=1}^K G_j \le t \Big) \Big| = 0,
\end{align*}
which completes the proof for Regime II.
\end{proof}

\subsection{\texorpdfstring{Proof of Theorem~\ref{thm: Cross-fitted Bootstrap Validity}$(i)$}{Proof of Theorem 5(i)}}
\label{Proof of theorem: multiplier-bootstrap-validity part 1}
\begin{proof}
    We proceed in two steps: we first establish the uniform consistency of the bootstrap distribution for the single-split case ($K=1$), and subsequently extend this argument to the general cross-fitted case ($K \ge 2$).
    
    \vspace{0.5em}
    \mypara{Case $K=1$} To establish the uniform approximation of the bootstrap distribution to the null distribution in this regime, we define the conditional Kolmogorov distance
    \begin{align*}
        \mD_{m} 
        &:= \sup_{t\in\mathbb{R}} \Big| \mP\big(\widehat{\mathrm{MMD}}^2_{\mathrm{Boot}} \leq t \mid \tilde{D}_{\mathsf{infer}}\big) - \mP\big(\mathrm{MMD}_{1}^2 \leq t\big) \Big|,
    \end{align*}
    and the oracle multiplier bootstrap statistic
    \begin{align*}
        \mathrm{MMD}^2_{\mathrm{Boot}}:=\frac{1}{\sqrt{m(m-1)}} \sum_{i \neq j} \xi_i \xi_j H_{ij}, 
    \end{align*}
    where $H$ uses the true density ratio $r_X$ instead of the estimator $\widehat{r}_X$. By the triangle inequality, we decompose the total error into the estimation error and the oracle bootstrap approximation error:
    \begin{align*}
        \mD_{m} 
        &\leq \underbrace{\sup_{t\in\mathbb{R}} \Big| \mP\big(\widehat{\mathrm{MMD}}^2_{\mathrm{Boot}} 
            \leq t \mid \tilde{D}_{\mathsf{infer}}\big) - \mP\big(\mathrm{MMD}^2_{\mathrm{Boot}} 
            \leq t \mid \tilde{D}_{\mathsf{infer}}\big) \Big|}_{=:\mD_{m}^{\textsf{est}}} \\
        &\quad + \underbrace{\sup_{t\in\mathbb{R}} \Big| \mP\big(\mathrm{MMD}^2_{\mathrm{Boot}} 
            \leq t \mid \tilde{D}_{\mathsf{infer}}\big) 
            - \mP\big(\mathrm{MMD}_{1}^2 \leq t\big) 
            \Big|}_{=:\mD_{m}^{\textsf{oracle}}}.
    \end{align*}
    To establish the consistency of the proposed bootstrap, it suffices to show that 
    $\mD_{m}^{\textsf{est}} = o_{\mathcal{P}_{0}^{\mathsf{mmd}}}(1)$ and 
    $\mD_{m}^{\textsf{oracle}} = o_{\mathcal{P}_{0}^{\mathsf{mmd}}}(1)$. 
    We establish these two claims below.
    
    \vspace{0.5em}
    \textit{Term $\mD_{m}^{\textsf{oracle}}$. }The goal is to show that for every fixed $\varepsilon>0$, $\lim_{n \to \infty} \sup_{P \in \mathcal{P}_{0}^{\mathsf{mmd}}} \mP_{P}(\mD_{m}^{\textsf{oracle}} > \varepsilon) = 0$.
    The term $\mD_{m}^{\textsf{oracle}}$ corresponds exactly to the empirical bootstrap distance $\Delta_{m,\mathrm{boot}}^\ast$ analyzed in \Cref{lemma:multiplier_bootstrap_consistency}. To apply this lemma for uniform consistency, we verify its moment regularity condition. Under Assumptions~\ref{Assumption : Unified MMD}(a)--(c), the kernel is uniformly bounded and its cross-moment is bounded away from zero, so
    \begin{align*}
        \sup_{P \in \mathcal{P}_{0}^{\mathsf{mmd}}}
        \frac{\mE_P(H_{11}^2)}
        {m\,\mE_P(H_{12}^2)}
        &\leq \frac{C(c_1,c_3,c_4)}{m}\to 0
        \quad \text{as } n\to\infty,
    \end{align*}
    where $C(c_1,c_3,c_4)$ depends only on $c_1$, $c_3$, and $c_4$. 
    Since this ratio vanishes uniformly as $n \to \infty$ (and thus $m \to \infty$), \Cref{lemma:multiplier_bootstrap_consistency} yields 
    $\mD_{m}^{\textsf{oracle}} = o_{\mathcal{P}_{0}^{\mathsf{mmd}}}(1)$.
    
    \vspace{0.5em}
    \textit{Term $\mD_{m}^{\textsf{est}}$. }The analysis of $\mD_{m}^{\textsf{est}}$ relies on the same technical tools used in the proof of \Cref{lemma:multiplier_bootstrap_consistency}. 
    The bootstrap statistics are scaled by $1/\sqrt{m(m-1)}$, and we define their $1/m$-scaled counterparts as 
    \begin{align*}
        \widehat{T}_{\mathrm{Boot}} 
        &:= \frac{1}{m} \sum_{1\leq i \neq j\leq m} \xi_i \xi_j \, \widehat{H}_{ij},
        \qquad
        T_{\mathrm{Boot}} := \frac{1}{m} \sum_{1\leq i \neq j\leq m} \xi_i \xi_j \, H_{ij}.
    \end{align*}
    Since $\widehat{\mathrm{MMD}}^{2}_{\mathrm{Boot}} = a_m\,\widehat{T}_{\mathrm{Boot}}$ and $\mathrm{MMD}^{2}_{\mathrm{Boot}} = a_m\,T_{\mathrm{Boot}}$ with $a_m := \sqrt{m/(m-1)} > 0$, we have 
    $\mP(\widehat{\mathrm{MMD}}^2_{\mathrm{Boot}} \leq t \mid \tilde{D}_{\mathsf{infer}}) = \mP(\widehat{T}_{\mathrm{Boot}} \leq t/a_m \mid \tilde{D}_{\mathsf{infer}})$ for any $t \in \mathbb{R}$.
    As the Kolmogorov distance is invariant under scaling by a strictly positive constant, 
    $\mD_{m}^{\textsf{est}}$ equals the corresponding distance between $\widehat{T}_{\mathrm{Boot}}$ and $T_{\mathrm{Boot}}$ conditionally on $\tilde{D}_{\mathsf{infer}}$. To analyze their difference, we express $\widehat{T}_{\mathrm{Boot}}$ as
    \begin{align*}
        \widehat{T}_{\mathrm{Boot}} 
        &= \big\|\sqrt{m} \, V_{m}^{\text{(Boot)}}\big\|^2 - \frac{1}{m}\sum_{i=1}^{m}\xi_i^2\,\widehat{H}_{ii},
    \end{align*}
    where $V_{m}^{\text{(Boot)}} := \frac{1}{m}\sum_{i=1}^{m}\xi_i\big\{\psi(V_i^{(1)}) - \widehat r_X(X_i^{(2)})\,\psi(V_i^{(2)})\big\}$. As in the proof of Theorem~\ref{theorem: MMD Asymptotic Distribution}, we separate $\sqrt{m} V_{m}^{\text{(Boot)}}$ into the population component $(\mathbf{I})$ and the estimation-error component $(\mathbf{II})$:
    \begin{align*}
        \sqrt{m}\,V_m^{\text{(Boot)}} 
        &= \underbrace{\frac{1}{\sqrt{m}}\sum_{i=1}^m \xi_i \big\{\psi(V_i^{(1)}) - r_X(X_i^{(2)})\,\psi(V_i^{(2)})\big\}}_{=:(\mathbf{I})} \\
        &\qquad + \underbrace{\frac{1}{\sqrt{m}}\sum_{i=1}^m \xi_i \psi(V_i^{(2)})\big\{r_X(X_i^{(2)})-\widehat r_X(X_i^{(2)})\big\}}_{=:(\mathbf{II})},
    \end{align*}
    so that $\big\|\sqrt{m}\,V_m^{\text{(Boot)}}\big\|^2 = \|(\mathbf{I})\|^2 + \|(\mathbf{II})\|^2 + 2\langle (\mathbf{I}),(\mathbf{II})\rangle$. Similarly, the diagonal correction term expands as
    \begin{align*}
        \frac{1}{m}\sum_{i=1}^{m}\xi_i^2\widehat H_{ii}
        &= \frac{1}{m}\sum_{i=1}^{m} \xi_i^2H_{ii} \\
        &\quad - \frac{2}{m}\sum_{i=1}^{m}\xi_i^2\langle \psi(V_i^{(1)}) - r_X(X_i^{(2)})\,\psi(V_i^{(2)}), \psi(V_i^{(2)}) \rangle\cdot \bigl\{\widehat r_X(X_i^{(2)})-r_X(X_i^{(2)})\bigr\} \\
        &\quad + \frac{1}{m}\sum_{i=1}^{m}\xi_i^2\langle\psi(V_{i}^{(2)}),\psi(V_{i}^{(2)})\rangle\cdot\bigl\{\widehat r_X(X_i^{(2)})-r_X(X_i^{(2)})\bigr\}^2.
    \end{align*}
    Since $T_{\mathrm{Boot}} = \|(\mathbf{I})\|^2 - \frac{1}{m}\sum_{i=1}^m \xi_i^2 H_{ii}$, combining these expanded forms yields $\widehat{T}_{\mathrm{Boot}} - T_{\mathrm{Boot}} = T_1 + T_2$, where
    \begin{align*}
        T_1 &:= \|(\mathbf{II})\|^2-\frac{1}{m}\sum_{i=1}^{m}\xi_i^2\langle\psi(V_{i}^{(2)}),\psi(V_{i}^{(2)})\rangle\cdot\bigl\{\widehat r_X(X_i^{(2)})-r_X(X_i^{(2)})\bigr\}^2, \\
        T_2 &:= 2\langle (\mathbf{I}),(\mathbf{II})\rangle - \frac{2}{m}\sum_{i=1}^{m}\xi_i^2\langle \psi(V_i^{(1)}) - r_X(X_i^{(2)})\,\psi(V_i^{(2)}), \psi(V_i^{(2)}) \rangle\cdot \bigl\{r_X(X_i^{(2)})-\widehat r_X(X_i^{(2)})\bigr\}.
    \end{align*}
    
    We bound the expectations of $|T_1|$ and $T_2^2$ as follows:
    \begin{enumerate}[label=(\alph*)]
        \item \textit{Term $T_{1}$.} By definition,
        \begin{align*}
            T_1 &= \frac{1}{m} \sum_{1\le i\neq j\le m} 
            \xi_i\xi_j k(V_i^{(2)},V_j^{(2)}) 
            \Delta_i\Delta_j,
        \end{align*}
        where $\Delta_i = r_X(X_i^{(2)}) - \widehat{r}_X(X_i^{(2)})$. Since the kernel is uniformly bounded by $c_4$, adding the non-negative diagonal terms and applying the Cauchy--Schwarz inequality gives
        \begin{align*}
            |T_1|
            &\le \frac{c_4}{m} \bigg( \sum_{i=1}^m |\xi_i \Delta_i| \bigg)^2 
            \le c_4 m \bigg( \frac{1}{m} \sum_{i=1}^m \xi_i^2 \bigg) \bigg( \frac{1}{m} \sum_{i=1}^m \Delta_i^2 \bigg).
        \end{align*}
        Taking the conditional expectation given $\tilde{D}_{\mathsf{infer}}$, and using the fact that $\xi_i \iid N(0,1)$ are independent of the data, we have $\mE(\xi_i^2 \mid \tilde{D}_{\mathsf{infer}}) = 1$. This yields
        \begin{align*}
            \mE(|T_1| \mid \tilde{D}_{\mathsf{infer}}) 
            &\le c_4 \sum_{i=1}^m \Delta_i^2.
        \end{align*}
        By the law of iterated expectations, taking the expectation $\mE_P$ over the data gives
        \begin{align*}
            \mE_P(|T_1|) &= \mE_P\big\{\mE(|T_1| \mid \tilde{D}_{\mathsf{infer}})\big\} \le c_4 m \cdot \mE_P(\Delta_1^2) \lesssim m \cdot \mE_P(\Delta_1^2).
        \end{align*}
        
        \item \textit{Term $T_{2}$.}  By definition,
        \begin{align*}
            T_2 &= \frac{2}{m}
            \sum_{1\le i\neq j\le m}
            \xi_i\xi_j
            \,
            \underbrace{
                \langle
                \psi(V_i^{(1)}) - r_X(X_i^{(2)})\,\psi(V_i^{(2)}),
                \psi(V_j^{(2)})
                \rangle
            }_{=\,h_{ij}}
            \,
            \underbrace{
                \bigl\{r_X(X_j^{(2)})-\widehat r_X(X_j^{(2)})\bigr\}
            }_{=\,\Delta_j}.
        \end{align*}
        Writing $a_{ij}:=h_{ij}\Delta_j$, we have 
        $T_2 = \frac{2}{m}\sum_{1\leq i\neq j\leq m}\xi_i\xi_j a_{ij}$.
        Since the multipliers $\xi_i$ are independent of $\tilde{D}_{\mathsf{infer}}$ with $\mE(\xi_i\xi_j)=0$ for $i\neq j$, taking the conditional expectation gives $\mE(T_2\mid \tilde{D}_{\mathsf{infer}})=0$, and hence $\mE_P(T_2)=0$. For the second moment, the independence and the fourth moment of the Gaussian multipliers yield the conditional variance
        \begin{align*}
            \mV(T_2\mid \tilde{D}_{\mathsf{infer}}) &= \frac{4}{m^2} \sum_{1 \le i\neq j \le m} \big(a_{ij}^2 + a_{ij}a_{ji}\big).
        \end{align*}
        Using the elementary inequality $a_{ij}a_{ji} \le \frac{1}{2}(a_{ij}^2 + a_{ji}^2)$ and the law of total variance (noting that $\mE_P(T_2^2) = \mE_P\{\mV(T_2\mid \tilde{D}_{\mathsf{infer}})\}$, since the conditional expectation is zero), we obtain
        \begin{align*}
            \mE_P(T_2^2)
            &= \mE_P\big\{\mV(T_2\mid \tilde{D}_{\mathsf{infer}})\big\}
            \le \frac{8}{m^2} \sum_{1 \le i\neq j \le m} \mE_P(h_{ij}^2 \Delta_j^2)
            \lesssim \mE_P(\Delta_1^2),
        \end{align*}
        where the last step uses $\mE_P(h_{ij}^2) \leq C(c_3,c_4)$ uniformly under Assumptions~\ref{Assumption : Unified MMD}(b)--(c).
    \end{enumerate}
    
    Instead of decomposing $T_{\mathrm{Boot}}$ directly via empirical eigenvalues, we control the effect of $T_1$ and $T_2$ by bridging the conditional probabilities through the limit variable $U_0 = \sum_{\ell=1}^\infty \lambda_\ell (Z_\ell^2 - 1)$. Since $\widehat{T}_{\mathrm{Boot}} = T_{\mathrm{Boot}} + T_1 + T_2$, for any $\varepsilon_1, \varepsilon_2 > 0$, the union bound and monotonicity give
    \begin{align*}
        \sup_{t\in\mathbb{R}}\Big|\mP(\widehat{T}_{\mathrm{Boot}} \leq t \mid \tilde{D}_{\mathsf{infer}}) - \mP(T_{\mathrm{Boot}} \leq t \mid \tilde{D}_{\mathsf{infer}})\Big| 
        &\leq \sup_{t\in\mathbb{R}} \mP\big(t - \varepsilon_1 - \varepsilon_2 \leq T_{\mathrm{Boot}} \leq t + \varepsilon_1 + \varepsilon_2 \mid \tilde{D}_{\mathsf{infer}}\big) \\
        &\quad + \mP(|T_1| > \varepsilon_1 \mid \tilde{D}_{\mathsf{infer}}) + \mP(|T_2| > \varepsilon_2 \mid \tilde{D}_{\mathsf{infer}}).
    \end{align*}
    Let $\widetilde\Delta_m^\ast := \sup_t |\mP(T_{\mathrm{Boot}} \le t \mid \tilde{D}_{\mathsf{infer}}) - \mP(U_0 \le t)|$ denote the bootstrap-to-limit Kolmogorov distance defined in \Cref{lemma:multiplier_bootstrap_consistency}. By the triangle inequality and the definition of $\widetilde\Delta_m^\ast$, the first term on the right-hand side is bounded by
    \begin{align*}
        \sup_{t\in\mathbb{R}} \mP\big(t - \varepsilon_1 - \varepsilon_2 \leq T_{\mathrm{Boot}} \leq t + \varepsilon_1 + \varepsilon_2 \mid \tilde{D}_{\mathsf{infer}}\big) 
        &\leq \sup_{t\in\mathbb{R}} \mP\big(t - \varepsilon_1 - \varepsilon_2 \leq U_0 \leq t + \varepsilon_1 + \varepsilon_2\big) + 2\widetilde\Delta_m^\ast \\
        &\lesssim \sqrt{\bigg( \frac{\varepsilon_1 + \varepsilon_2}{\sqrt{c_1}} \bigg)} + 2\widetilde\Delta_m^\ast,
    \end{align*}
    where the last step applies the anti-concentration bound for $U_0$ (\Cref{lemma:concentration inequality:infinitysum_chi}) using $\sum_{\ell=1}^\infty \lambda_\ell^2 = \mE_P(H_{12}^2) \geq c_1$. Taking expectations over $\tilde{D}_{\mathsf{infer}}$ and applying Markov's inequality to the deviation probabilities of $T_1$ and $T_2$, we obtain for any $s > 0$,
    \begin{align*}
        \mP_P(\mD_m^{\textsf{est}} > s) &\lesssim \frac{1}{s} \bigg\{ \sqrt{(\varepsilon_1 + \varepsilon_2)} + \mE_P(\widetilde\Delta_m^\ast) + \frac{\mE_P(|T_1|)}{\varepsilon_1} + \frac{\mE_P(T_2^2)}{\varepsilon_2^2} \bigg\}.
    \end{align*}
    We choose the auxiliary parameters to balance these terms
    \begin{align*}
        \varepsilon_1 &= \sqrt{m\, \mE_P(\Delta_1^2)}, \qquad 
        \varepsilon_2 = \big\{\mE_P(\Delta_1^2)\big\}^{1/3}.
    \end{align*}
    With these choices, the penalty terms respectively satisfy
    \begin{align*}
        \frac{\mE_P(|T_1|)}{\varepsilon_1} &\lesssim \sqrt{m\, \mE_P(\Delta_1^2)}, \qquad 
        \frac{\mE_P(T_2^2)}{\varepsilon_2^2} \lesssim \big\{\mE_P(\Delta_1^2)\big\}^{1/3}.
    \end{align*}
    Furthermore, by \Cref{lemma:multiplier_bootstrap_consistency}, $\mE_P(\widetilde\Delta_m^\ast) \lesssim \{ \mE_P(H_{11}^2) / (m \mE_P(H_{12}^2)) \}^{1/5} \lesssim m^{-1/5}$ uniformly over $\mathcal{P}_{0}^{\mathsf{mmd}}$. Taking the supremum over $P \in \mathcal{P}_{0}^{\mathsf{mmd}}$ yields
    \begin{align*}
        \sup_{P\in\mathcal{P}_{0}^{\mathsf{mmd}}} \mP_P(\mD_{m}^{\textsf{est}}>s) 
        &\lesssim
        \frac{1}{s}\Bigg(
        \frac{1}{c_{1}^{1/4}}\bigg[\Big\{m \sup_{P\in\mathcal{P}_{0}^{\mathsf{mmd}}} \mE_P(\Delta_1^2)\Big\}^{1/2} + \Big\{\sup_{P\in\mathcal{P}_{0}^{\mathsf{mmd}}} \mE_P(\Delta_1^2)\Big\}^{1/3}\bigg]^{1/2} \\
        &\qquad\qquad + m^{-1/5} + \Big\{m \sup_{P\in\mathcal{P}_{0}^{\mathsf{mmd}}} \mE_P(\Delta_1^2)\Big\}^{1/2} + \Big\{\sup_{P\in\mathcal{P}_{0}^{\mathsf{mmd}}} \mE_P(\Delta_1^2)\Big\}^{1/3}
        \Bigg).
    \end{align*}
    By Assumption~\ref{Assumption : Unified MMD}(b), $\sup_{P} \mE_P(\Delta_1^2) = o((n-m)^{-1/2})$. Furthermore, Assumption~\ref{Assumption : Unified MMD}(d) yields $m/\sqrt{(n-m)} \to \sqrt{c_{5}}$. Together, these imply that $m \sup_{P\in\mathcal{P}_{0}^{\mathsf{mmd}}} \mE_P(\Delta_1^2) \to 0$ and $\sup_{P} \mE_P(\Delta_1^2) \to 0$. Since $m \to \infty$, all terms in the upper bound vanish, establishing $\mD_{m}^{\textsf{est}} = o_{\mathcal{P}_{0}^{\mathsf{mmd}}}(1)$.
    
    \vspace{0.5em}
    Combining $\mD_{m}^{\textsf{oracle}} = o_{\mathcal{P}_{0}^{\mathsf{mmd}}}(1)$ and $\mD_{m}^{\textsf{est}} = o_{\mathcal{P}_{0}^{\mathsf{mmd}}}(1)$, we conclude that for every fixed $\varepsilon > 0$,
    \begin{align*}
        \lim_{n \to \infty}
        \sup_{P \in \mathcal{P}_{0}^{\mathsf{mmd}}}
        \mP_{P}
        \bigg(
        \sup_{t \in \mathbb{R}}
        \Big|
        \mP_{P}\big(
        \widehat{\mathrm{MMD}}^2_{\mathrm{Boot}}
        \leq t \mid \tilde{D}_{\mathsf{infer}}\big)
        -
        \mP_{P}\big(
        \mathrm{MMD}_{1}^2 \leq t\big)
        \Big| > \varepsilon
        \bigg) &= 0.
    \end{align*}
    
    \mypara{Extension to general $K \geq 2$} The preceding analysis establishes the uniform bootstrap consistency for a single fold ($K=1$). We now extend it to $K \geq 2$ via a single application of \Cref{Lemma: kolmogorov subadditive convolution}. For each $j$, let $\widehat{F}_{m,j}^*(x) := \mP_P\big(\widehat{\mathrm{MMD}}_{1,\mathrm{Boot},j}^2 \leq x \mid \tilde{D}_{\mathsf{infer}}\big)$ and $F_{0,\scriptscriptstyle{P}}(x) := \mP_P(G_j \leq x)$. The case $K=1$ established above gives, for every $\varepsilon > 0$ and every $j \in \{1,\ldots,K\}$,
    \begin{align}\label{eq: per fold bootstrap}
        \lim_{n\to\infty} \sup_{P\in\mathcal{P}_0^{\mathsf{mmd}}} \mP_P\bigg(\sup_{x\in\mathbb{R}} |\widehat{F}_{m,j}^*(x) - F_{0,\scriptscriptstyle{P}}(x)| > \varepsilon\bigg) = 0.
    \end{align}
    
    The multiplier collections $\{\xi_{i,j}\}_{i \in [m]}$ for distinct folds $j$ are mutually independent and independent of the data, so $\{\widehat{\mathrm{MMD}}_{1,\mathrm{Boot},j}^2\}_{j=1}^K$ are conditionally mutually independent given $\tilde{D}_{\mathsf{infer}}$. Choose $G_1, \ldots, G_K$ to be i.i.d.\ copies of $G$ that are independent of all data, hence remain mutually independent under any conditioning on $\tilde{D}_{\mathsf{infer}}$. Applying \Cref{Lemma: kolmogorov subadditive convolution} conditionally on $\tilde{D}_{\mathsf{infer}}$,
    \begin{align*}
        \sup_{t\in\mathbb{R}}\bigg|\mP_P\bigg(\sum_{j=1}^K \widehat{\mathrm{MMD}}_{1,\mathrm{Boot},j}^2 \leq t \,\big|\, \tilde{D}_{\mathsf{infer}}\bigg) - \mP_P\bigg(\sum_{j=1}^K G_j \leq t\bigg)\bigg|
        \;\leq\; \sum_{j=1}^K \sup_{t\in\mathbb{R}}|\widehat{F}_{m,j}^*(t) - F_{0,\scriptscriptstyle{P}}(t)|,
    \end{align*}
    where the unconditional CDF $\mP_P(\sum_{j=1}^{K} G_j \leq t)$ on the left coincides with its conditional counterpart given $\tilde{D}_{\mathsf{infer}}$ by the independence of $\{G_j\}$ from the data. The Kolmogorov distance is invariant under the deterministic scaling $x \mapsto x/K$, so writing ${}^{\dagger}\widehat{F}_m^*(t) := \mP_P^*({}^{\dagger}\widehat{\mathrm{MMD}}_{1,\mathrm{Boot}}^2 \leq t)$ and ${}^{\dagger}F_{0,\scriptscriptstyle{P}}(t) := \mP_P\big(K^{-1}\sum_{j=1}^K G_j \leq t\big)$,
    \begin{align}\label{eq: bootstrap kolmogorov subadd}
        \sup_{t\in\mathbb{R}} |{}^{\dagger}\widehat{F}_m^*(t) - {}^{\dagger}F_{0,\scriptscriptstyle{P}}(t)|
        \;\leq\; \sum_{j=1}^K \sup_{x\in\mathbb{R}} |\widehat{F}_{m,j}^*(x) - F_{0,\scriptscriptstyle{P}}(x)|.
    \end{align}
    For any $\varepsilon > 0$, the union bound applied to the right-hand side of \eqref{eq: bootstrap kolmogorov subadd}, together with \eqref{eq: per fold bootstrap}, gives
    \begin{align*}
        \lim_{n\to\infty} \sup_{P\in\mathcal{P}_0^{\mathsf{mmd}}} \mP_P\bigg(\sup_{t\in\mathbb{R}} |{}^{\dagger}\widehat{F}_m^*(t) - {}^{\dagger}F_{0,\scriptscriptstyle{P}}(t)| > \varepsilon\bigg) = 0.
    \end{align*}
    
    By \Cref{cor: Cross-fitted MMD Asymptotic Distribution} (Regime~II), the cross-fitted oracle statistic satisfies $\sup_{t} |\mP_P({}^{\dagger}\mathrm{MMD}_1^2 \leq t) - {}^{\dagger}F_{0,\scriptscriptstyle{P}}(t)| \to 0$ uniformly over $\mathcal{P}_0^{\mathsf{mmd}}$. The triangle inequality combining these two uniform limits concludes the proof of (i).
\end{proof}

\subsection{\texorpdfstring{Proof of Theorem~\ref{thm: Cross-fitted Bootstrap Validity}$(ii)$}{Proof of Theorem 5(ii)}}
\label{Proof of theorem: multiplier-bootstrap-validity part 2}
    
\begin{proof}
    Fix $\alpha\in(0,1)$ and an integer $K \geq 1$. Let $\big\{{}^{\dagger}\widehat{\mathrm{MMD}}^{2,(r)}_{1,\mathrm{Boot}}\big\}_{r=1}^R$ denote the $R$ independent cross-fitted multiplier bootstrap replicates conditionally on $\tilde{D}_{\mathsf{infer}}$. For each $P\in\mathcal{P}_{0}^{\mathsf{mmd}}$, recall that the limiting cross-fitted null distribution and its $(1-\alpha)$-quantile are
    \begin{align*}
        {}^{\dagger}F_{0,\scriptscriptstyle{P}}(t) = \mP_P\bigg(\frac{1}{K}\sum_{j=1}^K G_j \leq t\bigg),
        \qquad
        {}^{\dagger}q_{1-\alpha,\scriptscriptstyle{P}}
        =\inf\big\{t\in\mathbb R :   {}^{\dagger}F_{0,\scriptscriptstyle{P}}(t) \geq 1-\alpha\big\}.
    \end{align*}
    By combining the asymptotic null distribution of the cross-fitted statistic and Theorem~\ref{thm: Cross-fitted Bootstrap Validity}$(i)$, the conditional bootstrap CDF for a single replicate,
    \begin{align*}
        {}^{\dagger}\widehat F_m^*(t):=\mP_{P}^*\big({}^{\dagger}\widehat{\mathrm{MMD}}^{2,(1)}_{1,\mathrm{Boot}} \leq t\big)
        = \mP_{P}\big({}^{\dagger}\widehat{\mathrm{MMD}}^{2,(1)}_{1,\mathrm{Boot}} \leq t\mid \tilde{D}_{\mathsf{infer}}\big),
    \end{align*}
    satisfies, for every $\varepsilon>0$,
    \begin{align*}
        \sup_{P\in \mathcal{P}_{0}^{\mathsf{mmd}}}\mP_{P}\Big(\sup_{t\in\mathbb R}\big|{}^{\dagger}\widehat F_m^*(t)-{}^{\dagger}F_{0,\scriptscriptstyle{P}}(t)\big|>\varepsilon\Big)\to 0,\quad \text{as }n\to \infty.
    \end{align*}
    Define the empirical bootstrap CDF based on the $R$ multiplier replicates by
    \begin{align*}
        {}^{\dagger}\widehat F_m^{R}(t):=\frac{1}{R} \sum_{r=1}^R \mathds{1}\Big({}^{\dagger}\widehat{\mathrm{MMD}}^{2,(r)}_{1,\mathrm{Boot}} \le t\Big).
    \end{align*}
    Conditionally on $\tilde{D}_{\mathsf{infer}}$, the bootstrap replicates are i.i.d.\ with distribution function ${}^{\dagger}\widehat F_m^*$. Hence, by the Dvoretzky--Kiefer--Wolfowitz inequality, for any $\varepsilon>0$,
    \begin{align*}
        \mP_{P}\Big(\sup_{t\in\mathbb R}\big|{}^{\dagger}\widehat F_m^{R}(t)-{}^{\dagger}\widehat F_m^*(t)\big|>\varepsilon\given \tilde{D}_{\mathsf{infer}}\Big)\leq 2e^{-2R\varepsilon^2}.
    \end{align*}
    Taking expectations and noting that the bound depends only on $R$ and is independent of $P$ and $n$, we obtain
    \begin{align*}
        \sup_{P\in\mathcal{P}_{0}^{\mathsf{mmd}}}\mP_{P}\Big(\sup_{t\in\mathbb{R}}\big|{}^{\dagger}\widehat F_m^{R}(t)-{}^{\dagger}\widehat F_m^*(t)\big|>\varepsilon\Big)\leq2e^{-2R\varepsilon^2}\to 0, \quad \text{as } R\to\infty. 
    \end{align*}
    By the triangle inequality, we have
    \begin{align*}
        \sup_{t\in\mathbb{R}}\big|{}^{\dagger}\widehat F_m^{R}(t)-{}^{\dagger}F_{0,\scriptscriptstyle{P}}(t)\big|&\leq\sup_{t\in\mathbb{R}}\big|{}^{\dagger}\widehat F_m^{R}(t)-{}^{\dagger}\widehat F_m^*(t)\big|+\sup_{t\in\mathbb{R}}\big|{}^{\dagger}\widehat F_m^*(t)-{}^{\dagger}F_{0,\scriptscriptstyle{P}}(t)\big|.
    \end{align*}
    Hence, for any $\varepsilon>0$, by the union bound,
    \begin{align*}
        &\mP_{P}\Big(\sup_{t \in \mathbb{R}}\big|{}^{\dagger}\widehat F_m^{R}(t)-{}^{\dagger}F_{0,\scriptscriptstyle{P}}(t)\big|> \varepsilon\Big) \\
        &\leq\mP_{P}\Big(\sup_{t \in \mathbb{R}}\big|{}^{\dagger}\widehat F_m^{R}(t)-{}^{\dagger}\widehat F_m^*(t)\big|>\varepsilon/2\Big)+\mP_{P}\Big(\sup_{t \in \mathbb{R}}\big|{}^{\dagger}\widehat F_m^*(t)-{}^{\dagger}F_{0,\scriptscriptstyle{P}}(t)\big|>\varepsilon/2\Big).
    \end{align*}
    Taking the supremum over $P \in \mathcal{P}_{0}^{\mathsf{mmd}}$, the first term vanishes as $R \to \infty$ uniformly in $n$, and the second term vanishes as $n \to \infty$. Thus, taking the limits sequentially yields
    \begin{align*}
        \lim_{n,R \to \infty} \sup_{P\in \mathcal{P}_{0}^{\mathsf{mmd}}}\mP_{P}\Big(\sup_{t\in \mathbb{R}}\big|{}^{\dagger}\widehat F_m^{R}(t)-{}^{\dagger}F_{0,\scriptscriptstyle{P}}(t)\big|>\varepsilon\Big)= 0.
    \end{align*}
    Fix $\tau>0$ and define
    \begin{align*}
        \kappa(\tau):=\inf_{P\in \mathcal{P}_{0}^{\mathsf{mmd}}}\inf_{|t-q_{1-\alpha,\scriptscriptstyle{P}}^{\dagger}|\geq \tau}\big|{}^{\dagger}F_{0,\scriptscriptstyle{P}}(t)-(1-\alpha)\big|.
    \end{align*}
    By the assumed uniform separation condition in Theorem~\ref{thm: Cross-fitted Bootstrap Validity}$(ii)$, $\kappa(\tau)>0$ for every $\tau>0$. Define the event
    \begin{align*}
        A_{n,R}:=\Big\{\sup_{t\in\mathbb{R}}\big|{}^{\dagger}\widehat F_m^{R}(t)-{}^{\dagger}F_{0,\scriptscriptstyle{P}}(t)\big| < \kappa(\tau)/2\Big\}.
    \end{align*}
    By the preceding uniform CDF convergence,
    \begin{align*}
        \lim_{n,R \to \infty} \sup_{P\in \mathcal{P}_{0}^{\mathsf{mmd}}}\mP_{P}(A_{n,R}^c)= 0.
    \end{align*}
    It remains to show that on $A_{n,R}$ we have $|{}^{\dagger}\widehat q_{1-\alpha}^{R}-q_{1-\alpha,\scriptscriptstyle{P}}^{\dagger}|<\tau$. We proceed by contradiction. Suppose that ${}^{\dagger}\widehat q_{1-\alpha}^{R}> q_{1-\alpha,\scriptscriptstyle{P}}^{\dagger}+\tau$. Then, by the definition of the empirical $(1-\alpha)$-quantile,
    \begin{align*}
        {}^{\dagger}\widehat F_m^{R}\big(^{\dagger}q_{1-\alpha,\scriptscriptstyle{P}}+\tau\big) < 1-\alpha.
    \end{align*}
    Hence
    \begin{align*}
        {}^{\dagger}F_{0,\scriptscriptstyle{P}}(q_{1-\alpha,\scriptscriptstyle{P}}^{\dagger}+\tau)-{}^{\dagger}\widehat F_m^{R}(q_{1-\alpha,\scriptscriptstyle{P}}^{\dagger}+\tau) > {}^{\dagger}F_{0,\scriptscriptstyle{P}}(q_{1-\alpha,\scriptscriptstyle{P}}^{\dagger}+\tau)-(1-\alpha).
    \end{align*}
    Since ${}^{\dagger}F_{0,\scriptscriptstyle{P}}(q_{1-\alpha,\scriptscriptstyle{P}}^{\dagger}) \geq 1-\alpha$ by the definition of $^{\dagger}q_{1-\alpha,\scriptscriptstyle{P}}$ and ${}^{\dagger}F_{0,\scriptscriptstyle{P}}$ is non-decreasing, the right-hand side is at least $\kappa(\tau)$, which contradicts the event $A_{n,R}$. The case ${}^{\dagger}\widehat q_{1-\alpha}^{R} < q_{1-\alpha,\scriptscriptstyle{P}}^{\dagger}-\tau$ is treated analogously. Consequently,
    \begin{align*}
        \lim_{n,R \to \infty} \sup_{P\in \mathcal{P}_{0}^{\mathsf{mmd}}}\mP_{P}\Big(\big|{}^{\dagger}\widehat q_{1-\alpha}^{R}-q_{1-\alpha,\scriptscriptstyle{P}}^{\dagger}\big|>\tau\Big)=0.
    \end{align*}
    Let ${}^{\dagger}F_{m,\scriptscriptstyle{P}}(t):=\mP_{P}\big({}^{\dagger}\widehat{\mathrm{MMD}}_1^2 \le t\big)$ denote the exact finite-sample CDF of the cross-fitted test statistic. By the uniform distributional convergence of the cross-fitted statistic,
    \begin{align*}
        \sup_{P\in \mathcal{P}_{0}^{\mathsf{mmd}}}\sup_{t \in \mathbb{R}}\big|{}^{\dagger}F_{m,\scriptscriptstyle{P}}(t)-{}^{\dagger}F_{0,\scriptscriptstyle{P}}(t)\big|\to 0, \quad \text{as }n \to \infty.
    \end{align*}
    The rejection probability equals
    \begin{align*}
        \mP_{P}\big({}^{\dagger}\phi_\alpha^{R}=1\big)=\mP_{P}\big({}^{\dagger}\widehat{\mathrm{MMD}}_1^2>{}^{\dagger}\widehat q_{1-\alpha}^{R}\big).
    \end{align*}
    For any $\tau>0$, the rejection probability can be bounded from above and below by
    \begin{align*}
        &\mP_{P}\big({}^{\dagger}\widehat{\mathrm{MMD}}_1^2>q_{1-\alpha,\scriptscriptstyle{P}}^{\dagger}+\tau\big)-\mP_{P}\big(|{}^{\dagger}\widehat q_{1-\alpha}^{R}-q_{1-\alpha,\scriptscriptstyle{P}}^{\dagger}|>\tau\big) \\
        &\quad \leq
        \mP_{P}\big({}^{\dagger}\widehat{\mathrm{MMD}}_1^2>{}^{\dagger}\widehat q_{1-\alpha}^{R}\big) \\
        &\quad \leq
        \mP_{P}\big({}^{\dagger}\widehat{\mathrm{MMD}}_1^2>q_{1-\alpha,\scriptscriptstyle{P}}^{\dagger}-\tau\big)+\mP_{P}\big(|{}^{\dagger}\widehat q_{1-\alpha}^{R}-q_{1-\alpha,\scriptscriptstyle{P}}^{\dagger}|>\tau\big).
    \end{align*}
    Using the relation $\mP_{P}\big({}^{\dagger}\widehat{\mathrm{MMD}}_1^2 > t\big) = 1-{}^{\dagger}F_{m,\scriptscriptstyle{P}}(t)$ and replacing ${}^{\dagger}F_{m,\scriptscriptstyle{P}}$ by ${}^{\dagger}F_{0,\scriptscriptstyle{P}}$ up to a uniformly vanishing error as $n \to \infty$, we obtain
    \begin{align*}
        \limsup_{n,R \to \infty}\sup_{P\in \mathcal{P}_{0}^{\mathsf{mmd}}}\Big|\mP_{P}\big({}^{\dagger}\phi_\alpha^{R}=1\big)-\alpha\Big|
        \leq \sup_{P\in \mathcal{P}_{0}^{\mathsf{mmd}}}\sup_{|t-q_{1-\alpha,\scriptscriptstyle{P}}^{\dagger}|\le\tau}\big|{}^{\dagger}F_{0,\scriptscriptstyle{P}}(t)-(1-\alpha)\big|.
    \end{align*}
    It remains to show that the right-hand side can be made arbitrarily small. By Lemma~\ref{lemma:concentration inequality:infinitysum_chi}, the limiting distribution $G_1$ satisfies the anti-concentration bound. Because the convolution of independent random variables strictly preserves bounded densities, the cross-fitted limit $\frac{1}{K}\sum_{j=1}^K G_j$ inherits this anti-concentration property. Thus, for any $\varpi>0$,
    \begin{align*}
        \sup_{t\in\mathbb R} \mP_{P}\bigg( t < \frac{1}{K}\sum_{j=1}^K G_j \leq t+\varpi \bigg) \lesssim \sqrt{{\frac{\varpi}{\sum_{\ell=1}^{\infty}\lambda_{\ell}^2}}}.
    \end{align*}
    Using the definition of the limiting CDF ${}^{\dagger}F_{0,\scriptscriptstyle{P}}$ and the triangle inequality, we bound the local variation around the quantile:
    \begin{align*}
        \sup_{|t-q_{1-\alpha,\scriptscriptstyle{P}}^{\dagger}|\le\tau} \big| {}^{\dagger}F_{0,\scriptscriptstyle{P}}(t)-(1-\alpha) \big|
        &\leq \sup_{s\in\mathbb R} \big| {}^{\dagger}F_{0,\scriptscriptstyle{P}}(s+\tau)-{}^{\dagger}F_{0,\scriptscriptstyle{P}}(s) \big| \\
        &= \sup_{s\in\mathbb R} \mP_{P}\bigg( s < \frac{1}{K}\sum_{j=1}^K G_j \leq s+\tau \bigg) \\
        &\lesssim \sqrt{{\frac{\tau}{\mE_{P}(H_{12}^2)}}}.
    \end{align*}
    Under Assumption~\ref{Assumption : Unified MMD}(a), $\mE_{P}(H_{12}^2) \geq c_1 > 0$ uniformly over $P \in \mathcal{P}_{0}^{\mathsf{mmd}}$. Combining with the preceding bound,
    \begin{align*}
        \limsup_{n,R \to \infty} \sup_{P\in \mathcal{P}_{0}^{\mathsf{mmd}}} \Big| \mP_{P} \big({}^{\dagger}\phi_\alpha^{R}=1 \big) - \alpha \Big| 
        \lesssim \sqrt{{\frac{\tau}{c_{1}}}}.
    \end{align*}
    Since the left-hand side does not depend on $\tau$, letting $\tau\downarrow0$ yields the desired conclusion.
\end{proof}

\subsection{\texorpdfstring{Proof of \Cref{Example : Stable case}}{Proof of Example : Stable case}}\label{Appendix : Proof of Example : Stable case}
\begin{proof}
    We establish the asymptotic equivalence between the GCM statistic $T$ and its counterpart $\tilde{T}$ constructed using $\{(\tilde{X}_i,\tilde{Y}_i,\tilde{Z}_i)\}_{i=1}^n$. Throughout, we assume that $Y$ has a finite second moment and that $Y \independent Z \given X$, i.e., the null hypothesis holds. Let
    \begin{align*}
        T = \frac{\frac{1}{\sqrt{n}}\sum_{i=1}^n R_i}{\sqrt{\big\{ \frac{1}{n} \sum_{i=1}^n R_i^2 - \bigl(\frac{1}{n} \sum_{r=1}^n R_r\bigr)^2  \big\}}} =  \frac{\nu_{R}}{\hat{\sigma}_{R}},
    \end{align*}
    and $\tilde{T} =  {\nu}_{\tilde{R}}/\hat{\sigma}_{\tilde{R}}$. Let $\sigma_{\tilde{R}}^2 > 0$ denote the variance of $\{\tilde{Y} - f(\tilde{X})\}\{\tilde{Z} - g(\tilde{X})\}$ where $(\tilde{X},\tilde{Y},\tilde{Z}) $ is a random draw from the joint distribution $P_{XYZ}$. We begin with an upper bound for $|T - \tilde{T}|$: 
    \begin{align*}
        \Big|\frac{\nu_{R}}{\hat{\sigma}_{R}}-\frac{{\nu}_{\tilde{R}}}{\hat{\sigma}_{\tilde{R}}} \Big| &\leq
        \Big|\frac{\nu_{R}}{\hat{\sigma}_{R}}-\frac{{\nu}_{\tilde{R}}}{\hat{\sigma}_{R}}\Big|+\Big| \frac{{\nu}_{\tilde{R}}}{\hat{\sigma}_{R}}-\frac{{\nu}_{\tilde{R}}}{\hat{\sigma}_{\tilde{R}}} \Big| \\
        &\leq \frac{1}{\hat{\sigma}_{R}} \lvert \nu_{R}-{\nu}_{\tilde{R}}\rvert + \frac{\lvert {\nu}_{\tilde{R}}\rvert}{( \hat{\sigma}_{R}+\hat{\sigma}_{\tilde{R}})\hat{\sigma}_{R}\hat{\sigma}_{\tilde{R}}}\big| \hat{\sigma}_{R}^2-\hat{\sigma}_{\tilde{R}}^2\big|,
    \end{align*}
    from which the proof reduces to showing the convergence of the following four terms to zero in probability: (a) $\nu_R - \nu_{\tilde{R}}$, (b) $\hat{\sigma}_{R}^2-\hat{\sigma}_{\tilde{R}}^2$, (c) $\hat{\sigma}_R^2 - \sigma_{\tilde{R}}^2$ and (d) $\hat{\sigma}_{\tilde{R}}^2 - \sigma_{\tilde{R}}^2$. Under these convergence results, the asymptotic equivalence follows by the continuous mapping theorem along with the fact that $\nu_{\tilde{R}}$ is stochastically bounded by the central limit theorem. In what follows, we establish convergence of (a), (b), (c), and (d) to zero in probability in order. 
    \begin{enumerate}[itemsep=0.5em, parsep=5pt, label=(\alph*)]
        \item \textit{$\nu_R - \nu_{\tilde{R}}$. } The difference between $\nu_R$ and $\nu_{\tilde{R}}$ can be written as
        \begin{align*}
            {\frac{1}{\sqrt{n}} \sum_{i=1}^n (R_i - \tilde{R}_i}) = \underbrace{\frac{1}{\sqrt{n}} \sum_{i= n_1 + 1}^{\bar{n}_1} (R_i - \tilde{R}_i) \; \mathds{1}(\bar{n}_1 > n_1)}_{= \Delta_{1}} + \underbrace{\frac{1}{\sqrt{n}} \sum_{i= \bar{n}_1 + 1}^{n_1} (R_i - \tilde{R}_i) \; \mathds{1}(\bar{n}_1 \leq n_1)}_{=  \Delta_{2}}.
        \end{align*}
        Note that $Z_i$ is a fixed constant for a given index $i$, which allows us to show that $\mE(R_i) = 0$ for any $i \in [n]$. For example, when $i=1$, $Z_1$ equals $1$ (since $X_{1} = X_{1}^{(1)}$) and thus the law of total expectation yields
        \begin{align*}
            \mE(R_1) = \mE[\{1-g(X_{1})\}\{Y_{1}-f(X_{1})\}] =\mE\big[\{1-g(X_{1})\}\, \mE[\{Y_{1}-f(X_{1})\} \given X_1]\big] = 0,
        \end{align*}
        where we recall $f(X_1) = \mE(Y_1\given X_1)$. It also follows that $\mE(\tilde{R}_i) = 0$ for any $i \in [n]$ under the null hypothesis. This, together with the law of total expectation, shows that 
        \begin{align*}
            \mE{(\Delta_1)} = \mE{\Bigg\{\frac{1}{\sqrt{n}}\sum_{i=n_{1}+1}^{\bar{n}_1}\mathds{1}(\bar{n}_1 > n_1)\mE{\big( R_i - \tilde{R}_i\given \bar{n}_1\big)} \Bigg\}} = 0,
        \end{align*}
        and similarly $\mE{(\Delta_2)} = 0$. Thus, we have $\mE(\nu_R - \nu_{\tilde{R}}) = 0$. 
        
        Now consider the variance of $\nu_R - \nu_{\tilde{R}}$. Since $\mE(\nu_R - \nu_{\tilde{R}}) = 0$ and $\mE(\Delta_1\Delta_2) =0$, we have 
        \begin{align*}
            \mV(\nu_R - \nu_{\tilde{R}}) = \mV(\Delta_1) + \mV(\Delta_2). 
        \end{align*} 
        For $\mV(\Delta_1)$, we have
        \begin{align*}
            \mV(\Delta_1) & =\mE\bigl\{\mV\big(\Delta_1 \given \bar{n}_1\big)\big\}+\mV\big\{\underbrace{\mE\big(\Delta_1 \given \bar{n}_1\big)}_{=0}\big\} \\
            &=
            \mE{\bigg[
                \operatorname{var}\bigg\{\frac{1}{\sqrt{n}} \sum_{i= n_1 + 1}^{\bar{n}_1} (R_i - \tilde{R}_i)\; \mathds{1}(\bar{n}_1 > n_1) \;\Big|\; \bar{n}_1\bigg\}
                \bigg]} \\
            &=
            \mE{\bigg[\frac{1}{n}\sum_{i= n_1 + 1}^{\bar{n}_1}\mathds{1}(\bar{n}_1 > n_1)\;\mE{\big\{\big(R_{i}-\tilde{R}_{i} \big)^2 \; \big|\; \bar{n}_1\big\}} \bigg]}\\
            & \overset{\mathrm{(i)}}{\leq} 2 
            \mE{\bigg\{\frac{\bar{n}_1-n_1}{n}\;\frac{1}{\bar{n}_1-n_1}\sum_{i=n_1+1}^{\bar{n}_1}\mE{\bigl(R_{i}^2+\tilde{R}_{i}^2 \given \bar{n}_1 \bigr)}\bigg\}} \\
            & \overset{\mathrm{(ii)}}{\leq} 
            \frac{4}{n} \mE\left(\left|\bar{n}_1-n_1\right|\right) \mV(Y_1)\\ 
            &\overset{\mathrm{(iii)}}{\leq} \frac{2}{\sqrt{n}} \mV(Y_1), 
        \end{align*}
        where $\mathrm{(i)}$ follows from the inequality $(x - y)^2 \leq 2x^2 + 2y^2$ and $\mathrm{(ii)}$ uses the law of total variance along with the fact that $R_i^2 \leq \{Y_i - f(X_i)\}^2$ and $\tilde{R}_i^2 \leq \{\tilde{Y}_i - f(\tilde{X}_i)\}^2$ since $Z_i, \tilde{Z}_i \in \{1,2\}$. For the last inequality $\mathrm{(iii)}$, we use $\mE(|\bar{n}_1 - n_1|) \leq \sqrt{n}/2$. The same bound holds for $\mV(\Delta_2)$ and thus 
        \begin{align*}
            \mV(\nu_R - \nu_{\tilde{R}}) \leq \frac{4}{\sqrt{n}} \mV(Y_1). 
        \end{align*}
        Combining the results with Chebyshev's inequality now shows that $\nu_R - \nu_{\tilde{R}}$ converges to zero in probability.

        \item \textit{$\hat{\sigma}_{R}^2-\hat{\sigma}_{\tilde{R}}^2$.} We show that 
        \begin{align*}
            \hat{\sigma}_{R}^2-\hat{\sigma}_{\tilde{R}}^2 = \bigg\{ \frac{1}{n} \sum_{i=1}^n R_i^2 - \Big(\frac{1}{n} \sum_{r=1}^n R_r\Big)^2 \bigg\} - \bigg\{ \frac{1}{n} \sum_{i=1}^n \tilde{R}_i^2 - \Big(\frac{1}{n} \sum_{r=1}^n \tilde{R}_r\Big)^2  \bigg\}
        \end{align*}
        converges to zero in probability. We decompose this into two terms
        \begin{align*}
            \mathbb{(I)} =  \frac{1}{n} \sum_{i=1}^n R_i^2 - \frac{1}{n} \sum_{i=1}^n \tilde{R}_i^2 \quad \text{and} \quad
            \mathbb{(II)}  =  \Big(\frac{1}{n} \sum_{r=1}^n R_r\Big)^2 - \Big(\frac{1}{n} \sum_{r=1}^n \tilde{R}_r\Big)^2,
        \end{align*}
        and show each converges to zero in probability. For the first term $\mathbb{(I)}$, we have
        \begin{align*}
            \frac{1}{n} \sum_{i=1}^n (R_i^2 - \tilde{R}_i^2) = \underbrace{\frac{1}{n}\sum_{n_{1}+1}^{\bar{n}_1}(R_{i}^2-\tilde{R}_{i}^2)\;\mathds{1}(\bar{n}_1 > n_{1})}_{=  \tilde{\Delta}_1}+\underbrace{\frac{1}{n}\sum_{\bar{n}_1+1}^{n_{1}}(R_{i}^2-\tilde{R}_{i}^2)\;\mathds{1}(\bar{n}_1 \leq n_{1})}_{=  \tilde{\Delta}_2}.
        \end{align*}
        Using the law of total expectation, we obtain
        \begin{align*}
            \mE(|\tilde{\Delta}_1|)  &= \mE{\bigg[\Big|\frac{1}{n}\sum_{i=n_{1}+1}^{\bar{n}_1}(R_{i}^2-\tilde{R}_{i}^2)\;\mathds{1}(\bar{n}_1 > n_{1})\Big|\bigg]} \\
            &\leq \mE{\bigg\{\frac{\bar{n}_1-n_1}{n}\;\frac{1}{\bar{n}_1-n_1}\sum_{i=n_{1}+1}^{\bar{n}_1} \mathds{1}(\bar{n}_1 > n_1) \; \mE\bigl(\lvert R_{i}^2 - \tilde{R}_{i}^2 \rvert \given \bar{n}_1 \bigr) \bigg\}} \\
            &\leq \frac{2}{n} \mE(|\bar{n}_1 - n_1|) \mV(Y_1)  \leq \frac{\mV(Y_1)}{\sqrt{n}},
        \end{align*}
        where the first inequality is derived from Jensen's inequality, and the remaining steps follow from the previous results. A similar argument applies to $\tilde{\Delta}_2$. By Markov's inequality, $\mathbb{(I)}$ converges to zero in probability.
        For the second term $\mathbb{(II)}$, we have
        \begin{align*}
            \bigg\lvert
            \Big(\frac{1}{n}\sum_{r=1}^n R_r\Big)^2 - \Big(\frac{1}{n} \sum_{r=1}^n \tilde{R}_r\Big)^2
            \bigg\rvert \leq \bigg\lvert \frac{1}{n} \sum_{r=1}^n R_r + \frac{1}{n} \sum_{r=1}^n \tilde{R}_r \bigg\rvert\;\bigg\lvert\frac{1}{n} \sum_{r=1}^n R_r - \frac{1}{n} \sum_{r=1}^n \tilde{R}_r\bigg\rvert,
        \end{align*}
        which converge to zero in probability by the previous results. 
        \item \textit{$\hat{\sigma}_R^2 - \sigma_{\tilde{R}}^2$ and $\hat{\sigma}_{\tilde{R}}^2 - \sigma_{\tilde{R}}^2$.}  
        The latter converges to zero in probability by the law of large numbers. For the former, write 
        $\hat{\sigma}_R^2 - \sigma_{\tilde{R}}^2 = (\hat{\sigma}_R^2 - \hat{\sigma}_{\tilde{R}}^2) + (\hat{\sigma}_{\tilde{R}}^2 - \sigma_{\tilde{R}}^2)$, where the first difference converges to zero by part~(b) and the second by the law of large numbers.
    \end{enumerate}
    This completes the proof which shows that $T$ and $\tilde{T}$ are asymptotically equivalent for the stable case.
\end{proof}

\subsection{\texorpdfstring{Proof of Example~\ref{Example : Unstable case}}{Proof of Example : Unstable case}}\label{Appendix : Proof of Example : Unstable case}

\begin{proof}
    For simplicity, we let $Z \in \{0,1\}$ (rather than $\{1,2\}$ as in the main text), with $Z=1$ for the first group and $Z=0$ for the second, and apply the same convention to $\tilde Z$. We write $\stackrel{p}{\to}$ and $\xrightarrow{d}$ for convergence in probability and in distribution, respectively.

    We show that $T_n$ under the fixed group-size design is not coupling-stable with its i.i.d.\ counterpart $\tilde{T}_n$, by interpreting the fixed design as conditioning the i.i.d.\ statistic on the group count and comparing their asymptotic distributions. Define
    \begin{align*}
        \tilde T_n &= \frac{1}{\sqrt{n}}\sum_{i=1}^n \tilde{\zeta}_i, \qquad 
        \tilde V_n = \frac{1}{\sqrt{n}}\sum_{i=1}^n (\tilde Z_i - p),
    \end{align*}
    where $\tilde{\zeta}_i = \tilde Y_i(\tilde Z_i - g(\tilde X_i))$ and $p = \mP(\tilde Z_1=1)$. Assuming $\mE(|\tilde\zeta_1|^{2+\delta})<\infty$ for some $\delta>0$, the multivariate central limit theorem implies that the joint distribution converges to a bivariate normal distribution
    \begin{align*}
        \binom{\tilde T_n}{\tilde V_n} \;\xrightarrow{d}\; \binom{\tilde T}{\tilde V} \sim N(\mathbf{0}, \Sigma),
    \end{align*}
    where the covariance matrix $\Sigma$ is given by
    \begin{align*}
        \Sigma = \begin{pmatrix}
            \mV(\tilde{\zeta}_1) & \mathrm{Cov}(\tilde{\zeta}_1,\tilde Z_1-p) \\
            \mathrm{Cov}(\tilde{\zeta}_1,\tilde Z_1-p) & p(1-p)
        \end{pmatrix}.
    \end{align*}
    Consequently, the unconditional limit of the i.i.d. statistic is $\tilde T_n \xrightarrow{d} N\bigl(0, \mV(\tilde{\zeta}_1)\bigr)$.
    
    Under the fixed group-size design, the data are generated as an i.i.d.\ sample subject to the constraint $\sum_{i=1}^n \tilde Z_i = n_1$. Assuming $n_1/n \to p$, this constraint is equivalent to $\tilde V_n = a_n$, where $a_n = \frac{n_1 - np}{\sqrt{n}} \to 0$. Since $T_n$ uses the identical functional form as $\tilde T_n$, its finite-sample distribution coincides with the conditional distribution of the i.i.d.\ statistic: 
    \begin{align*}
        T_n \;\overset{d}{=}\; \bigg(\tilde T_n \;\big|\; \sum_{i=1}^n \tilde Z_i = n_1\bigg) \;\overset{d}{=}\; (\tilde T_n \mid \tilde V_n = a_n).
    \end{align*}
    We must be careful in taking the asymptotic limit because $\mP(\tilde V_n = a_n) \to 0$ as $n \to \infty$. To rigorously justify conditioning on this shrinking event, we invoke the conditional limit theorem for random vectors with a discrete component \citep[Theorem 2]{Holst1979twoconditionallimit}. The requisite regularity conditions are readily satisfied: the unnormalized sum $\sum_{i=1}^n \tilde Z_i$ is a sufficient statistic for $p$, the Bernoulli variable $\tilde Z_1$ satisfies the characteristic function bound $\sup_{\varepsilon \le |t| \le \pi} |\mE(e^{it\tilde Z_1})| < 1$, and the joint distribution converges to a Gaussian limit. Consequently, the exact conditional distribution of $T_n$ weakly converges to the conditional distribution of its continuous limit, yielding $T_n \xrightarrow{d} (\tilde T \mid \tilde V = 0)$.
    
    For a bivariate normal vector, this conditional distribution is Gaussian with variance reduced by the projection onto $\tilde V$.
    \begin{align*}
        T_n \;\xrightarrow{d}\; N\bigg( 0,\; \mV(\tilde{\zeta}_1) - \frac{\mathrm{Cov}(\tilde{\zeta}_1,\tilde Z_1-p)^2}{p(1-p)} \bigg).
    \end{align*}
    
    Coupling stability in the sense of Definition~\ref{def:stability} requires $|T_n - \tilde T_n| \xrightarrow{p} 0$. If this convergence in probability holds, then $T_n - \tilde T_n = o_p(1)$, and hence $T_n$ and $\tilde T_n$ must share the same marginal weak limit. We now evaluate the covariance term. Under the null hypothesis $Y \independent Z \mid X$, we have $\tilde Y_1 \independent \tilde Z_1 \mid \tilde X_1$. Using the law of total expectation, we obtain the following expansion.
    \begin{align*}
        \mathrm{Cov}(\tilde{\zeta}_1, \tilde Z_1-p) 
        &= \mE\{\tilde Y_1(\tilde Z_1 - g(\tilde X_1))(\tilde Z_1 - p)\} \\
        &= \mE \big[ \mE(\tilde Y_1 \mid \tilde X_1) \cdot \mE\{(\tilde Z_1 - g(\tilde X_1))(\tilde Z_1 - p) \mid \tilde X_1\} \big] \\
        &= \mE \big\{ f(\tilde X_1) g(\tilde X_1) (1 - g(\tilde X_1)) \big\}.
    \end{align*}
    Whenever $\mE\{f(\tilde X_1)g(\tilde X_1)(1-g(\tilde X_1))\} \neq 0$, the asymptotic variance of $T_n$ is strictly smaller than that of $\tilde T_n$. Because their marginal asymptotic distributions differ, the necessary condition $|T_n - \tilde{T}_n|\stackrel{p}{\to}0$ therefore fails, and $T_n$ is not coupling-stable. 
\end{proof}

\section{Supporting Lemmas} \label{Section: Supporting Lemmas}
In this section, we gather several lemmas from the existing literature, some of which we prove or adapt as they are employed in our framework. The proof of the following lemma can be found, for example, in \citet{Mulzer2018FivePO}.
\begin{lemma} \label{Lemma: Concentration inequality}
    Let $Z_1,\ldots,Z_n$ be i.i.d.~Bernoulli random variables with success probability $p \in [0,1]$ and $S_n = \sum_{i=1}^n Z_i$. For any $\delta \in [0,1]$, it holds that 
    \begin{align*}
        & \mP\{S_n \geq (1 + \delta) np\} \leq e^{- \frac{np \delta^2}{3}} \quad \text{and} \quad \mP\{S_n \leq (1 - \delta) np \} \leq e^{- \frac{np \delta^2}{3}}.
    \end{align*}
\end{lemma}
The following is the uniform central limit theorem result in \citet[][Lemma 18]{shah2020hardness}.
\begin{lemma} (\citealp[][Lemma 18]{shah2020hardness}) \label{Lemma: Uniform CLT}
    Let $\mathcal{P}$ be a family of distributions for a random variable $\zeta \in \mathbb{R}$ and suppose that $\zeta_1,\zeta_2,\ldots$ are i.i.d.~copies of $\zeta$. For each $n \in \mathbb{N}$, let $S_n \coloneqq n^{-1/2} \sum_{i=1}^n \zeta_i$. Suppose that for all $P \in \mathcal{P}$, we have $\mE_P(\zeta) = 0$, $\mE_P(\zeta^2)= 1$ and $\mE_P(|\zeta|^{2+\eta}) < c$ for some $\eta,c>0$. We have that 
    \begin{align*}
        \lim_{n \rightarrow \infty} \sup_{P \in \mathcal{P}} \sup_{t \in \mathbb{R}} \big| \mP_P (S_n \leq t) - \Phi(t) \big| = 0.
    \end{align*} 
\end{lemma}

The next lemma corresponds to \citet[][Lemma S8]{lundborg2024projected} on conditional uniform central limit theorem. 
\begin{lemma} (\citealp[][Lemma S8]{lundborg2024projected}) \label{Lemma: conditional clt}
    Let $(X_{n, i})_{n \in \mathbb{N}, i \in [n]}$ be a triangular array of real-valued random variables and let $(\mathcal{F}_n)_{n \in \mathbb{N}}$ be a filtration on $\mathcal{F}$. Assume that
    \begin{enumerate}
        \item $X_{n, 1}, \dots, X_{n, n}$ are conditionally independent given $\mathcal{F}_n$, for each $n \in \mathbb{N}$;
        \item $\mE_P(X_{n, i} \given \mathcal{F}_n) = 0$ for all $n \in \mathbb{N}, i \in [n]$;
        \item $\bigl| n^{-1} \sum_{i=1}^n \mE_P(X_{n, i}^2 \given \mathcal{F}_n) - 1\bigr| = o_{\mathcal{P}}(1)$;
        \item there exists $\delta > 0$ such that
        \begin{align*}
            \frac{1}{n}\sum_{i=1}^n \mE_{P}\bigl(|X_{n, i}|^{2+\delta} \given \mathcal{F}_n\bigr) = o_{\mathcal{P}}(n^{\delta/2}).
        \end{align*}
    \end{enumerate}
    Then $S_n = n^{-1/2} \sum_{m=1}^n X_{n,m}$ converges uniformly in distribution to $N(0, 1)$, i.e. 
    \begin{align*}
        \lim_{n \to \infty} \sup_{P \in \mathcal{P}} \sup_{x \in \mathbb{R}} |\mP_P(S_n \leq x) - \Phi(x)| = 0.
    \end{align*}
\end{lemma}

The next lemma corresponds to \citet[][Lemma 20]{shah2020hardness} on uniform Slutsky's theorem. 
\begin{lemma} (\citealp[][Lemma 20]{shah2020hardness}) \label{Lemma: uniform Slutsky}
    Let $\mathcal{P}$ be a family of distributions that determines the law of a sequences $\left(V_n\right)_{n \in \mathbb{N}}$ and $\left(W_n\right)_{n \in \mathbb{N}}$ of random variables. Suppose
    
    \begin{align*}
        \lim _{n \rightarrow \infty} \sup _{P \in \mathcal{P}} \sup _{t \in \mathbb{R}}\left|\mP_P\left(V_n \leq t\right)-\Phi(t)\right|=0.
    \end{align*}
    
    \noindent Then we have the following. 
    \begin{itemize}
        \item[(a)] If $W_n=o_{\mathcal{P}}(1)$, we have $\lim _{n \rightarrow \infty} \sup _{P \in \mathcal{P}} \sup _{t \in \mathbb{R}}\left|\mP_P\left(V_n+W_n \leq t\right)-\Phi(t)\right|=0$.
        \item[(b)] If $W_n=1+o_{\mathcal{P}}(1)$, we have $\lim _{n \rightarrow \infty} \sup _{P \in \mathcal{P}} \sup _{t \in \mathbb{R}}\left|\mP_P\left(V_n / W_n \leq t\right)-\Phi(t)\right|=0$.
    \end{itemize}
\end{lemma}

\medskip

The next lemma states that the Kolmogorov distance between two independent convolutions is bounded by the sum of the pairwise Kolmogorov distances. We use it to handle the cross-fitted analyses in \Cref{Proof of theorem: MMD Asymptotic Distribution} and \Cref{Proof of theorem: multiplier-bootstrap-validity part 1}, where finitely many independent (or conditionally independent) statistics are aggregated.

\begin{lemma}[{Kolmogorov subadditivity; \citealp[][Lemma~5.1]{mattner2024convolution}}]
    \label{Lemma: kolmogorov subadditive convolution}
    
    Let $K \in \mathbb{N}$ and let $\{Z_j\}_{j=1}^K$ and $\{W_j\}_{j=1}^K$ be two collections of real-valued random variables defined on a common probability space. Suppose that, within each collection, the $K$ variables are mutually independent. Then
    \begin{align}\label{eq: kolmogorov subadditive}
        \sup_{t\in\mathbb{R}}\bigg|\mP\bigg(\sum_{j=1}^K Z_j \leq t\bigg) - \mP\bigg(\sum_{j=1}^K W_j \leq t\bigg)\bigg|
        \;\leq\;
        \sum_{j=1}^K \sup_{t\in\mathbb{R}}\big|\mP(Z_j \leq t) - \mP(W_j \leq t)\big|.
    \end{align}
    Moreover, the same inequality holds when both probabilities are replaced by conditional probabilities given a common $\sigma$-field $\mathcal{G}$, provided that, within each of $\{Z_j\}_{j=1}^K$ and $\{W_j\}_{j=1}^K$, the variables are mutually independent conditionally on $\mathcal{G}$.
    
\end{lemma}
The unconditional inequality is the case of \citet[][Lemma~5.1]{mattner2024convolution} corresponding to the translation-invariant test class $\mathcal{F} = \{\mathds{1}_{(-\infty,t]} \,:\, t \in \mathbb{R}\}$, for which $\|P-Q\|_{\mathcal{F}}$ coincides with the Kolmogorov distance. The conditional version follows by applying the unconditional inequality fiberwise to the regular conditional joint laws of $(Z_1,\ldots,Z_K)$ and $(W_1,\ldots,W_K)$ given $\mathcal{G}$, which factorize as products of conditional marginals under the stated conditional independence.

\medskip

The following two lemmas provide moment bounds for canonical $U$-statistics. Lemma~\ref{Lemma : moment bound U stat} gives a general inequality for decoupled kernels \citep{gine2000exponential}, while Lemma~\ref{Lemma : moment bound T1} specializes this to the block statistic $T_{1,1}$, showing that its $(2+\delta)$-moment is controlled by the kernel moment. Standard decoupling results \citep{delapena1995decoupling,delapena1999decoupling} then extend the bound to the original statistic.

\begin{lemma}(\citealp[][Equation~(3.3)]{gine2000exponential})
    \label{Lemma : moment bound U stat}
    Let $h_{i,j}:=h_{i,j}(X_i^{(1)},X_j^{(2)})$ be canonical kernels such that
    $\mE |h_{i,j}|^p<\infty$ for all $1\leq i,j\leq n$. Then, for any $p>2$, there exists a constant $C_p>0$ such that
    \begin{align*}
        \mE\Big| \sum_{i=1}^n \sum_{j=1}^n h_{i,j} \Big|^p
        \leq C_p \max \Biggl\{&
        p^p \biggl(\sum_{i=1}^n \sum_{j=1}^n \mE\, h_{i,j}^2\biggr)^{\!p/2}, p^{3p/2}\, \mE_1 \max_{1 \leq i \leq n} 
        \biggl(\sum_{j=1}^n \mE_2\, h_{i,j}^2\biggr)^{\!p/2}, \\
        & p^{3p/2}\, \mE_2 \max_{1 \leq j \leq n} \biggl(\sum_{i=1}^n \mE_1\, h_{i,j}^2\biggr)^{p/2},  p^{2p}\, \mE\max_{1 \leq i,j \leq n} |h_{i,j}|^p
        \Biggr\}.
    \end{align*}
    Here, $\mE_1$ and $\mE_2$ denote expectations with respect to $X^{(1)}$ and $X^{(2)}$, respectively.
\end{lemma}

\begin{lemma}[Moment bound for $T_{1,1}$ in \Cref{Proof of theorem: MMD Asymptotic Distribution}]
    \label{Lemma : moment bound T1}
    Let $\mathcal{W}$ be a measurable space, $P$ a probability measure on $\mathcal{W}$, and $H : \mathcal{W} \times \mathcal{W} \to \mathbb{R}$ a symmetric canonical kernel. 
    Let $\{W_i^{(1)}\}_{i=1}^B$ and $\{W_j^{(2)}\}_{j=1}^B$ be mutually independent i.i.d.\ samples from $P$. Define the decoupled statistic
    \begin{align*}
        T_{1,1}^{\mathrm{dec}}
        := \frac{1}{B}\sum_{i=1}^B \sum_{j=1}^B H(W_i^{(1)},W_j^{(2)})
    \end{align*}
    Then, for any $\delta > 0$, there exists a constant $C_\delta > 0$, independent of $B$, such that
    \begin{align*}
        \mE\bigl(|T_{1,1}^{\mathrm{dec}}|^{2+\delta}\bigr) \;\leq\; C_\delta \,\mE\bigl(|H(W_1,W_2)|^{2+\delta}\bigr).
    \end{align*}
    Moreover, by classical decoupling inequalities \citep{delapena1995decoupling,delapena1999decoupling}, the same bound holds for the block-level statistic $T_{1,1}$ up to a universal constant independent of $B$, where
    \begin{align*}
        T_{1,1} = \frac{1}{B}\sum_{1 \leq i \neq j \leq B} H(W_i,W_j).
    \end{align*}
\end{lemma}

\begin{proof}
    Fix $p = 2+\delta$ for some $\delta > 0$, and write $h_{ij} := H(W_i^{(1)}, W_j^{(2)})$. Applying \Cref{Lemma : moment bound U stat} together with $\max\{a_1,\ldots,a_m\} \leq a_1+\cdots+a_m$,
    \begin{align*}
        \mE\bigg|\sum_{i=1}^B \sum_{j=1}^B h_{ij}\bigg|^p
        \lesssim A_1 + A_2 + A_3 + A_4,
    \end{align*}
    where multiplicative constants independent of $B$ are absorbed into $\lesssim$, and
    \begin{align*}
        \begin{alignedat}{2}
            A_1 &= \Big\{\sum_{i=1}^B \sum_{j=1}^B \mE(h_{ij}^2)\Big\}^{p/2}, 
            &\qquad A_2 &= \mE_1\Bigg[ \max_{1\leq i\le B} 
            \Big\{\sum_{j=1}^B \mE_2(h_{ij}^2)\Big\}^{p/2}\Bigg], \\
            A_3 &= \mE_2\Bigg[ \max_{1\leq j\le B} 
            \Big\{\sum_{i=1}^B \mE_1(h_{ij}^2)\Big\}^{p/2}\Bigg], 
            &\qquad A_4 &= \mE\Big(\max_{1\leq i,j\le B} |h_{ij}|^p\Big).
        \end{alignedat}
    \end{align*}
    
    We bound each term in turn.
    
    \begin{enumerate}[label=(\alph*)]
        \item \textit{Term $A_1$.} By the i.i.d.\ structure, $\sum_{i=1}^B \sum_{j=1}^B \mE(h_{ij}^2) = B^2 \mE(h_{12}^2)$, and hence
        \begin{align*}
            A_1 \;\lesssim\; B^p \{\mE(h_{12}^2)\}^{p/2} \;\leq\; B^p \,\mE(|h_{12}|^p),
        \end{align*}
        where the last inequality uses Jensen's inequality.
        
        \vspace{0.3em}
        
        \item \textit{Terms $A_2$ and $A_3$.} Replacing the maximum by the sum,
        \begin{align*}
            \mE_1\bigg[\max_{1\le i\le B} 
            \Big\{\sum_{j=1}^B \mE_2 (h_{ij}^2)\Big\}^{p/2}\bigg]
            &\leq \sum_{i=1}^B \mE_1 \Big\{\sum_{j=1}^B \mE_2 (h_{ij}^2)\Big\}^{p/2}
            = B \cdot \mE_1\Big\{\sum_{j=1}^B \mE_2 (h_{1j}^2)\Big\}^{p/2},
        \end{align*}
        where the equality follows from {the i.i.d.\ structure}. Since $p>2$ and $x\mapsto x^{p/2}$ is convex, Jensen's inequality gives
        \begin{align*}
            \Big\{\sum_{j=1}^B \mE_2 (h_{1j}^2)\Big\}^{p/2}
            &\leq B^{p/2-1} \sum_{j=1}^B \{\mE_2(h_{1j}^2)\}^{p/2},
        \end{align*}
        and combining these with one further application of Jensen's inequality yields
        \begin{align*}
            A_2 \;\leq\; B^{1+p/2}\,\mE(|h_{12}|^p).
        \end{align*}
        The same bound holds for $A_3$ by interchanging the roles of $(W_i^{(1)})$ and $(W_j^{(2)})$.
        
        \vspace{0.3em}
        
        \item \textit{Term $A_4$.} Bounding the maximum by the sum,
        \begin{align*}
            A_4 \;\leq\; B^2 \,\mE(|h_{12}|^p).
        \end{align*}
    \end{enumerate}
    
    Combining these bounds and noting $B^{1+p/2} \leq B^p$ and $B^2 \leq B^p$ for $p \geq 2$,
    \begin{align*}
        \mE\bigg|\sum_{i=1}^B \sum_{j=1}^B h_{ij}\bigg|^p 
        \;\lesssim\; B^p \,\mE(|h_{12}|^p).
    \end{align*}
    Normalizing by $B^{-p}$ gives $\mE(|T_{1,1}^{\mathrm{dec}}|^p) \lesssim \mE(|h_{12}|^p)$, which establishes the first bound. The corresponding bound for the block-level statistic $T_{1,1}$ then follows from classical decoupling inequalities \citep{delapena1995decoupling,delapena1999decoupling}, up to a universal constant independent of $B$.
\end{proof}

The following lemma provides a concentration inequality for an infinite sum of chi-squared random variables.

\begin{lemma}(\citealp[][Lemma A.7]{Fan2026testofindependece})
    \label{lemma:concentration inequality:infinitysum_chi}
    Let $\{\alpha_n\}_{n \geq 1}$ be a sequence of real numbers satisfying $\alpha_1 \geq \alpha_2 \geq \ldots \geq 0$ and $\sum_{n=1}^{\infty} \alpha_n^2 = 1$, and let $\{\eta_n\}_{n \geq 1} \iid \chi_1^2$. Then, for any $\varpi > 0$,
    \begin{align*}
        \sup_{z \in \mathbb{R}} \mP\bigg\{z < \sum_{n=1}^{\infty} \alpha_n(\eta_n - 1) \leq z + \varpi\bigg\} \leq \sqrt{\bigg(\frac{4 \varpi}{\pi}\bigg)}.   
    \end{align*}
\end{lemma}
\mypara{Remark} \Cref{lemma:concentration inequality:infinitysum_chi} also holds for finite sums. For any finite sequence $\alpha_1 \ge \cdots \ge \alpha_L > 0$ with $\sum_{n=1}^L \alpha_n^2 = 1$, we can simply set $\alpha_n = 0$ for all $n > L$. This directly yields
\begin{align*}
    \sup_{z \in \mathbb{R}} \mP\bigg\{z < \sum_{n=1}^L \alpha_n(\eta_n - 1) \leq z + \varpi\bigg\} \leq \sqrt{\bigg(\frac{4 \varpi}{\pi}\bigg)}. 
\end{align*}

The following lemma adapts the non-asymptotic distributional approximation results of \citet{Fan2026testofindependece}, originally developed for independence testing with two random vectors, to the single-variate degenerate $U$-statistic setting used throughout our proofs.

\begin{lemma}[Degenerate $U$-statistic approximation]
    \label{lemma:degenerate_ustat_approx}
    Let $X_1,\ldots,X_n \iid P$ take values in a measurable space $\mathcal{X}$, and let $h : \mathcal{X} \times \mathcal{X} \to \mathbb{R}$ be a measurable symmetric kernel with $\mE\{h(X_1,X_2)^2\} \in (0, \infty)$, satisfying the degeneracy condition
    \begin{align*}
        \mE\{h(x, X_1)\} = 0 \qquad \text{for all } x\in\mathcal{X}.
    \end{align*}
    Define the degenerate $U$-statistic
    \begin{align*}
        U_n := \frac{1}{n(n-1)}\sum_{1\le i\ne j\le n} h(X_i,X_j),
    \end{align*}
    and let $Q : L^2(P) \to L^2(P)$ be the Hilbert--Schmidt integral operator
    \begin{align*}
        (Qg)(x) = \int h(x,y)\, g(y)\, dP(y),
    \end{align*}
    with eigenvalues $\{\lambda_{\ell \geq 1}\}$ ordered so that $|\lambda_1| \geq |\lambda_2| \geq \cdots$. Define
    \begin{align*}
        U_0 := \sum_{\ell=1}^\infty \lambda_\ell (V_\ell^2 - 1), 
        \qquad V_1, V_2, \ldots \iid N(0,1),
    \end{align*}
    and
    \begin{align*}
        \Delta_n := \sup_{t\in\mathbb{R}} \big| \mP\bigl(\sqrt{\{n(n-1)\}}\,U_n \le t\bigr) - \mP(U_0 \le t) \big|.
    \end{align*}
    Then {for every $0 < \delta \leq 1$, there exists a constant $C_\delta > 0$ depending only on $\delta$} such that
    \begin{align*}
        \Delta_n \leq C_\delta \Bigg(n^{-1/5} + \Bigg[
        \underbrace{\frac{\mE(|h(X_1,X_2)|^{2+\delta})}{n^{\delta/2}\, \{\mE(h(X_1,X_2)^2)\}^{(2+\delta)/2}}}_{=:\, \rho_\delta}
        \Bigg]^{1/(2\delta + 5)}
        \Bigg).
    \end{align*}
\end{lemma}

\noindent The argument parallels that of Theorem~3.3 in \citet{Fan2026testofindependece}, with modifications confined to the reduction from the two-sample independence-testing setting of \citet{Fan2026testofindependece} to the single-sample degenerate $U$-statistic considered here. We therefore omit the details.

\Cref{lemma:degenerate_ustat_approx} serves as a building block for the following non-asymptotic multiplier bootstrap approximation, which bounds the Kolmogorov distance between the conditional law of the bootstrap statistic and the non-Gaussian limit $U_0$.

\begin{lemma}[Multiplier bootstrap consistency] 
    \label{lemma:multiplier_bootstrap_consistency}
    
    Adopt the setting of \Cref{lemma:degenerate_ustat_approx}. Assume additionally that $h$ is positive semidefinite, and admits the spectral expansion
    \begin{align*}
        h(x,y) &= \sum_{\ell=1}^\infty \lambda_\ell \phi_\ell(x)\phi_\ell(y),
    \end{align*}
    with eigenvalues ordered as $\lambda_1 \ge \lambda_2 \ge \cdots \ge 0$. Here, the eigenfunctions $\{\phi_\ell\}_{\ell\ge1}$ are mutually uncorrelated with zero mean and unit variance, meaning $\mE\{\phi_\ell(X_1)\} = 0$ and $\mE\{\phi_\ell(X_1)\phi_m(X_1)\} = \mathds{1}(\ell=m)$ for all $\ell, m$. We also assume the diagonal identity
    \begin{align*}
        h(x,x) &= \sum_{\ell=1}^\infty \lambda_\ell \phi_\ell^2(x)
    \end{align*}
    holds almost surely.
    
    Let $\xi_1,\ldots,\xi_n \iid N(0,1)$ be independent of the data $\mathbb X_n = (X_1,\ldots,X_n)$. Define the original statistic $T_n$ and its multiplier bootstrap counterpart $T_n^\ast$ as
    \begin{align*}
        T_n := \frac1n \sum_{1\le i\ne j\le n} h(X_i,X_j), \qquad T_n^\ast := \frac1n \sum_{1\le i\ne j\le n} \xi_i\xi_j h(X_i,X_j).
    \end{align*}
    Let $U_0 := \sum_{\ell=1}^\infty \lambda_\ell(Z_\ell^2-1)$ with $Z_1,Z_2,\ldots \iid N(0,1)$. Define the relevant distance metrics as
    \begin{align*}
        \widetilde\Delta_n^\ast &:= \sup_{t\in\mathbb R} \big| \mP_\xi(T_n^\ast\le t\mid \mathbb X_n) - \mP(U_0\le t) \big|, \\
        \Delta_n &:= \sup_{t\in\mathbb R} \big| \mP(T_n\le t)-\mP(U_0\le t) \big|, \\
        \Delta_{n,\mathrm{boot}}^\ast &:= \sup_{t\in\mathbb R} \big| \mP_\xi(T_n^\ast\le t\mid\mathbb X_n) - \mP(T_n\le t) \big|.
    \end{align*}
    Then, for every $s>0$, there exists a universal constant $C>0$ such that
    \begin{align*}
        \mP\big(\widetilde\Delta_n^\ast\ge s\big) &\le \frac{C}{s} \bigg[ \frac{\mE\{h(X_1,X_1)^2\}}{n\,\mE\{h(X_1,X_2)^2\}} \bigg]^{1/5}, \\
        \mP\big(\Delta_{n,\mathrm{boot}}^\ast\ge s\big) &\le \frac1s \bigg\{ \Delta_n + C\bigg[ \frac{\mE\{h(X_1,X_1)^2\}}{n\,\mE\{h(X_1,X_2)^2\}} \bigg]^{1/5} \bigg\}.
    \end{align*}
    Consequently, if $\Delta_n\to0$ and $\mE\{h(X_1,X_1)^2\} / [n\,\mE\{h(X_1,X_2)^2\}]\to0$ as $n \to \infty$, then
    \begin{align*}
        \Delta_{n,\mathrm{boot}}^\ast &\to 0
    \end{align*}
    in probability.
\end{lemma}

\begin{proof}
    For $L\ge1$, define the truncated kernel
    \begin{align*}
        h_L(x,y) &:= \sum_{\ell=1}^L \lambda_\ell \phi_\ell(x)\phi_\ell(y),
    \end{align*}
    and define
    \begin{align*}
        T_{n,L}^\ast &:= \frac1n \sum_{1\le i\ne j\le n} \xi_i\xi_j h_L(X_i,X_j), \qquad U_{0,L} := \sum_{\ell=1}^L \lambda_\ell(Z_\ell^2-1).
    \end{align*}
    Let $\Phi_L(x) := \{\phi_1(x),\ldots,\phi_L(x)\}^\top$, $\Lambda_L := \mathrm{diag}(\lambda_1,\ldots,\lambda_L)$, and
    \begin{align*}
        W_L &:= \frac1{\sqrt n} \sum_{i=1}^n \xi_i\Phi_L(X_i).
    \end{align*}
    Conditional on $\mathbb X_n$,
    \begin{align*}
        W_L\mid\mathbb X_n &\sim N(0,\widehat\Sigma_L), \qquad \widehat\Sigma_L := \frac1n \sum_{i=1}^n \Phi_L(X_i)\Phi_L(X_i)^\top.
    \end{align*}
    Using the algebraic identity $\sum_{1\le i\ne j\le n}\xi_i\xi_j h_L(X_i,X_j) = \sum_{1\le i,j\le n}\xi_i\xi_j h_L(X_i,X_j) - \sum_{i=1}^{n} \xi_i^2 h_L(X_i,X_i)$, we can write
    \begin{align*}
        T_{n,L}^\ast &= \bar{Q}_{n,L}^\ast - \bar{D}_{n,L}^\ast,
    \end{align*}
    where
    \begin{align*}
        \bar{Q}_{n,L}^\ast &:= W_L^\top\Lambda_L W_L - \mathrm{tr}(\widehat\Sigma_L\Lambda_L), \qquad \bar{D}_{n,L}^\ast := \frac1n \sum_{i=1}^n (\xi_i^2-1)h_L(X_i,X_i).
    \end{align*}
    This centering uses the fact that $\mathrm{tr}(\widehat\Sigma_L\Lambda_L) = \frac1n \sum_{i=1}^n h_L(X_i,X_i)$.
    
    We first compare $\bar{Q}_{n,L}^\ast$ with $U_{0,L}$. Conditional on $\mathbb X_n$, the quadratic form $W_L^\top \Lambda_L W_L$ has the same distribution as $\sum_{j=1}^L \hat{\gamma}_j Z_j^2$, where $Z_j \iid N(0,1)$ and $\hat{\gamma}_1 \ge \cdots \ge \hat{\gamma}_L$ are the eigenvalues of the symmetric matrix $\sqrt{\Lambda_L} \widehat\Sigma_L \sqrt{\Lambda_L}$. Noting that $\mathrm{tr}(\widehat\Sigma_L \Lambda_L) = \sum_{j=1}^L \hat{\gamma}_j$, we have the distributional equivalence
    \begin{align*}
        \bar{Q}_{n,L}^\ast &\stackrel{d}{=} \sum_{j=1}^L \hat{\gamma}_j(Z_j^2 - 1).
    \end{align*}
    The limit statistic can be written as $U_{0,L} = \sum_{j=1}^L \lambda_j(Z_j^2 - 1)$, where $\lambda_1 \ge \cdots \ge \lambda_L$ are the eigenvalues of $\Lambda_L$. Coupling them on the same probability space using the same sequence $\{Z_j\}_{j=1}^L$, we define the error
    \begin{align*}
        A_{n,L} &:= \sum_{j=1}^L (\hat{\gamma}_j - \lambda_j)(Z_j^2 - 1), \qquad \text{so that} \quad \bar{Q}_{n,L}^\ast \stackrel{d}{=} U_{0,L} + A_{n,L}.
    \end{align*}
    This coupling is used only for the quadratic part $\bar{Q}_{n,L}^\ast$; no independence between $A_{n,L}$ and $\bar{D}_{n,L}^\ast$ is required.
    
    Let $\beta_L := \sum_{\ell=1}^L \lambda_\ell^2$, and define $\omega_L(r) := \sup_{t\in\mathbb R} \mP(t-r\le U_{0,L}\le t+r)$. Choose $L$ large enough such that $\beta_L > 0$. By the anti-concentration lemma for weighted sums of centered chi-square random variables (\Cref{lemma:concentration inequality:infinitysum_chi}),
    \begin{align*}
        \omega_L(r) &\lesssim \sqrt{\frac{r}{\beta_{L}^{1/2}}}.
    \end{align*}
    Therefore, for any $\varepsilon_1>0$,
    \begin{align*}
        \sup_{t\in\mathbb R} \Big| \mP_\xi(\bar{Q}_{n,L}^\ast\le t\mid\mathbb X_n) - \mP(U_{0,L}\le t) \Big| &\le \omega_L(\varepsilon_1) + \mP_Z(|A_{n,L}|>\varepsilon_1\mid\mathbb X_n).
    \end{align*}
    Conditional on $\mathbb X_n$, the variance of $A_{n,L}$ is $2\sum_{j=1}^L (\hat{\gamma}_j - \lambda_j)^2$. By the Hoffman--Wielandt theorem for symmetric matrices, the sum of squared eigenvalue differences is bounded by the squared Frobenius norm of the matrix difference. Thus,
    \begin{align*}
        \mV_Z(A_{n,L}\mid\mathbb X_n) &\le 2 \big\| \sqrt{\Lambda_L} \widehat\Sigma_L \sqrt{\Lambda_L} - \Lambda_L \big\|_F^2 \\
        &= 2 \mathrm{tr} \Big[ \big( \sqrt{\Lambda_L} (\widehat\Sigma_L - I_L) \sqrt{\Lambda_L} \big)^2 \Big].
    \end{align*}
    Using the cyclic property of the trace, this equals $2 \mathrm{tr}[((\widehat\Sigma_L - I_L)\Lambda_L)^2]$, which expands to
    \begin{align*}
        \mV_Z(A_{n,L}\mid\mathbb X_n) &\le 2 \sum_{\ell=1}^L \sum_{k=1}^L \lambda_\ell\lambda_k (\widehat\Sigma_{L,\ell k}-\delta_{\ell k})^2,
    \end{align*}
    where $\delta_{\ell k}$ is the Kronecker delta (which equals $1$ if $\ell = k$ and $0$ otherwise). Taking expectation over $\mathbb X_n$, and using the orthonormality of $\{\phi_\ell\}$ in $L^2(P)$,
    \begin{align*}
        \mE \big\{ \mV_Z(A_{n,L}\mid\mathbb X_n) \big\}
        &\le 2 \sum_{\ell=1}^L \sum_{k=1}^L \lambda_\ell\lambda_k \mE \big[ (\widehat\Sigma_{L,\ell k}-\delta_{\ell k})^2 \big]  \\
        &\le \frac{2}{n} \mE \bigg[ \sum_{\ell=1}^L \sum_{k=1}^L \lambda_\ell\lambda_k \phi_\ell^2(X_1)\phi_k^2(X_1) \bigg]  \\
        &= \frac{2}{n} \mE\{h_L(X_1,X_1)^2\}  \\
        &\le \frac{2}{n} \mE\{h(X_1,X_1)^2\}.
    \end{align*}
    The last inequality follows from $\lambda_\ell\ge0$ and the diagonal identity, which imply $0\le h_L(x,x)\le h(x,x)$.
    
    Next, since $\xi_i\sim N(0,1)$, $\mV_\xi(\xi_i^2-1)=2$. Therefore,
    \begin{align*}
        \mV_\xi(\bar{D}_{n,L}^\ast\mid\mathbb X_n) &= \frac2{n^2} \sum_{i=1}^n h_L(X_i,X_i)^2.
    \end{align*}
    Taking expectation,
    \begin{align*}
        \mE \big\{ \mV_\xi(\bar{D}_{n,L}^\ast\mid\mathbb X_n) \big\} &= \frac{2}{n} \mE\{h_L(X_1,X_1)^2\} \le \frac{2}{n} \mE\{h(X_1,X_1)^2\}.
    \end{align*}
    Now fix $\varepsilon_1,\varepsilon_2>0$. Since $T_{n,L}^\ast=\bar{Q}_{n,L}^\ast-\bar{D}_{n,L}^\ast$,
    \begin{align*}
        &\sup_{t\in\mathbb R} \big| \mP_\xi(T_{n,L}^\ast\le t\mid\mathbb X_n) - \mP(U_{0,L}\le t) \big|  \\
        &\qquad\le \omega_L(\varepsilon_1) + \omega_L(\varepsilon_2) + \mP_Z(|A_{n,L}|>\varepsilon_1\mid\mathbb X_n) + \mP_\xi(|\bar{D}_{n,L}^\ast|>\varepsilon_2\mid\mathbb X_n).
    \end{align*}
    By Chebyshev's inequality and the preceding variance bounds,
    \begin{align*}
        &\mE \bigg[\sup_{t\in\mathbb R} \big| \mP_\xi(T_{n,L}^\ast\le t\mid\mathbb X_n) - \mP(U_{0,L}\le t) \big| \bigg] \\
        &\qquad\lesssim \sqrt{ \frac{\varepsilon_1+\varepsilon_2}{\beta_{L}^{1/2}} } + \frac{\mE\{h(X_1,X_1)^2\}}{n\varepsilon_1^2} + \frac{\mE\{h(X_1,X_1)^2\}}{n\varepsilon_2^2}.
    \end{align*}
    It remains to pass from the truncated quantities to the infinite-dimensional ones. Let $\eta>0$. A standard smoothing argument gives
    \begin{align*}
        \widetilde\Delta_n^\ast
        &\le \sup_{t\in\mathbb R} \Big| \mP_\xi(T_{n,L}^\ast\le t\mid\mathbb X_n) - \mP(U_{0,L}\le t) \Big| \\
        &\quad + \mP_\xi(|T_n^\ast-T_{n,L}^\ast|>\eta\mid\mathbb X_n) + \mP(|U_0-U_{0,L}|>\eta) \\
        &\quad + \sup_{t\in\mathbb R}\mP(t-2\eta\le U_0\le t+2\eta) + \sup_{t\in\mathbb R}\mP(t-2\eta\le U_{0,L}\le t+2\eta).
    \end{align*}
    Since the spectral expansion holds in $L^2(P \times P)$, we obtain
    \begin{align*}
        \mE \bigg[\mE_\xi \Big[ (T_n^\ast-T_{n,L}^\ast)^2 \mid\mathbb X_n \Big]\bigg]
        &= \frac{2(n-1)}{n} \mE\{(h-h_L)(X_1,X_2)^2\} \\
        &= \frac{2(n-1)}{n} \sum_{\ell>L}\lambda_\ell^2 \to 0, \; \text{as }L \to \infty.
    \end{align*}
    and similarly,
    \begin{align*}
        \mE\{(U_0-U_{0,L})^2\} &= 2\sum_{\ell>L}\lambda_\ell^2 \to 0,\; \text{as }L \to \infty.
    \end{align*}
    Hence, by Markov's inequality, the two approximation probabilities vanish after taking expectation and then letting $L\to\infty$. Also, $\beta_L\to \sum_{\ell=1}^\infty \lambda_\ell^2 = \mE\{h(X_1,X_2)^2\}$, and the anti-concentration bounds for $U_{0,L}$ and $U_0$ imply that the two smoothing terms are bounded by a constant times
    \begin{align*}
        \sqrt{\frac{\eta}{\{\mE(h(X_1,X_2)^2)\}^{1/2}} }
    \end{align*}
    in the limit. Letting first $L\to\infty$ and then $\eta\downarrow0$, we get
    \begin{align*}
        \mE(\widetilde\Delta_n^\ast) &\lesssim \sqrt{\frac{\varepsilon_1+\varepsilon_2}{\{\mE(h(X_1,X_2)^2)\}^{1/2}}} + \frac{\mE\{h(X_1,X_1)^2\}}{n\varepsilon_1^2} + \frac{\mE\{h(X_1,X_1)^2\}}{n\varepsilon_2^2}.
    \end{align*}
    Set $\varepsilon_1=\varepsilon_2=\varepsilon$. Balancing the two terms gives
    \begin{align*}
        \varepsilon &= \bigg[ \frac{\mE\{h(X_1,X_1)^2\} [\,\mE\{h(X_1,X_2)^2\}]^{1/4}}{n} \bigg]^{2/5}.
    \end{align*}
    Substitution yields
    \begin{align*}
        \mE(\widetilde\Delta_n^\ast) &\lesssim \bigg[ \frac{\mE\{h(X_1,X_1)^2\}}{n\,\mE\{h(X_1,X_2)^2\}} \bigg]^{1/5}.
    \end{align*}
    Therefore, by Markov's inequality,
    \begin{align*}
        \mP\big(\widetilde\Delta_n^\ast\ge s\big) &\le \frac{C}{s} \bigg[ \frac{\mE\{h(X_1,X_1)^2\}}{n\,\mE\{h(X_1,X_2)^2\}} \bigg]^{1/5}.
    \end{align*}
    Finally, by the triangle inequality,
    \begin{align*}
        \Delta_{n,\mathrm{boot}}^\ast &\le \widetilde\Delta_n^\ast + \Delta_n.
    \end{align*}
    Taking expectations and applying Markov's inequality gives
    \begin{align*}
        \mP\big(\Delta_{n,\mathrm{boot}}^\ast\ge s\big) &\le \frac1s \bigg\{ \Delta_n + C\bigg[ \frac{\mE\{h(X_1,X_1)^2\}}{n\,\mE\{h(X_1,X_2)^2\}} \bigg]^{1/5} \bigg\}.
    \end{align*}
    This proves the desired result.
\end{proof}

The following lemma, adapted from \citet[Lemma~A.1]{neykov2021minimax} to accommodate a specified marginal distribution of $Z$, is the key technical ingredient in the proof of Theorem~\ref{Theorem: negative result}.

\begin{lemma}[Adaptation of {\citet[Lemma~A.1]{neykov2021minimax}} for Specified Marginals]\label{Lemma: hardness marginal fixed}
    Let $M \in (0, \infty]$ and let $(X, Y, Z)$ take values in $[-M, M]^{d_X} \times [-M, M]^{d_Y} \times \{1, 2\}$, with $\mP(Z = 1) = \lambda_n \in (0, 1)$ and $\mP(Z = 2) = 1 - \lambda_n$. Let $\{(X_i, Y_i, Z_i)\}_{i=1}^n \iid (X, Y, Z)$. Then, for any $\delta > 0$, there exists a constant $C > 0$ such that for any $\varepsilon > 0$ and any Borel set $D \subseteq ([-M,M]^{d_X+d_Y} \times \{1, 2\})^n \times [0, 1]$, one can construct i.i.d.\ random vectors $\{(\widetilde{X}_i, \widetilde{Y}_i, \widetilde{Z}_i)\}_{i=1}^n$ satisfying $\widetilde{Y}_i \independent \widetilde{Z}_i \mid \widetilde{X}_i$ and
    \begin{enumerate}[label=(\roman*)]
        \item $\displaystyle \mP\Big( \max_{i \in [n]} \|(\widetilde{X}_i, \widetilde{Y}_i) - (X_i, Y_i)\|_\infty < \varepsilon,\  \widetilde{Z}_i = Z_i \text{ for all } i \in [n] \Big) > 1 - \delta;$ 
        
        \item for any $U \sim \operatorname{Unif}[0,1]$ independent of $\{(\widetilde{X}_i, \widetilde{Y}_i, \widetilde{Z}_i)\}_{i=1}^n$,
        \begin{align*}
            \mP\Big(\big(\{(\widetilde{X}_i, \widetilde{Y}_i, \widetilde{Z}_i)\}_{i=1}^n, U\big) \in D\Big) \leq C \cdot \mu(D),
        \end{align*}
        where $\mu$ is the product of the counting measure on $\{1, 2\}^n$ and the Lebesgue measure on $\mathbb{R}^{(d_X + d_Y)n} \times [0, 1]$.
    \end{enumerate}
\end{lemma}

\begin{proof}
    The proof is a direct modification of Lemma~A.1 in
    \citet{neykov2021minimax}. We spell out the modification because the
    conditioning variable in the present problem is $X$, rather than $Z$.
    
    We first reduce to the bounded-density case, exactly as in Step I of
    Lemma~A.1 in \citet{neykov2021minimax}. When $M=\infty$, we truncate the
    distribution so that, with probability at least $1-\delta/2$, the
    truncated variables coincide with the original variables. Hence it
    suffices to consider the case $M<\infty$. As in Lemma~A.1, we may also
    replace the distribution by an arbitrarily close bounded-density
    approximation. In the present setting, to guarantee membership in
    $\mathcal P_M^{ac}$, we additionally mix each conditional distribution
    of $(X,Y)$ given $Z=z$ with a small uniform component on $[-M,M]^2$.
    This modification can be made with probability at most $\delta/2$ and
    ensures that, for each $z\in\{1,2\}$, the conditional density of
    $(X,Y)$ given $Z=z$ is bounded above and bounded away from zero on
    $[-M,M]^2$. By standard coupling arguments, these preliminary approximations introduce an error of at most $\delta$ to the final probability bounds, which can be safely absorbed or bounded.
    
    We now describe the construction in the one-dimensional case ($d_X=d_Y=1$). The
    higher-dimensional case follows by the same product-partition argument
    as in Remark~A.2 of \citet{neykov2021minimax}. Let
    $\{B_1,\ldots,B_m\}$ and $\{C_1,\ldots,C_m\}$ be equal-length partitions
    of $[-M,M]$, where $B_j$ partitions the support of $Y$ and $C_k$
    partitions the support of $X$. Further divide each $C_k$ into $m$
    subintervals
    \begin{align*}
        C_{1k},\ldots,C_{mk}.
    \end{align*}
    The important point is that the fine subinterval $C_{jk}$ is indexed by
    the $Y$-bin $j$, but not by the group label $Z$.
    
    Given a draw $(X,Y,Z)$ with $X\in C_k$, $Y\in B_j$, and $Z=i$, define
    \begin{align*}
        \widetilde Z:=Z,
    \end{align*}
    draw $\widetilde X$ uniformly on $C_{jk}$, and draw $\widetilde Y$
    uniformly on $B_j$. Applying this construction independently to each
    observation gives
    $\{(\widetilde X_i,\widetilde Y_i,\widetilde Z_i)\}_{i=1}^n$.
    
    Since $X$ and $\widetilde X$ both belong to $C_k$, and $Y$ and
    $\widetilde Y$ both belong to $B_j$, we have
    \begin{align*}
        \|(\widetilde X,\widetilde Y)-(X,Y)\|_\infty
        &\leq \frac{2M}{m}.
    \end{align*}
    Also $\widetilde Z=Z$ by construction. Hence, choosing $m$ sufficiently
    large so that $2M/m<\varepsilon$ yields property~\textup{(i)}, after
    combining with the preliminary truncation and bounded-density
    approximation.
    
    Next, we verify conditional independence. Fix $x\in C_{jk}$. Then the
    index $j$ is determined by $x$. Therefore, for any Borel set
    $A\subseteq[-M,M]$ and any $i\in\{1,2\}$,
    \begin{align*}
        \mP(\widetilde Y\in A\mid \widetilde X=x,\widetilde Z=i)
        &=
        \frac{|A\cap B_j|}{|B_j|}.
    \end{align*}
    The right-hand side does not depend on $i$. Therefore
    \begin{align*}
        \widetilde Y\independent \widetilde Z\mid \widetilde X.
    \end{align*}
    Thus the constructed law belongs to $\mathcal P_0$.
    
    We now check the mutual absolute continuity condition. Since the
    preliminary approximation gives a conditional density of $(X,Y)$ given
    $Z=i$ bounded away from zero, every cell $C_k\times B_j$ has positive
    probability under both labels $i=1,2$. Hence, for every $j,k$,
    \begin{align*}
        \mP(X\in C_k,Y\in B_j\mid Z=i)>0,
        \qquad i=1,2.
    \end{align*}
    Because the subinterval $C_{jk}$ depends only on the $Y$-bin $j$ and not
    on the group label $Z=i$, the conditional distributions
    $\mathcal L(\widetilde X\mid \widetilde Z=1)$ and
    $\mathcal L(\widetilde X\mid \widetilde Z=2)$ put positive density on
    exactly the same collection of intervals $\{C_{jk}:j,k\in[m]\}$.
    Consequently,
    \begin{align*}
        \mathcal L(\widetilde X\mid \widetilde Z=1)
        \ll
        \mathcal L(\widetilde X\mid \widetilde Z=2), \quad
        \mathcal L(\widetilde X\mid \widetilde Z=2)
        \ll
        \mathcal L(\widetilde X\mid \widetilde Z=1).
    \end{align*}
    Therefore the constructed law belongs to $\mathcal P_M^{ac}$.
    
    It remains to prove property~\textup{(ii)}. This follows from the same
    permutation-averaging and density-domination argument as in
    Lemma~A.1 of \citet{neykov2021minimax}. The only difference is that the
    present group variable $Z$ is binary with fixed marginal probabilities
    \begin{align*}
        \mP(Z=1) = \lambda_n, \quad
        \mP(Z=2) = 1-\lambda_n.
    \end{align*}
    This affects the cell-probability bound only through
    \begin{align*}
        \lambda_n+(1-\lambda_n) &= 1.
    \end{align*}
    Indeed, for every pair of cells $C_k$ and $B_j$,
    \begin{align*}
        \mP(X\in C_k,Y\in B_j)
        &=
        \sum_{z\in\{1,2\}}
        \mP(Z=z)
        \int_{C_k\times B_j}
        p_{X,Y\mid Z=z}(x,y)\,dx\,dy \\
        &\leq
        \lambda_n L |C_k||B_j|
        +
        (1-\lambda_n)L |C_k||B_j| \\
        &=
        L|C_k||B_j|.
    \end{align*}
    Thus the fixed marginal distribution of $Z$ does not change the
    domination argument. Applying Lemma~A.1 of
    \citet{neykov2021minimax} with the above relabeling of variables gives
    a constant $C<\infty$ such that, for every Borel set $D$,
    \begin{align*}
        \mP\left(
        \bigl(
        \{(\widetilde X_i,\widetilde Y_i,\widetilde Z_i)\}_{i=1}^n,
        U
        \bigr)\in D
        \right)
        &\leq
        C\mu(D),
    \end{align*}
    where $U$ is an auxiliary random variable and $\mu$ is the dominating measure as defined in Lemma~A.1 of \citet{neykov2021minimax}. This proves property~\textup{(ii)} and completes the proof.
\end{proof}

\section{Extended Literature and Technical Details}
\label{sec:extended_literature_technical}

\subsection{Comprehensive Review of Related Literature}
\label{Subsection : Comprehensive Review of Related Literature}
In the main text, we focused on recent developments directly addressing nonparametric conditional two-sample testing to maintain a concise narrative. In this subsection, we provide a broader overview of related statistical literature. Specifically, we discuss classical approaches targeting specific distributional aspects, enumerate various methodologies developed for conditional independence testing, and outline key advancements in direct density ratio estimation.

\mypara{Testing Conditional Moments and Goodness-of-Fit} While nonparametric comparisons of two conditional distributions have received limited attention until recently, related problems have been explored in the literature. These include testing for the equality of conditional moments \citep{hall1990bootstrap, kulasekera1995comparison,kulasekera1997smoothing,fan1998test,neumeyer2003nonparametric, PardoFernndez2015} and goodness-of-fit testing for pre-specified conditional distributions \citep{Andrews1997, zheng2000consistent, fan2006nonparametric}. 

\mypara{Conditional Independence Testing (CIT)} The problem of testing for conditional independence has been extensively studied, resulting in a variety of methods. Beyond the methods mentioned in the main text, recent improvements to the Generalized Covariance Measure \citep{shah2020hardness} include strategies such as weighting \citep{scheidegger2022weighted} and alternative projection techniques \citep{chakraborty2024doubly}. Other notable approaches include various kernel-based tests \citep{zhang2012kernel, doran2014permutation, pogodin2024practical}, binning-based tests \citep{kim2022local, neykov2024nearly}, regression-based tests \citep{dai2022significance}, and tests under the model-X framework \citep{candes2018panning, berrett2020conditional, liu2022fast, tansey2022holdout}.

\mypara{Density Ratio Estimation} To address the instability of separately estimating individual densities to form a ratio, \citet{tsuboi2009direct} developed methods that directly estimate the density ratio without involving density estimation. \citet{kanamori2010theoretical} provided a comparative analysis of such methods, while \citet{kanamori2009least} reformulated the problem as a least-squares optimization admitting a closed-form solution. Furthermore, \citet{liu2017trimmed} introduced trimmed estimation to improve robustness against extreme values. More recent developments in density ratio estimation include \citet{choi2021featurized, rhodes2020telescoping, choi2022density}.

\subsection{Importance-Weighted Test Statistics}
\label{sec:importance_weighted_statistics}

{The two examples below illustrate how the general principle of density ratio estimation in \main~yields basic conditional two-sample test statistics with Gaussian calibration.}

\begin{example}[Mean comparison] \normalfont
    Given a feature map $\psi : \mathcal{X} \times \mathcal{Y} \to \mathbb{R}$, consider the test statistic
    \begin{align*}
        \frac{1}{n_1} \sum_{i=1}^{n_1} \psi\bigl(V_i^{(1)}\bigr) - \frac{1}{n_2} \sum_{i=1}^{n_2} r_X\bigl(X_i^{(2)}\bigr)\,\psi\bigl(V_i^{(2)}\bigr).
    \end{align*}
    for the hypotheses in \main. The expectation of this statistic is equal to zero under the null hypothesis. Moreover, since the test statistic is simply a linear combination of independent random variables, it can be calibrated using the Gaussian approximation.
\end{example}

\begin{example}[Rank sum statistic] \normalfont \label{Example: Rank sum statistic}
    Instead of comparing the mean, one can compare the stochastic order of two distributions using ranks.
    Specifically, given a feature map $\psi : \mathcal{X} \times \mathcal{Y} \to \mathbb{R}$, a rank sum statistic based on the transformed samples can be computed as
    \begin{align*}
        \frac{1}{n_1 n_2} \sum_{i=1}^{n_1} \sum_{j=1}^{n_2} r_X\bigl(X_j^{(2)}\bigr)\,\mathds{1}\bigl\{\psi\bigl(V_j^{(2)}\bigr) < \psi\bigl(V_i^{(1)}\bigr)\bigr\}.
    \end{align*}
    Under the null hypothesis, and assuming no ties among transformed samples, one can verify that the expectation is equal to {1/2}. As in \cite{hu2024two}, the test statistic can be shown to be asymptotically Gaussian under suitable regularity conditions. Therefore, the critical value can be determined based on this Gaussian approximation. The power, however, depends on $\psi$. \citet{hu2024two} takes $\psi$ as an estimate of $f_{Y\sgiven X}^{(1)}(\cdot \given \cdot) /  f_{Y\sgiven X}^{(2)}(\cdot \given \cdot)$.
\end{example}

\subsection{Convergence Rates of Density Ratios}
\label{sec:convergence_rates}

Assumption~\ref{Assumption: classifier}(b) and Assumption~\ref{Assumption : Unified MMD}(b) impose convergence-rate conditions on the density ratio estimator. In terms of mean squared error (MSE), the former requires the conditional MSE to decay at rate $o_P(m^{-1})$, while the latter requires the unconditional MSE to decay at rate $o((n-m)^{-1/2})$. The latter is the milder of the two and is attainable under nonparametric smoothness alone: if $r_X$ belongs to a $\beta$-H\"older class on $\mathbb{R}^d$, the minimax optimal MSE rate is $O(n^{-2\beta/(2\beta + d)})$ \citep{nguyen2010estimating}, which satisfies $o(n^{-1/2})$ whenever $\beta > d/2$.

The former, matching the assumption in \citet{hu2024two}, requires the stricter near-parametric rate $o_P(m^{-1})$, which under sample splitting with $m/n \to 0$ is delivered by the following classes of estimators.

\begin{itemize}
    \item \textbf{Parametric models.} When {the density-ratio model is correctly specified} (e.g., an exponential-family form for $r_X$ {itself, not necessarily for the underlying densities}), least-squares estimators such as cLSIF and uLSIF achieve an MSE rate of $O(m^{-1})$ under standard regularity conditions \citep[Theorems~13.6 and~13.7]{Sugiyama_Suzuki_Kanamori_2012}.
    
    \item \textbf{Kernel methods.} When $r_X$ lies in a reproducing kernel Hilbert space (RKHS), kernel-based uLSIF attains an MSE rate of $m^{-1/(1+\gamma)}$, where $\gamma \in (0, 1)$ is {the bracketing-entropy exponent of the RKHS ball, controlled by the eigenvalue decay of the kernel} \citep[Theorem~14.16]{Sugiyama_Suzuki_Kanamori_2012}. As $\gamma \to 0${, which occurs for finite-dimensional RKHSs or sufficiently fast eigenvalue decay, the rate approaches the near-parametric $m^{-1}$}.
    
    \item \textbf{Deep learning.} When $r_X$ is $\beta$-H\"older on an approximately $d^\ast$-dimensional manifold embedded in $\mathbb{R}^d$, deep ReLU network estimators attain an MSE rate of $m^{-2\beta/(2\beta + d^\ast)} (\log m)^c$ \citep[Theorem~3.1]{zheng2022nonparametric}, {mitigating the curse of the ambient dimension $d$ by adapting to the intrinsic dimension $d^\ast$; this rate approaches $m^{-1}$ up to polylogarithmic factors when $\beta \gg d^\ast$, and draws on the manifold-adaptive approximation machinery of \citet{Jiao2023}}.
\end{itemize}

Assumption~\ref{Assumption: classifier}(b) is verified in the three structural settings listed above, whereas Assumption~\ref{Assumption : Unified MMD}(b) is verified under nonparametric smoothness of $r_X$ alone.

\section{Additional Numerical Experiments} \label{section: Additional Experiments}

This section complements the numerical study of \main. Appendix~\ref{Section: Experimental Details} gives the experimental details, and the remaining subsections report additional simulation results: the effect of density ratio estimation error on DRT methods (Appendix~\ref{Appendix : Density Ratio Estimation Error Analysis}), real data results for CIT and conditional kernel methods (Appendix~\ref{Appendix: CIT Real Data Results}), and two sensitivity analyses of Algorithm~\ref{Algorithm: Converting CIT into C2ST}, one assessing its effect on the CIT approach (Appendix~\ref{Appendix: With_Without_Algorithm_1}) and one examining the adjustment parameter $\varepsilon$ (Appendix~\ref{Appendix: Sensitivity Analysis by epsilon}).

\subsection{Experimental Details} 
\label{Section: Experimental Details}

{The implementation details of our numerical experiments cover density ratio estimation, the classifier-based test, the MMD-based tests, and the CIT approaches.}

\begin{algorithm}[t!]
    \caption{Cross-Fitted Multiplier Bootstrap Critical Value for Quadratic-Time MMD}
    \label{Algorithm: Cross-Fitted Multiplier Bootstrap MMD}
    \begin{algorithmic}[1]
        \Require Number of folds $K \geq 1$, cross-fitted test sets 
        $\{D_{\mathsf{test},j}\}_{j=1}^{K}$ with size $m$ per group, plug-in kernel contributions 
        $\{\widehat H^{(j)}\}_{j=1}^{K}$ from Section~\ref{Section: Kernel MMD}, number of bootstrap draws $R$, level $\alpha \in (0,1)$ \vskip .5em
        
        \For{$r=1,\dots,R$}
        \State Generate i.i.d. multipliers  $\xi_{i,j}^{(r)} \sim N(0,1)$, independent of the data.
        
        \For{$j=1,\dots,K$}
        \State Compute the fold-wise bootstrap statistic
        \begin{align*}
            \widehat{\mathrm{MMD}}^{2,(r)}_{1,\mathrm{Boot},j}
            =
            \frac{1}{\sqrt{m(m-1)}}
            \sum_{1 \leq i \neq i' \leq m}
            \xi_{i,j}^{(r)} \xi_{i',j}^{(r)}
            \widehat H_{ii'}^{(j)}.
        \end{align*}
        \EndFor
        
        \State Compute the cross-fitted bootstrap statistic
        \begin{align*}
            {}^{\dagger}\widehat{\mathrm{MMD}}^{2,(r)}_{1,\mathrm{Boot}}
            =
            \frac{1}{K}
            \sum_{j=1}^{K}
            \widehat{\mathrm{MMD}}^{2,(r)}_{1,\mathrm{Boot},j}.
        \end{align*}
        \EndFor \vskip .3em
        
        \State Form the collection 
        $\big\{{}^{\dagger}\widehat{\mathrm{MMD}}^{2,(r)}_{1,\mathrm{Boot}}\big\}_{r=1}^{R}$.
        
        \State Let
        ${}^{\dagger}\widehat{\mathrm{MMD}}^{2,\bullet 1}_{1,\mathrm{Boot}}
        \leq \cdots \leq
        {}^{\dagger}\widehat{\mathrm{MMD}}^{2,\bullet R}_{1,\mathrm{Boot}}$
        denote the ordered values.
        
        \State Define the empirical critical value
        \begin{align*}
            {}^{\dagger}\widehat q_{1-\alpha}^{R}
            =
            {}^{\dagger}\widehat{\mathrm{MMD}}^{2,\bullet \lceil R(1-\alpha)\rceil}_{1,\mathrm{Boot}}.
        \end{align*}
        
        \State \textbf{Output:} Critical value 
        ${}^{\dagger}\widehat q_{1-\alpha}^{R}$.
    \end{algorithmic}
\end{algorithm}

\mypara{Density Ratio Estimation} 
We estimate the density ratio $r_X(x) = f_{X}^{(1)}(x)/f_{X}^{(2)}(x)$ using the probabilistic classification-based approach of \citet[Section~3]{sugiyama2010density}. We focus on two classifiers: linear logistic regression (LLR) and kernel logistic regression (KLR).

Given samples $\{(X_i^{(1)},Y_i^{(1)})\}_{i=1}^{n_1} \iid P_{XY}^{(1)}$ and $\{(X_j^{(2)},Y_j^{(2)})\}_{j=1}^{n_2} \iid P_{XY}^{(2)}$, consider $\{(X_i, \ell_i)\}_{i=1}^{n}$, where $(X_1, \ldots, X_n) = (X_1^{(1)}, \ldots, X_{n_1}^{(1)}, X_1^{(2)}, \ldots, X_{n_2}^{(2)})$, $n = n_1 + n_2$, and $\ell_i = \mathds{1}(i \geq n_1 + 1)$. Denote $X_i = \big(X_i(1), \ldots, X_i(p)\big)^\top$.

\begin{itemize}
    \item For the LLR method, we use $\boldsymbol{\beta} = (\beta_0, \beta_1, \ldots, \beta_p)^\top$ and model the posterior probability as
    \begin{align*}
        \eta(X_i; \boldsymbol{\beta}) = \mP(\ell = 1 \given X_i) = \frac{1}{1 + \exp\bigl(-\beta_0 - \sum_{j=1}^{p} \beta_{j} X_i(j)\bigr)}.
    \end{align*}
    The estimated coefficients $\hat{\boldsymbol{\beta}}$ are obtained by minimizing the negative log-likelihood.
    
    \item For the KLR method \citep{zhu2005kernel}, {we use $\boldsymbol{\beta} = (\beta_0, \beta_1, \ldots, \beta_n)^\top$ and set $\eta(X_i; \boldsymbol{\beta}) = 1/\{1 + \exp(-\theta(X_i; \boldsymbol{\beta}))\}$, where $\theta(X_i; \boldsymbol{\beta}) = \beta_0 + \sum_{j=1}^{n} \beta_j k(X_i, X_j)$} and $k(x, y) = \exp(-\|x - y\|^2 / \sigma^2)$. The estimated coefficients $\hat{\boldsymbol{\beta}}$ are obtained by minimizing the penalized negative log-likelihood
    \begin{align*}
        -\sum_{i=1}^n \left\{\ell_i \theta(X_i; \boldsymbol{\beta}) - \log\!\bigl(1 + \exp(\theta(X_i; \boldsymbol{\beta}))\bigr)\right\} + \frac{\lambda}{2} \|\theta\|^2_{\mathcal{H}_k},
    \end{align*}
    where $\mathcal{H}_k$ is the RKHS generated by $k$ and $\lambda$ is a regularization parameter.
\end{itemize}

The density ratio estimate is then
\begin{align}
    \hat{r}_X(X_i) = \frac{n_2}{n_1} \cdot \frac{\eta(X_i; \hat{\boldsymbol{\beta}})}{1 - \eta(X_i; \hat{\boldsymbol{\beta}})}.
\end{align}

For the joint density ratio, we use $(X_i, Y_i)$ in place of $X_i$. We set $\sigma^2 = 200$ following \citet{hu2024two}, and fix $\lambda = 0.0005$ throughout our simulations.

\mypara{MMD-based Tests} {For the MMD-based tests in \main, we use a Gaussian kernel with the bandwidth fixed at $1$ across all main experiments. A comparison with the median heuristic and other bandwidth choices is provided in Section~\ref{sec:bandwidth_selection}. For the linear-time (MMD-$\ell$, $\gamma = 0$) and block-wise (MMD-$b$, $\gamma = 0.5$) variants, we use 2-fold cross-fitting ($K = 2$) with equal splitting ratio. For the quadratic-time variant (MMD-$q$, $\gamma = 1$), we use an 8:2 splitting ratio for density ratio estimation and test statistic computation, with calibration performed via the multiplier bootstrap described below.}

\mypara{Multiplier Bootstrap Algorithm}
For the quadratic-time variant $\gamma = 1$, we calibrate the test statistic via the multiplier bootstrap introduced in Section~\ref{Section: Kernel MMD}. Algorithm~\ref{Algorithm: Cross-Fitted Multiplier Bootstrap MMD} provides the pseudocode. The number of independent multiplier draws is $R = 299$.

\mypara{Classifier-based Test} {We implement the classifier-based test of Section~\ref{Section: Classifier-based Approach} with the weighted classifier $\widehat{h}$ from \eqref{Eq: classifier} trained on $D_{\mathsf{clf}}$. In our main experiments, $\widehat{h}$ is fit via \texttt{ranger} (Random Forests with $\texttt{num.trees} = 300$, $\texttt{max.depth} = 6$).} {On $D_{\mathsf{infer}}$ we take $m = \lfloor 0.2 \,\lvert D_{\mathsf{infer}}\rvert / 2 \rfloor$, yielding an $8{:}2$ split between $D_{\mathsf{ratio}}$ and $\tilde{D}_{\mathsf{eval}}$ per group. For the cross-validated variant $^\dagger$CLF defined in \eqref{Eq: cv_Acc statistic}, we use $K = 2$ folds of $D_{\mathsf{infer}}$ with the same $8{:}2$ within-fold ratio. The classifier $\widehat{h}$ is fit once on $D_{\mathsf{clf}}$ and held fixed across folds.}

\mypara{Randomized Conditional Independence Test} 
{The RCIT method of \citet{strobl2019approximate} is implemented with its default hyperparameters: the Lindsay--Pilla--Basak approximation for the null distribution, with $100$ random Fourier features for the conditioning set and $5$ for the non-conditioning sets.}

\mypara{Regression Methods for CIT}
{As stated in \main, we use Random Forest (via \texttt{ranger}) for both the generalized covariance measure \citep{shah2020hardness} and the projected covariance measure \citep{lundborg2024projected}, and XGBoost (via \texttt{xgboost}) for the WGSC procedure \citep{williamson2023general}. Each procedure internally invokes both a continuous and a binary regression; the hyperparameter settings are summarized in Table~\ref{tab : CIT regression method}.}

\begin{table}[t]
    \centering
    \caption{Regression methods and hyperparameter settings used in the CIT procedures. $d$ denotes the input dimension of the regression.}
    \resizebox{0.95\textwidth}{!}{%
        \begin{tabular}{cccc}
            \toprule
            \textbf{Regression Method} & $\mathsf{R}$\textbf{ Implementation} & \textbf{Tuning Parameters} & \textbf{Description} \\
            \midrule
            \multirow{2}{*}{Random Forest} & \multirow{2}{*}{\texttt{ranger}} & $\mathtt{num.trees} = 500$ & number of trees \\
            & & $\mathtt{mtry} = \lfloor \sqrt{d} \rfloor$ & number of candidate variables per split \\
            \midrule
            \multirow{3}{*}{XGBoost} & \multirow{3}{*}{\texttt{xgboost}} & $\mathtt{max\_depth} = 6$ & maximum tree depth \\
            & & $\eta = 0.3$ & learning rate \\
            & & $\mathtt{nrounds} = 100$ & number of boosting rounds \\
            \bottomrule
        \end{tabular}
    }
    \label{tab : CIT regression method}
\end{table}

\noindent The code for reproducing all of our simulation results (including those in \Cref{Appendix : Density Ratio Estimation Error Analysis,Appendix: CIT Real Data Results,Appendix: With_Without_Algorithm_1,Appendix: Sensitivity Analysis by epsilon}) and for more detailed settings is available on GitHub: \url{https://github.com/suman-cha/Cond2ST}.

\subsection{Impact of Density Ratio Estimation Errors on DRT Methods}
\label{Appendix : Density Ratio Estimation Error Analysis}
\begin{figure}[t]
    \centering
    \includegraphics[width=1.0\textwidth]{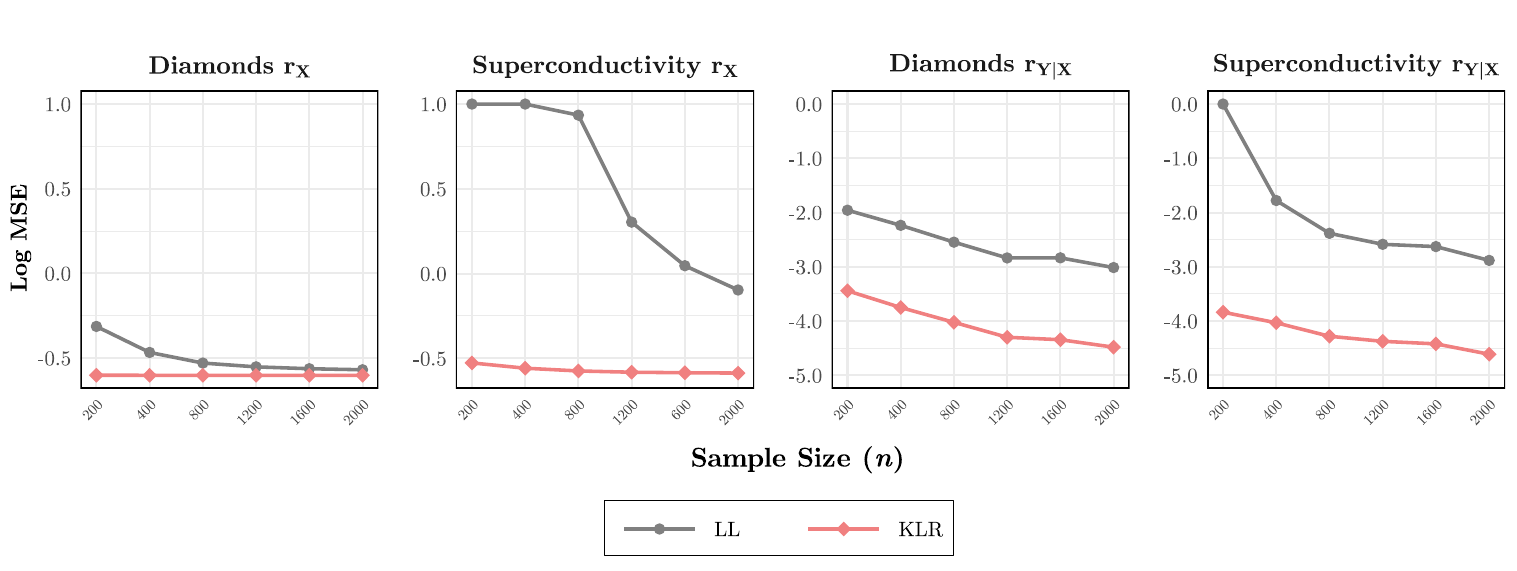}
    \caption{Log-scaled mean squared errors of the marginal density ratio $r_X(x)$ (\emph{first two panels}) and the conditional density ratio $r_{Y \given X}(y \given x)$ (\emph{last two panels}) for the LLR and KLR methods across sample sizes $n \in \{200, 400, 800, 1200, 1600, 2000\}$. Each point reports the median over $500$ simulations under the null hypothesis, for both the diamonds and superconductivity datasets.}
    \label{fig:density_ratio_mse}
\end{figure}
To complement the real-data analysis in \main, we examine how the accuracy of density ratio estimation affects the type~I error of DRT methods. Figure~\ref{fig:density_ratio_mse} reports the log-scaled mean squared error (MSE) of the marginal density ratio $r_X$ and the conditional density ratio $r_{Y \given X}$ for the LLR and KLR estimators, under $500$ null simulations per configuration of sample size, dataset, and estimation method.

On the low-dimensional diamonds dataset, both LLR and KLR attain low MSE for both density ratios, with the gap between the two methods narrowing as $n$ grows. This is consistent with the findings in \main, where LLR suffices to control the type~I error in low-dimensional settings. On the high-dimensional superconductivity dataset, the two methods diverge substantially: KLR maintains low and stable MSE across all sample sizes, whereas LLR incurs orders-of-magnitude larger MSE for both density ratios, particularly at small $n$. Although LLR improves with growing sample size, its estimation accuracy remains well below that of KLR and provides an empirical account of the poor type~I error control observed in \main. These results demonstrate that the validity of DRT methods depends empirically on the accuracy of the underlying density ratio estimator.

\subsection{Sensitivity to the Density-Ratio Split Proportion} 
\label{Appendix: Split Ratio Sensitivity}

\begin{table}[t!]
\centering
\caption{{Rejection rates of four DRT methods at three split proportions $(n-2m)/n \in \{0.2, 0.5, 0.8\}$ and four sample sizes $n \in \{200, 500, 1000, 2000\}$, across the three bounded synthetic scenarios. All entries are averages over $500$ repetitions at level $\alpha = 0.05$. The proportion $(n-2m)/n$ is the fraction of the sample allocated to density-ratio estimation.}}
\label{tab:split_ratio_bounded}
\footnotesize
\setlength{\tabcolsep}{4pt}
\begin{tabular}{llc cccc cccc}
    \toprule
    && & \multicolumn{4}{c}{Type~I error (Null)} & \multicolumn{4}{c}{Power (Alternative)} \\
    \cmidrule(lr){4-7} \cmidrule(lr){8-11}
    Scenario & Method & $\frac{(n-2m)}{n}$ & $n=200$ & $500$ & $1000$ & $2000$ & $200$ & $500$ & $1000$ & $2000$ \\
    \midrule
    \multirow{12}{*}{S1(B)} & \multirow{3}{*}{MMD-$\ell$}
    & 0.2 & 0.060 & 0.030 & 0.046 & 0.050 & 0.082 & 0.086 & 0.130 & 0.178 \\
    & & 0.5 & 0.050 & 0.040 & 0.032 & 0.044 & 0.060 & 0.076 & 0.094 & 0.148 \\
    & & 0.8 & 0.040 & 0.044 & 0.042 & 0.038 & 0.058 & 0.054 & 0.078 & 0.082 \\
    \cmidrule(l){2-11}
    & \multirow{3}{*}{MMD-$b$}
    & 0.2 & 0.292 & 0.178 & 0.150 & 0.094 & 0.382 & 0.530 & 0.790 & 0.976 \\
    & & 0.5 & 0.062 & 0.064 & 0.044 & 0.038 & 0.122 & 0.220 & 0.458 & 0.818 \\
    & & 0.8 & 0.054 & 0.030 & 0.024 & 0.044 & 0.072 & 0.108 & 0.146 & 0.344 \\
    \cmidrule(l){2-11}
    & \multirow{3}{*}{MMD-$q$}
    & 0.2 & 0.810 & 0.736 & 0.678 & 0.652 & 0.924 & 0.996 & 1.000 & 1.000 \\
    & & 0.5 & 0.238 & 0.200 & 0.174 & 0.142 & 0.558 & 0.906 & 0.994 & 1.000 \\
    & & 0.8 & 0.092 & 0.078 & 0.074 & 0.062 & 0.240 & 0.474 & 0.752 & 0.962 \\
    \cmidrule(l){2-11}
    & \multirow{3}{*}{CP}
    & 0.2 & 0.066 & 0.040 & 0.054 & 0.056 & 0.360 & 0.688 & 0.936 & 0.994 \\
    & & 0.5 & 0.060 & 0.046 & 0.060 & 0.062 & 0.562 & 0.936 & 0.986 & 0.996 \\
    & & 0.8 & 0.066 & 0.068 & 0.068 & 0.068 & 0.434 & 0.842 & 0.952 & 0.982 \\
    \midrule
    \multirow{12}{*}{S2(B)} & \multirow{3}{*}{MMD-$\ell$}
    & 0.2 & 0.050 & 0.038 & 0.046 & 0.042 & 0.138 & 0.272 & 0.400 & 0.660 \\
    & & 0.5 & 0.044 & 0.040 & 0.060 & 0.050 & 0.114 & 0.178 & 0.296 & 0.494 \\
    & & 0.8 & 0.062 & 0.044 & 0.048 & 0.034 & 0.096 & 0.140 & 0.174 & 0.306 \\
    \cmidrule(l){2-11}
    & \multirow{3}{*}{MMD-$b$}
    & 0.2 & 0.078 & 0.054 & 0.040 & 0.048 & 0.648 & 0.992 & 1.000 & 1.000 \\
    & & 0.5 & 0.026 & 0.054 & 0.048 & 0.034 & 0.338 & 0.826 & 1.000 & 1.000 \\
    & & 0.8 & 0.048 & 0.066 & 0.040 & 0.046 & 0.156 & 0.366 & 0.716 & 0.980 \\
    \cmidrule(l){2-11}
    & \multirow{3}{*}{MMD-$q$}
    & 0.2 & 0.476 & 0.270 & 0.236 & 0.162 & 1.000 & 1.000 & 1.000 & 1.000 \\
    & & 0.5 & 0.082 & 0.064 & 0.060 & 0.052 & 0.968 & 1.000 & 1.000 & 1.000 \\
    & & 0.8 & 0.054 & 0.046 & 0.044 & 0.060 & 0.496 & 0.956 & 1.000 & 1.000 \\
    \cmidrule(l){2-11}
    & \multirow{3}{*}{CP}
    & 0.2 & 0.034 & 0.046 & 0.046 & 0.044 & 0.056 & 0.038 & 0.036 & 0.070 \\
    & & 0.5 & 0.048 & 0.052 & 0.040 & 0.048 & 0.050 & 0.072 & 0.054 & 0.056 \\
    & & 0.8 & 0.056 & 0.050 & 0.040 & 0.048 & 0.076 & 0.066 & 0.034 & 0.058 \\
    \midrule
    \multirow{12}{*}{S3(B)} & \multirow{3}{*}{MMD-$\ell$}
    & 0.2 & 0.152 & 0.092 & 0.090 & 0.050 & 0.272 & 0.594 & 0.666 & 0.630 \\
    & & 0.5 & 0.074 & 0.056 & 0.044 & 0.062 & 0.360 & 0.586 & 0.664 & 0.614 \\
    & & 0.8 & 0.054 & 0.046 & 0.070 & 0.064 & 0.284 & 0.476 & 0.614 & 0.596 \\
    \cmidrule(l){2-11}
    & \multirow{3}{*}{MMD-$b$}
    & 0.2 & 0.524 & 0.364 & 0.274 & 0.164 & 0.712 & 0.806 & 0.866 & 0.888 \\
    & & 0.5 & 0.084 & 0.080 & 0.062 & 0.046 & 0.624 & 0.650 & 0.734 & 0.780 \\
    & & 0.8 & 0.046 & 0.048 & 0.044 & 0.040 & 0.464 & 0.602 & 0.664 & 0.650 \\
    \cmidrule(l){2-11}
    & \multirow{3}{*}{MMD-$q$}
    & 0.2 & 0.946 & 0.926 & 0.888 & 0.908 & 0.994 & 0.996 & 0.998 & 1.000 \\
    & & 0.5 & 0.416 & 0.350 & 0.276 & 0.250 & 0.808 & 0.898 & 0.954 & 0.970 \\
    & & 0.8 & 0.088 & 0.090 & 0.090 & 0.070 & 0.656 & 0.710 & 0.832 & 0.866 \\
    \cmidrule(l){2-11}
    & \multirow{3}{*}{CP}
    & 0.2 & 0.038 & 0.062 & 0.044 & 0.042 & 0.282 & 0.288 & 0.384 & 0.474 \\
    & & 0.5 & 0.040 & 0.066 & 0.054 & 0.050 & 0.292 & 0.374 & 0.496 & 0.640 \\
    & & 0.8 & 0.050 & 0.058 & 0.048 & 0.034 & 0.288 & 0.404 & 0.558 & 0.680 \\
    \bottomrule
\end{tabular}
\end{table}

The density-ratio-based tests of Section~\ref{Section: Approach via Density Ratio Estimation} partition the evaluation sample into a density-ratio estimation set of size $n-2m$ and a test-statistic set of size $2m$. The split proportion $(n-2m)/n$ controls a trade-off: allocating a larger share to density-ratio estimation reduces the plug-in error required by Assumption~\ref{Assumption : Unified MMD}(b), while allocating more to the test statistic increases its finite-sample signal at a given $\widehat{r}_X$. We examine this trade-off in the three bounded scenarios of Section~\ref{Section: Numerical Experiments}, where $r_X$ is bounded and the plug-in effect is isolated from density-ratio estimation error. For each scenario and each sample size $n \in \{200, 500, 1000, 2000\}$, we run four density-ratio tests (MMD-$\ell$, MMD-$b$, MMD-$q$, and CP) at three proportions $(n-2m)/n \in \{0.2, 0.5, 0.8\}$, where $(n-2m)/n$ is the fraction of the sample allocated to density-ratio estimation. The proportion $(n-2m)/n = 0.8$ is the $8{:}2$ split used in the main experiments. Rejection rates are averaged over $500$ repetitions at level $\alpha = 0.05$, and the complete results are reported in Table~\ref{tab:split_ratio_bounded}.

The sensitivity of the MMD statistics to $(n-2m)/n$ increases with the block parameter $\gamma$. For MMD-$q$ ($\gamma = 1$), the null rejection rate is severely inflated under the small allocation $(n-2m)/n = 0.2$ and settles near the nominal level at $(n-2m)/n = 0.8$. MMD-$b$ ($\gamma = 0.5$) shows the same pattern on a smaller scale, with a null rejection rate of $0.524$ in S3B at $n = 200$ under $(n-2m)/n = 0.2$ that settles to $0.046$ at $(n-2m)/n = 0.8$. MMD-$\ell$ ($\gamma = 0$) is the least sensitive, with null rejection rates near $\alpha$ at the larger allocations and a mild inflation only at $(n-2m)/n = 0.2$ under the smallest sample sizes. This gradient across $\gamma$ is consistent with Assumption~\ref{Assumption : Unified MMD}(b): larger $\gamma$ aggregates more kernel interactions and therefore requires tighter plug-in error, which is attained only at larger $(n-2m)/n$.

CP shows relatively insensitive result to $(n-2m)/n$. Its null rejection rate lies near nominal level across all scenarios, sample sizes, and proportions considered, and its power varies only marginally across proportions. CP is, however, substantially less powerful than MMD-$b$ or MMD-$q$ in S2B, where the alternative induces dispersion rather than a mean shift.

Under the alternative, the larger allocation $(n-2m)/n = 0.8$ reduces power relative to smaller allocations, most visibly for MMD-$b$ and MMD-$q$ at small sample sizes. The gap closes as $n$ grows and is below $0.05$ at $n=2{,}000$ for MMD-$q$ in S1B and S2B, consistent with the asymptotic negligibility of the plug-in error under Assumption~\ref{Assumption : Unified MMD}(b) when $(n-2m)/n$ is held above the level needed to control its rate. These observations support $(n-2m)/n = 0.8$ as the default for the MMD family in the main experiments: it controls size under all three scenarios while retaining power at moderate and large sample sizes. For CP, which does not rely on a plug-in variance, the choice of $(n-2m)/n$ does not affect size and the power differences are small.

\subsection{Bandwidth Selection for MMD-based Tests} \label{sec:bandwidth_selection}

\begin{figure}[t!]
    \centering
    \includegraphics[width=1\textwidth]{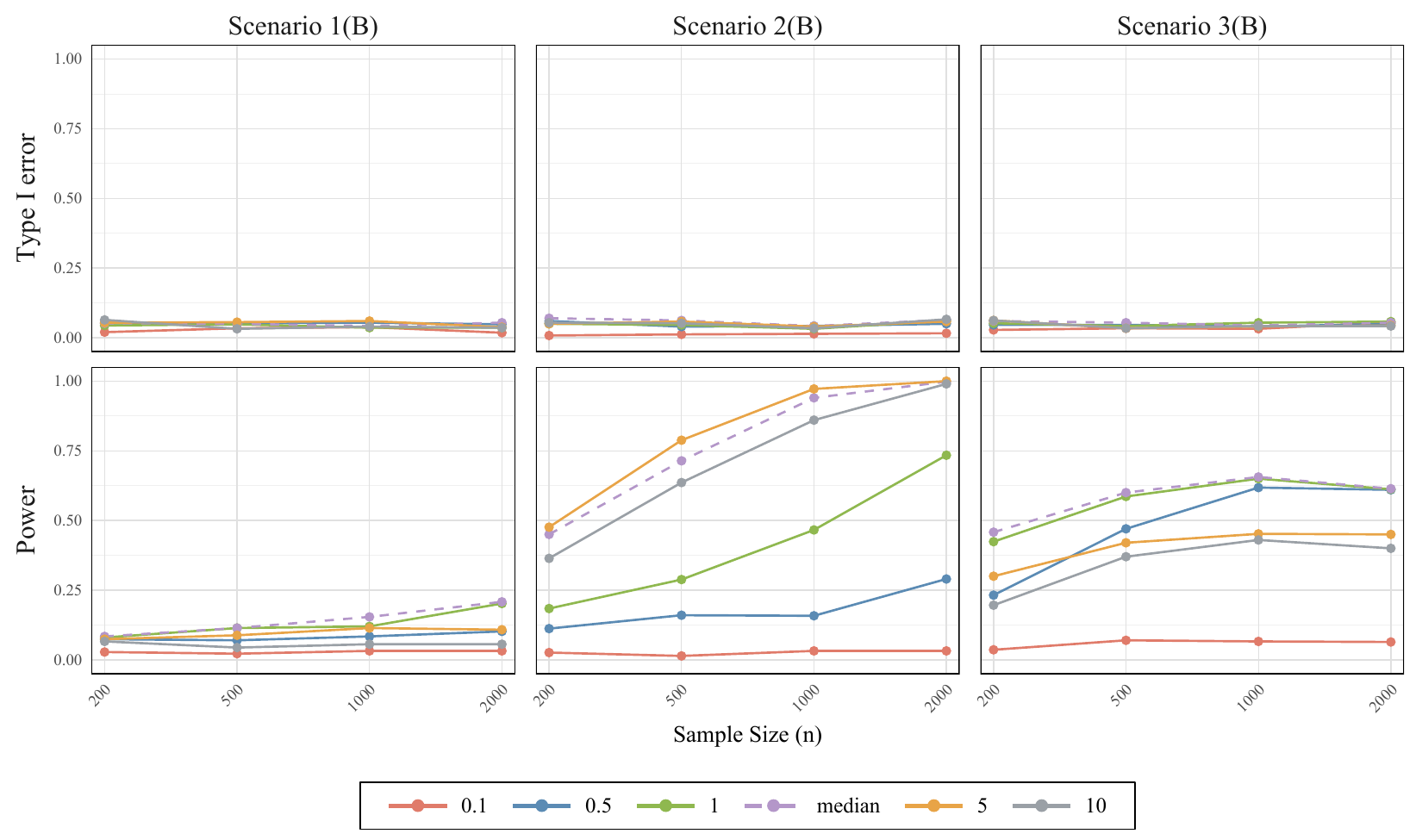}
    \captionsetup{skip=10pt}  
    \caption{{Rejection rates of ${}^{\dagger}\mathrm{MMD}\text{-}\ell$ in Scenarios~1(B), 2(B), and 3(B) under six bandwidths $h \in \{0.1, 0.5, 1, h_{\mathrm{med}}, 5, 10\}$, with the label ``median'' denoting $h_{\mathrm{med}}$. The top row reports type~I error rates under the null hypothesis and the bottom row reports power under the alternative hypothesis, each as a function of $n$. Results are averaged over $500$ repetitions with $\alpha = 0.05$.}}
    \label{fig:bandwidth_comparison}
\end{figure}

{For MMD-based tests we use a Gaussian kernel with a fixed bandwidth $h = 1$ throughout the main experiments. A fixed bandwidth keeps the kernel deterministic and independent of the data, which simplifies implementation and is aligned with the analyses in \citet{gretton2012kernel} and \citet{schrab2023mmd}. To assess the sensitivity of test performance to this choice, we repeat the ${}^{\dagger}\mathrm{MMD}\text{-}\ell$ experiments of Scenarios~1(B), 2(B), and 3(B) under six bandwidths, $h \in \{0.1, 0.5, 1, h_{\mathrm{med}}, 5, 10\}$, where $h_{\mathrm{med}}$ denotes the median heuristic. We report the bounded settings so that the effect of $h$ can be examined without confounding from density-ratio estimation error. Figure~\ref{fig:bandwidth_comparison} reports the type~I error rate and the power across $n \in \{200, 500, 1000, 2000\}$.}

The median heuristic attains power that is broadly competitive with, and in several configurations somewhat higher than, the best fixed bandwidth, whereas the two endpoints $h = 0.1$ and $h = 10$ lead to noticeably lower power. The type~I error rate remains close to the nominal level $\alpha = 0.05$ for all six bandwidths, so the variation reflects a power trade-off rather than a size concern. Throughout the main text, we nevertheless use $h = 1$ to keep the kernel a fixed user-specified function, which is the setting under which our theoretical analysis operates.
\subsection{Additional Simulations} 
\label{sec:additional_simulations}
\begin{figure}[t!]
    \centering
    \includegraphics[width=1\textwidth]{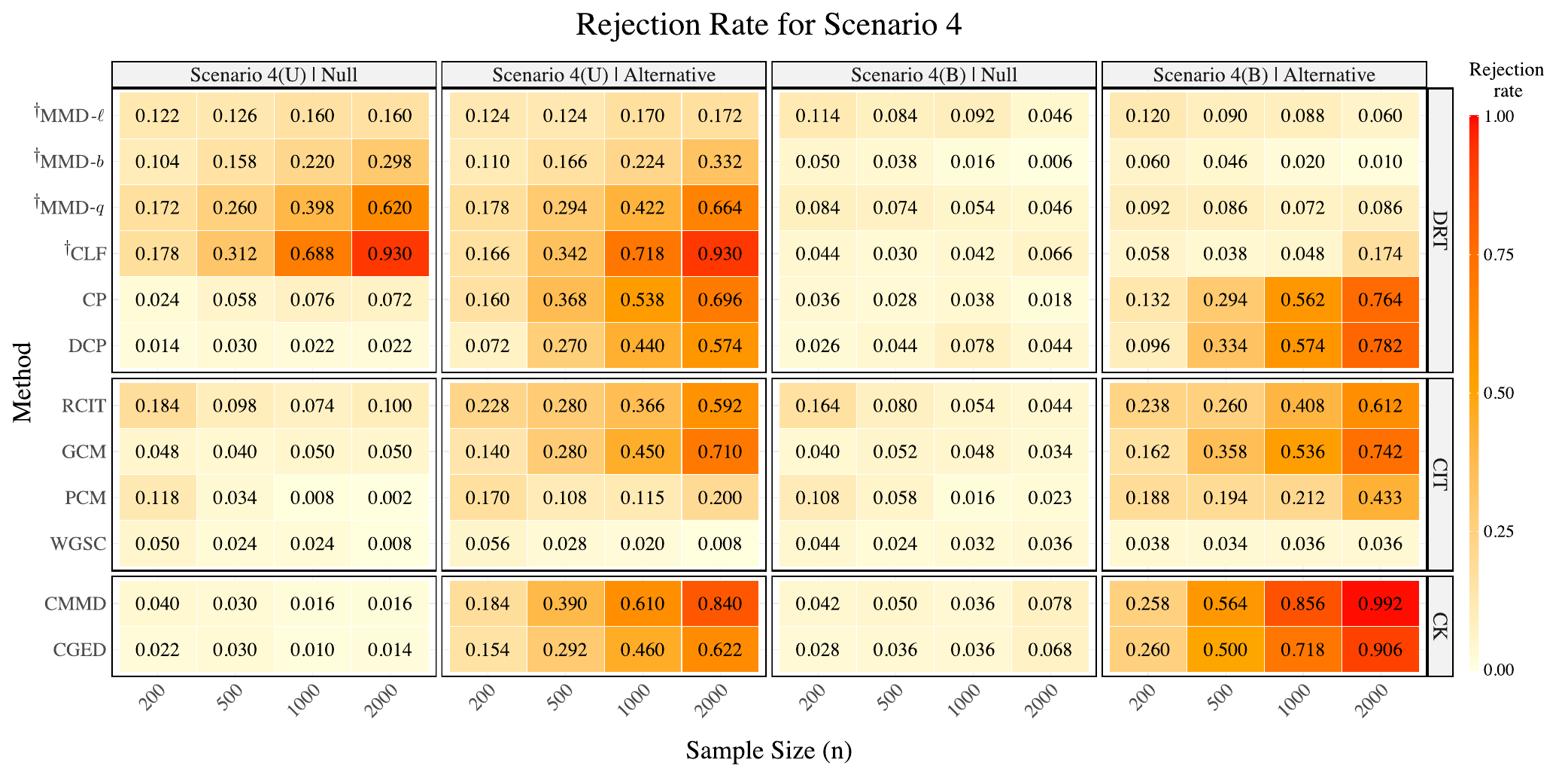}
    \captionsetup{skip=0pt}  
    \caption{{Rejection rates for Scenario~4 under null and alternative hypotheses, shown for both unbounded~(U) and bounded~(B) settings. Results are averaged over $500$ repetitions with $\alpha = 0.05$.}}
    \label{fig:Scenario 4}
\end{figure}

\begin{figure}[t!]
    \centering
    \includegraphics[width=1\textwidth]{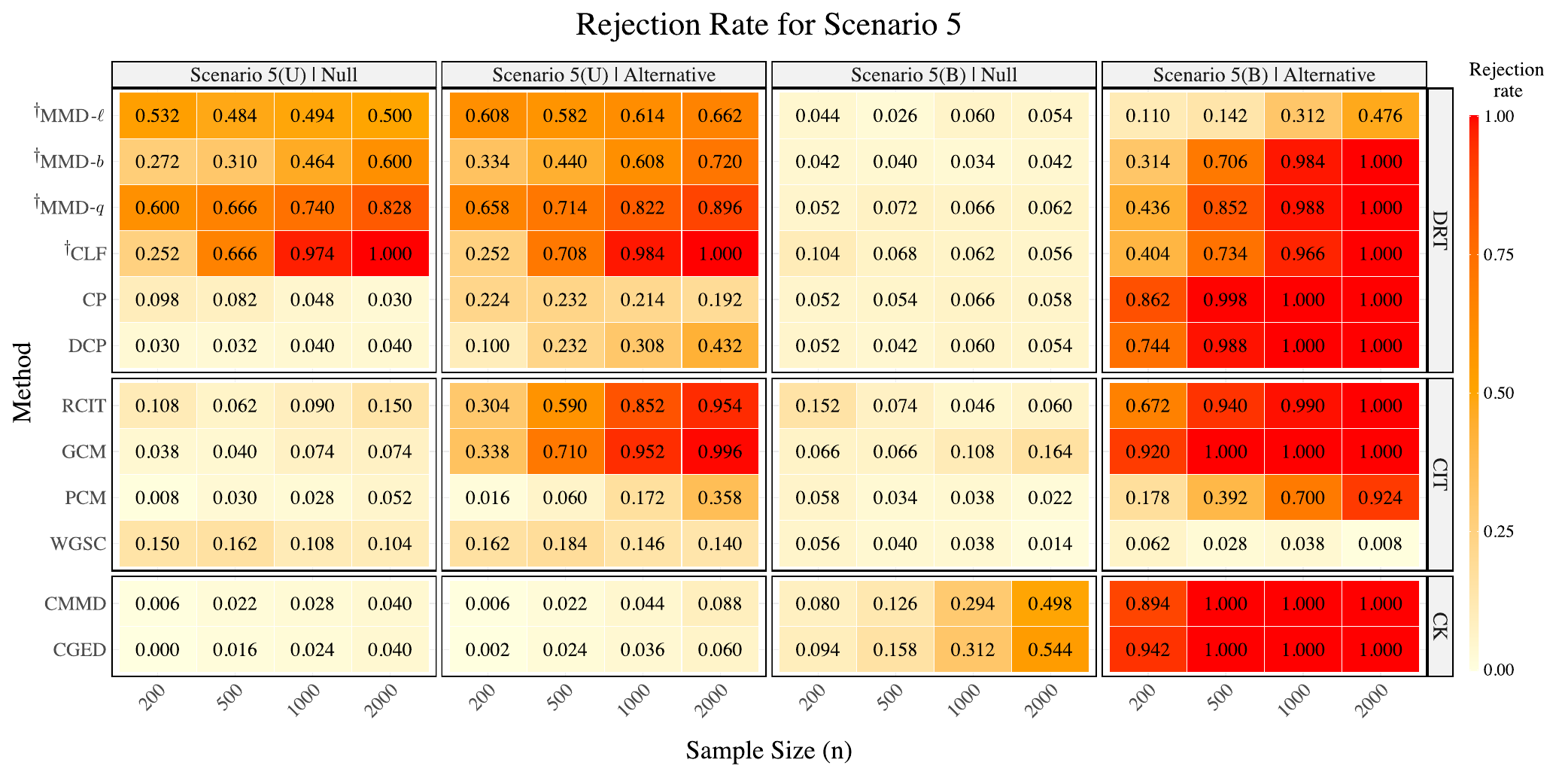}
    \captionsetup{skip=0pt}  
    \caption{Rejection rates for Scenario 5 under null and alternative hypotheses, shown for both unbounded (U) and bounded (B) settings. Results are averaged over 500 repetitions with significance level $\alpha = 0.05$.}
    \label{fig:Scenario 5}
\end{figure}

We complement the synthetic study in \main~with two further scenarios in which the covariate distributions depart from the Gaussian setting.

\mypara{Scenario 4: Linear Model with Heavy-Tailed Covariates} 
{This scenario examines performance under heavy-tailed covariate distributions. For each $j \in \{1, 2\}$, we generate $x^{(j)} = \mu^{(j)} + \tau^{(j)}$ with $\tau^{(j)} = (\tau_1^{(j)}, \ldots, \tau_p^{(j)})^\top$ and independent components $\tau_i^{(j)}$ following a $t$-distribution with $\nu_j$ degrees of freedom. We set $\nu_1 = 5$ and $\nu_2 = 4$ so that group~2 has heavier tails, which produces an unbounded marginal density ratio. The location vectors are $\mu^{(1)} = (1, 1, -1, -1, 0, \ldots, 0)^\top$ and $\mu^{(2)} = \mathbf{0}_p$. In the unbounded case~(4U), we use the full support of the $t$-distributions, whereas in the bounded case~(4B) we truncate the support of both distributions to $[-3, 3]^p$. The conditional distribution follows the specification of Scenario~\ref{subsec: Experimental Setup}: $y^{(j)} \given x^{(j)} = \delta^{(j)} + x^{(j)\top} \beta + \epsilon^{(j)}$, where $\beta = (1, -1, -1, 1, 0, \ldots, 0)^\top$ and $\epsilon^{(j)}$ follows a $t$-distribution with $2$ degrees of freedom. Under the null hypothesis we set $\delta^{(1)} = \delta^{(2)} = 0$, and under the alternative we set $\delta^{(1)} = 0$ and $\delta^{(2)} = 0.5$.}

\mypara{Scenario 5: Nonlinear Model with Beta-Distributed Covariates}
{This scenario examines performance when covariates follow group-specific product Beta distributions on $(0, 1)^p$. For group~1 the components of $x^{(1)}$ are independent $\mathrm{Beta}(0.5, 2)$, and for group~2 the components of $x^{(2)}$ are independent $\mathrm{Beta}(2, 2)$. In the unbounded case~(5U), the full support $(0, 1)^p$ is used, producing an unbounded marginal density ratio near the boundary. In the bounded case~(5B), we truncate the support to $[a_0, 1 - a_0]^p$ with $a_0 = 0.1$, producing a bounded density ratio. The conditional response is generated as
    \begin{align*}
        y^{(j)} \given x^{(j)} = \sin\bigl(2\pi x_1^{(j)}\bigr) + 0.5 \log\bigl(1 + 10\, x_2^{(j)}\bigr) + 0.3\, \bigl(x_3^{(j)} - 0.5\bigr)^2 + \epsilon^{(j)} + \delta^{(j)},
    \end{align*}
    where $\epsilon^{(j)} \sim N(0, 1)$, with $\delta^{(1)} = \delta^{(2)} = 0$ under the null and $\delta^{(1)} = 0$, $\delta^{(2)} = 0.5$ under the alternative.}

\subsection{Real Data Analysis for CIT and Conditional Kernel Methods}\label{Appendix: CIT Real Data Results}

\begin{figure}[t!]
    \centering
    \includegraphics[width=1\textwidth]{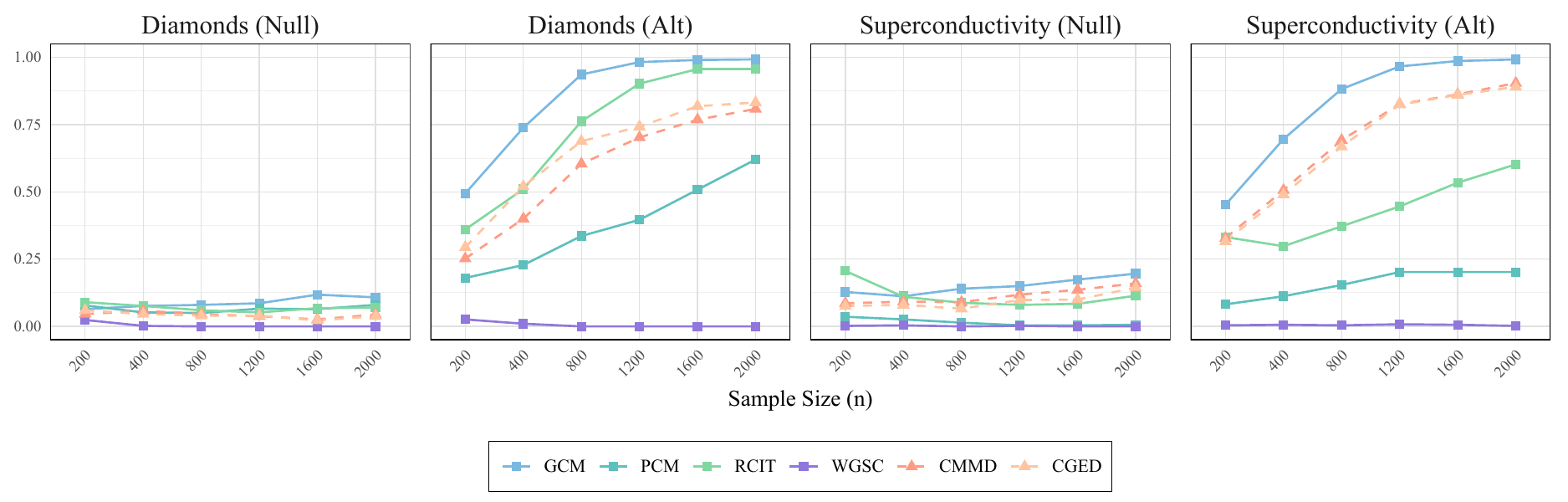}
    \caption{Rejection rates of CIT methods (GCM, PCM, RCIT, WGSC) and conditional kernel methods (CMMD, CGED) on the diamonds and superconductivity datasets under null and alternative hypotheses. Results are averaged over 500 repetitions with significance level $\alpha = 0.05$.}
    \label{fig:real CIT}
\end{figure}

We present the results for CIT and conditional kernel methods applied to the diamonds and superconductivity datasets, complementing the DRT analysis in \main. Figure \ref{fig:real CIT} reports the rejection rates of GCM, PCM, RCIT, WGSC, CMMD, and CGED under both the null and alternative hypotheses across the sample sizes specified in \Cref{subsec: Experimental Setup}.

For the low-dimensional diamonds dataset, all six methods control type~I error close to the nominal level $\alpha = 0.05$, and power increases with the sample size for every method except WGSC. Among the CIT methods, RCIT and GCM achieve the highest power; the conditional kernel methods CMMD and CGED achieve power comparable to RCIT and GCM at moderate to large sample sizes.

In the high-dimensional superconductivity dataset, the type~I error control is more variable. GCM exhibits null rejection rates that drift upward as the sample size grows, and RCIT shows inflation at small sample sizes, whereas PCM, WGSC, CMMD, and CGED stay near $\alpha$. Power patterns are broadly comparable to those in the diamonds dataset. Compared with the DRT methods reported in \main, the CIT and conditional kernel methods exhibit smaller size distortions in this high-dimensional regime, where the main source of DRT distortion is density-ratio estimation error.

\subsection{Sensitivity Analysis of CIT Methods to Algorithm~\ref{Algorithm: Converting CIT into C2ST}} 
\label{Appendix: With_Without_Algorithm_1}

\begin{table}[t!]
    \centering
    \caption{{Rejection rates of the CIT methods for Scenarios 1(U), 2(U), and 3(U) under the null and alternative hypotheses. Algorithm~\ref{Algorithm: Converting CIT into C2ST} is applied ($\checkmark$) or not ($\times$) as indicated in the ``Alg.~1'' column. The regression-based methods PCM, GCM, and WGSC are evaluated under Random Forest ($\texttt{rf}$) and XGBoost ($\texttt{xgb}$).}}
    \resizebox{\textwidth}{!}{%
        \begin{tabular}{ccccccccccc}
            \toprule
            \multirow{2}{*}{Scenario} & \multirow{2}{*}{Hypothesis} & \multirow{2}{*}{$n$} & \multirow{2}{*}{Alg.~1} & \multirow{2}{*}{RCIT} & \multicolumn{2}{c}{PCM} & \multicolumn{2}{c}{GCM} & \multicolumn{2}{c}{WGSC} \\
            \cmidrule(lr){6-7} \cmidrule(lr){8-9} \cmidrule(lr){10-11}
            & & & & & \texttt{rf} & \texttt{xgb} & \texttt{rf} & \texttt{xgb} & \texttt{rf} & \texttt{xgb} \\
            \midrule
            \multirow{8}{*}{S1U}
            & \multirow{4}{*}{Null}
            & \multirow{2}{*}{500}  & \checkmark & $0.056$ & $0.048$ & $0.056$ & $0.026$ & $0.074$ & $0.248$ & $0.050$ \\
            & &                        & $\times$   & $0.210$ & $0.040$ & $0.106$ & $0.034$ & $0.088$ & $0.070$ & $0.076$ \\
            & & \multirow{2}{*}{2000} & \checkmark & $0.048$ & $0.011$ & $0.088$ & $0.042$ & $0.074$ & $0.282$ & $0.016$ \\
            & &                        & $\times$   & $0.500$ & $0.007$ & $0.064$ & $0.052$ & $0.064$ & $0.038$ & $0.022$ \\
            \cmidrule{2-11}
            & \multirow{4}{*}{Alt}
            & \multirow{2}{*}{500}  & \checkmark & $0.228$ & $0.164$ & $0.102$ & $0.300$ & $0.240$ & $0.252$ & $0.048$ \\
            & &                        & $\times$   & $0.536$ & $0.204$ & $0.098$ & $0.448$ & $0.500$ & $0.086$ & $0.078$ \\
            & & \multirow{2}{*}{2000} & \checkmark & $0.610$ & $0.398$ & $0.198$ & $0.728$ & $0.728$ & $0.282$ & $0.024$ \\
            & &                        & $\times$   & $0.914$ & $0.478$ & $0.210$ & $0.832$ & $0.972$ & $0.086$ & $0.028$ \\
            \midrule
            \multirow{8}{*}{S2U}
            & \multirow{4}{*}{Null}
            & \multirow{2}{*}{500}  & \checkmark & $0.080$ & $0.060$ & $0.094$ & $0.048$ & $0.100$ & $0.270$ & $0.042$ \\
            & &                        & $\times$   & $0.080$ & $0.058$ & $0.074$ & $0.058$ & $0.098$ & $0.058$ & $0.056$ \\
            & & \multirow{2}{*}{2000} & \checkmark & $0.038$ & $0.009$ & $0.066$ & $0.042$ & $0.072$ & $0.272$ & $0.030$ \\
            & &                        & $\times$   & $0.054$ & $0.002$ & $0.066$ & $0.048$ & $0.078$ & $0.046$ & $0.020$ \\
            \cmidrule{2-11}
            & \multirow{4}{*}{Alt}
            & \multirow{2}{*}{500}  & \checkmark & $0.886$ & $0.992$ & $0.494$ & $0.062$ & $0.094$ & $0.320$ & $0.330$ \\
            & &                        & $\times$   & $0.868$ & $0.988$ & $0.566$ & $0.052$ & $0.072$ & $0.364$ & $0.494$ \\
            & & \multirow{2}{*}{2000} & \checkmark & $0.982$ & $1.000$ & $1.000$ & $0.042$ & $0.058$ & $0.486$ & $0.900$ \\
            & &                        & $\times$   & $0.980$ & $1.000$ & $1.000$ & $0.040$ & $0.062$ & $0.960$ & $0.994$ \\
            \midrule
            \multirow{8}{*}{S3U}
            & \multirow{4}{*}{Null}
            & \multirow{2}{*}{500}  & \checkmark & $0.104$ & $0.058$ & $0.096$ & $0.058$ & $0.086$ & $0.268$ & $0.038$ \\
            & &                        & $\times$   & $0.078$ & $0.052$ & $0.084$ & $0.040$ & $0.090$ & $0.056$ & $0.060$ \\
            & & \multirow{2}{*}{2000} & \checkmark & $0.070$ & $0.002$ & $0.064$ & $0.060$ & $0.058$ & $0.274$ & $0.026$ \\
            & &                        & $\times$   & $0.054$ & $0.007$ & $0.064$ & $0.048$ & $0.072$ & $0.044$ & $0.024$ \\
            \cmidrule{2-11}
            & \multirow{4}{*}{Alt}
            & \multirow{2}{*}{500}  & \checkmark & $0.726$ & $0.781$ & $0.744$ & $0.228$ & $0.260$ & $0.426$ & $0.708$ \\
            & &                        & $\times$   & $0.752$ & $0.776$ & $0.774$ & $0.234$ & $0.262$ & $0.742$ & $0.754$ \\
            & & \multirow{2}{*}{2000} & \checkmark & $0.778$ & $0.420$ & $0.802$ & $0.200$ & $0.218$ & $0.472$ & $0.790$ \\
            & &                        & $\times$   & $0.788$ & $0.450$ & $0.794$ & $0.212$ & $0.214$ & $0.816$ & $0.802$ \\
            \bottomrule
        \end{tabular}
    }
    \label{tab:scenario_u_performance_by_alg1}
\end{table}

{We examine how Algorithm~\ref{Algorithm: Converting CIT into C2ST} affects the finite-sample behavior of four CIT methods. In addition to RCIT, the regression-based procedures PCM, GCM, and WGSC are evaluated under Random Forest ($\texttt{rf}$) and XGBoost ($\texttt{xgb}$). Table~\ref{tab:scenario_u_performance_by_alg1} reports the rejection rates across the three scenarios, with each cell displayed as ``$\checkmark / \times$'' for the configurations with and without the algorithm.}

{The effect on RCIT is scenario-dependent. In Scenario~1(U), omitting the algorithm produces a type~I error that grows with $n$, reaching $0.500$ at $n=2{,}000$, whereas applying the algorithm yields rejection rates close to the nominal level $\alpha = 0.05$ once $n \geq 500$. In Scenarios~2(U) and 3(U), by contrast, the null rejection rates remain near $\alpha$ under both configurations, and the power profiles also differ only modestly.}

{For PCM and GCM, the null rejection rates are comparable with and without the algorithm in every scenario and sample size, so size control is preserved. Differences between the two configurations under the alternative shrink as $n$ grows and become negligible in Scenarios 2(U) and 3(U) at $n=2{,}000$, which is the empirical counterpart to the vanishing sample-loss rate of \main.} {WGSC deviates from this pattern. Under $\texttt{rf}$, its null rejection rate is inflated to roughly $0.25$--$0.28$ across all $n$ and all three scenarios whenever Algorithm~\ref{Algorithm: Converting CIT into C2ST} is applied, while it remains near $\alpha$ without the algorithm.}

\subsection{\texorpdfstring{Sensitivity Analysis of Algorithm~\ref{Algorithm: Converting CIT into C2ST} to $\varepsilon$}{Sensitivity Analysis of Algorithm: Converting CIT into C2ST to epsilon}}
\label{Appendix: Sensitivity Analysis by epsilon}

{The adjustment parameter $\varepsilon$ determines the size of the constructed dataset $\mathcal{D}_{\tilde{n}}$ through $\tilde{n} = \lfloor k^{\ast} n \rfloor$, with $k^{\ast}$ specified in Algorithm~\ref{Algorithm: Converting CIT into C2ST}. We assess how this choice affects finite-sample behavior using three candidates, $\varepsilon \in \bigl\{n^{-1},\; \{\log(n)\}^{-1},\; \{\log(n)\}^{-1/2}\bigr\}$, ordered from fast to slow decay. The analysis covers Scenarios 1(U)--3(U) under unbounded marginal density ratios, with $n \in \{200, 500, 1000\}$.}

{Table~\ref{tab:sensitivity_varepsilon} reports rejection rates under the null and alternative hypotheses. Across the three scenarios, both the null and alternative rejection rates vary only mildly with the choice of $\varepsilon$, with larger fluctuations confined to the smallest sample size $n=200$.}

\begin{table}[t!]
    \centering
    \caption{{Sensitivity of Algorithm~\ref{Algorithm: Converting CIT into C2ST} to $\varepsilon$ under Scenarios 1(U)--3(U). Rejection rates are based on $500$ repetitions with $\alpha = 0.05$.}}
    \vspace{-5pt}
    \small
    \resizebox{\textwidth}{!}{%
        \setlength{\tabcolsep}{6pt}
        \renewcommand{\arraystretch}{0.95}
        \begin{tabular}{cc c cccc c cccc}
            \toprule
            & & & \multicolumn{4}{c}{\textbf{Null}} & & \multicolumn{4}{c}{\textbf{Alternative}} \\
            \cmidrule(lr){4-7} \cmidrule(lr){9-12}
            Scenario & $n$ & $\varepsilon$ & RCIT & PCM & GCM & WGSC & & RCIT & PCM & GCM & WGSC \\
            \midrule
            \multirow{9}{*}{1(U)} 
            & \multirow{3}{*}{200} 
            & $n^{-1}$                & 0.164 & 0.064 & 0.022 & 0.072 & & 0.220 & 0.070 & 0.080 & 0.068 \\
            & & $\{\log(n)\}^{-1}$       & 0.166 & 0.064 & 0.036 & 0.076 & & 0.240 & 0.068 & 0.076 & 0.082 \\
            & & $\{\log(n)\}^{-1/2}$     & 0.168 & 0.072 & 0.018 & 0.062 & & 0.224 & 0.074 & 0.072 & 0.066 \\
            \cmidrule(lr){2-12}
            & \multirow{3}{*}{500} 
            & $n^{-1}$                & 0.090 & 0.038 & 0.026 & 0.052 & & 0.250 & 0.050 & 0.126 & 0.040 \\
            & & $\{\log(n)\}^{-1}$       & 0.056 & 0.066 & 0.024 & 0.050 & & 0.228 & 0.092 & 0.170 & 0.048 \\
            & & $\{\log(n)\}^{-1/2}$     & 0.080 & 0.036 & 0.036 & 0.044 & & 0.244 & 0.062 & 0.166 & 0.046 \\
            \cmidrule(lr){2-12}
            & \multirow{3}{*}{1000} 
            & $n^{-1}$                & 0.046 & 0.052 & 0.030 & 0.038 & & 0.376 & 0.078 & 0.258 & 0.048 \\
            & & $\{\log(n)\}^{-1}$       & 0.064 & 0.050 & 0.024 & 0.056 & & 0.394 & 0.098 & 0.262 & 0.054 \\
            & & $\{\log(n)\}^{-1/2}$     & 0.062 & 0.040 & 0.030 & 0.040 & & 0.394 & 0.078 & 0.288 & 0.038 \\
            \midrule
            \multirow{9}{*}{2(U)} 
            & \multirow{3}{*}{200} 
            & $n^{-1}$                & 0.168 & 0.070 & 0.030 & 0.064 & & 0.694 & 0.418 & 0.026 & 0.200 \\
            & & $\{\log(n)\}^{-1}$       & 0.162 & 0.074 & 0.044 & 0.060 & & 0.730 & 0.466 & 0.044 & 0.192 \\
            & & $\{\log(n)\}^{-1/2}$     & 0.164 & 0.068 & 0.028 & 0.064 & & 0.724 & 0.500 & 0.038 & 0.178 \\
            \cmidrule(lr){2-12}
            & \multirow{3}{*}{500} 
            & $n^{-1}$                & 0.080 & 0.054 & 0.020 & 0.046 & & 0.874 & 0.824 & 0.028 & 0.276 \\
            & & $\{\log(n)\}^{-1}$       & 0.074 & 0.060 & 0.046 & 0.042 & & 0.906 & 0.864 & 0.050 & 0.330 \\
            & & $\{\log(n)\}^{-1/2}$     & 0.078 & 0.040 & 0.024 & 0.038 & & 0.902 & 0.812 & 0.030 & 0.324 \\
            \cmidrule(lr){2-12}
            & \multirow{3}{*}{1000} 
            & $n^{-1}$                & 0.094 & 0.044 & 0.028 & 0.038 & & 0.958 & 0.968 & 0.022 & 0.576 \\
            & & $\{\log(n)\}^{-1}$       & 0.086 & 0.032 & 0.020 & 0.030 & & 0.968 & 0.990 & 0.026 & 0.604 \\
            & & $\{\log(n)\}^{-1/2}$     & 0.064 & 0.042 & 0.032 & 0.040 & & 0.964 & 0.980 & 0.022 & 0.626 \\
            \midrule
            \multirow{9}{*}{3(U)} 
            & \multirow{3}{*}{200} 
            & $n^{-1}$                & 0.160 & 0.070 & 0.026 & 0.070 & & 0.688 & 0.704 & 0.216 & 0.626 \\
            & & $\{\log(n)\}^{-1}$       & 0.166 & 0.062 & 0.028 & 0.064 & & 0.688 & 0.726 & 0.204 & 0.636 \\
            & & $\{\log(n)\}^{-1/2}$     & 0.112 & 0.060 & 0.026 & 0.076 & & 0.698 & 0.722 & 0.210 & 0.644 \\
            \cmidrule(lr){2-12}
            & \multirow{3}{*}{500} 
            & $n^{-1}$                & 0.074 & 0.038 & 0.006 & 0.048 & & 0.722 & 0.792 & 0.250 & 0.692 \\
            & & $\{\log(n)\}^{-1}$       & 0.076 & 0.064 & 0.030 & 0.038 & & 0.734 & 0.804 & 0.250 & 0.708 \\
            & & $\{\log(n)\}^{-1/2}$     & 0.084 & 0.054 & 0.030 & 0.038 & & 0.730 & 0.782 & 0.252 & 0.694 \\
            \cmidrule(lr){2-12}
            & \multirow{3}{*}{1000} 
            & $n^{-1}$                & 0.076 & 0.042 & 0.038 & 0.048 & & 0.766 & 0.822 & 0.216 & 0.738 \\
            & & $\{\log(n)\}^{-1}$       & 0.064 & 0.034 & 0.034 & 0.020 & & 0.774 & 0.826 & 0.214 & 0.762 \\
            & & $\{\log(n)\}^{-1/2}$     & 0.060 & 0.038 & 0.032 & 0.036 & & 0.768 & 0.822 & 0.208 & 0.760 \\
            \bottomrule
    \end{tabular}}
    \label{tab:sensitivity_varepsilon}
\end{table}

\subsection{A Refined Bound for $k^{\ast}$} 
\label{Appendix: refined k star}

The $k^{\ast}$ of Algorithm~\ref{Algorithm: Converting CIT into C2ST} is obtained by applying the multiplicative Chernoff bound (Lemma~\ref{Lemma: Concentration inequality}) to the overflow events $\{\tilde{n}_1 > n_1\}$ and $\{\tilde{n}_2 > n_2\}$, where $\tilde{n}_1 \sim \mathrm{Binomial}(\tilde{n}, n_1/n)$ and $\tilde{n} = \lfloor k n \rfloor$. Because the Chernoff bound is conservative, the resulting $k^{\ast}$ may be smaller than necessary, so the random subsample discards more observations than required. A tighter alternative is obtained by directly inverting the exact Binomial tail. Let
\begin{align*}
    q_j(k) \coloneqq \mP\bigl(\tilde{n}_j > n_j\bigr), \qquad \tilde{n}_1 \sim \mathrm{Binomial}(\lfloor k n \rfloor,\, n_1/n), \qquad \tilde{n}_2 = \lfloor k n \rfloor - \tilde{n}_1,
\end{align*}

and define

\begin{align*}
    k^{\ast}_{\mathrm{ref}} \coloneqq \max\Bigl\{\, k \in (0, 1) :\; \max_{j \in \{1, 2\}} q_j(k) \leq \varepsilon \,\Bigr\}.
\end{align*}

A coupling argument shows that each $q_j(k)$ is non-decreasing in $k$, so the defining set has the form $(0, k^{\ast}_{\mathrm{ref}}]$ and $k^{\ast}_{\mathrm{ref}}$ is computable by bisection over $(0, 1)$. Because the Chernoff inequality upper bounds $q_j(k)$, the constraint $\max_j q_j(k) \leq \varepsilon$ is weaker than the Chernoff-based constraint used in \main, so $k^{\ast}_{\mathrm{ref}} \geq k^{\ast}$ by construction. Thus $k^{\ast}_{\mathrm{ref}}$ retains at least as many observations while meeting the same $\varepsilon$-tail guarantee. Table~\ref{tab:kstar_comparison} compares the two values for balanced designs with $\varepsilon = 1/\log(n)$. At $n=2{,}000$, $k^{\ast}_{\mathrm{ref}}$ retains $\tilde{n} = 3{,}930$ observations out of $2n = 4{,}000$, against $3{,}785$ for the Chernoff-based $k^{\ast}$.

\begin{table}[t!]
    \centering
    \caption{{Comparison of the Chernoff-based $k^{\ast}$ and the refined $k^{\ast}_{\mathrm{ref}}$ for balanced designs with $n_1 = n_2 = n$ and $\varepsilon = 1/\log(n)$. The effective sample size is $\tilde{n} = \lfloor k \cdot 2n \rfloor$.}}
    \label{tab:kstar_comparison}
    \small
    \begin{tabular}{rr cccc}
        \toprule
        $n$ & $\varepsilon$ & $k^{\ast}$ & $k^{\ast}_{\mathrm{ref}}$ & $\tilde{n}$ (Chernoff) & $\tilde{n}$ (refined) \\
        \midrule
        200  & 0.189 & 0.854 & 0.958 & 341  & 383  \\
        500  & 0.161 & 0.901 & 0.970 & 900  & 970  \\
        1000 & 0.145 & 0.927 & 0.977 & 1853 & 1954 \\
        2000 & 0.132 & 0.946 & 0.983 & 3785 & 3930 \\
        \bottomrule
    \end{tabular}
\end{table}

For reproducibility, we provide an \texttt{R} implementation of $k^{\ast}_{\mathrm{ref}}$ below. The function inverts the Binomial tail probability by bisection over $(0, 1)$.

\vspace{0.5em}

\begin{center}
\begin{minipage}{0.72\textwidth}
\begin{lstlisting}[
language=R,
caption={\textsf{R} implementation for $k^{\ast}_{\mathrm{ref}}$ via exact Binomial tail computation.},
basicstyle=\ttfamily\scriptsize,
breaklines=true,
columns=fullflexible,
showstringspaces=false,
frame=single,
framesep=2pt,
aboveskip=0.3em,
belowskip=0.3em
]
compute_k_star_refined <- function(n1, n2, epsilon, tol = 1e-8) {
n <- n1 + n2
tail_prob <- function(k) {
tilde_n <- ceiling(k * n)
if (tilde_n < 2) return(0)
p1 <- 1 - pbinom(n1, size = tilde_n, prob = n1 / n)
p2 <- 1 - pbinom(n2, size = tilde_n, prob = n2 / n)
return(max(p1, p2))
}
k_low <- 0.01
k_high <- 1 - 1e-10
if (tail_prob(k_low) > epsilon) return(k_low)
while (k_high - k_low > tol) {
k_mid <- (k_low + k_high) / 2
if (tail_prob(k_mid) <= epsilon) k_low <- k_mid
else k_high <- k_mid
}
return(k_low)
}
\end{lstlisting}
\end{minipage}
\end{center}
        
\subsection{Power Comparison with the Oracle CIT} 
\label{Appendix: Oracle Comparison}

\begin{table}[t]
\centering
\caption{Rejection rates under Scenario~3(U) comparing the oracle CIT (O), applied to $n$ i.i.d.\ triples $(X_i, Y_i, Z_i)$, with the subsampled procedure (S) applied to two independent samples of fixed sizes $n_1 = n_2 = n/2$ via Algorithm~\ref{Algorithm: Converting CIT into C2ST} with $\varepsilon = 1/\log(n)$. All entries are averages over $500$ repetitions at $\alpha = 0.05$.}
\label{tab:oracle_comparison}
\footnotesize
\setlength{\tabcolsep}{5pt}
\begin{tabular}{l cc cc cc cc}
    \toprule
    & \multicolumn{2}{c}{$n = 200$} & \multicolumn{2}{c}{$n = 500$} & \multicolumn{2}{c}{$n = 1000$} & \multicolumn{2}{c}{$n=2{,}000$} \\
    \cmidrule(lr){2-3} \cmidrule(lr){4-5} \cmidrule(lr){6-7} \cmidrule(lr){8-9}
    Test & O & S & O & S & O & S & O & S \\
    \midrule
    GCM  & 0.087 & 0.071 & 0.141 & 0.117 & 0.219 & 0.212 & 0.297 & 0.285 \\
    PCM  & 0.409 & 0.367 & 0.555 & 0.513 & 0.673 & 0.649 & 0.677 & 0.677 \\
    RCIT & 0.500 & 0.460 & 0.792 & 0.722 & 0.840 & 0.894 & 0.958 & 0.938 \\
    WGSC & 0.267 & 0.232 & 0.490 & 0.434 & 0.615 & 0.584 & 0.665 & 0.653 \\
    \bottomrule
\end{tabular}
\end{table}

Algorithm~\ref{Algorithm: Converting CIT into C2ST} discards an $O\bigl(\sqrt{\{n \log(1/\varepsilon)\}}\bigr)$ fraction of the observations to restore the i.i.d.\ structure required by a conditional independence test, as established in \main. To quantify the resulting finite-sample power loss, we compare the subsampled procedure with an oracle CIT that operates directly on genuine i.i.d.\ triples. In the oracle setting, we generate $n$ i.i.d.\ copies of $(X_i, Y_i, Z_i)$ with $Z_i \sim \mathrm{Bernoulli}(n_1/n)$ and $(X_i, Y_i) \given Z_i \sim P_{XY}^{(Z_i)}$, and apply the CIT to all $n$ observations. In the subsampled setting, we apply Algorithm~\ref{Algorithm: Converting CIT into C2ST} to two independent samples of fixed sizes $n_1 = n_2 = n/2$ with $\varepsilon = 1/\log(n)$, matching the specification of Section~\ref{Section: Numerical Experiments}. The comparison is carried out for RCIT, GCM, PCM, and WGSC under Scenario~3(U) of Section~\ref{Section: Synthetic Data Examples}, where power grows with $n$ so that the gap is most visible. Rejection rates are averaged over $500$ repetitions at $\alpha = 0.05$ for $n \in \{200, 500, 1000, 2000\}$, and reported in Table~\ref{tab:oracle_comparison}.

The subsampled procedure tracks the oracle closely across the four tests and four sample sizes. The mean absolute difference $|$O $-$ S$|$ is $0.030$ and the maximum is $0.070$ (RCIT at $n = 500$). The only case in which the subsampled procedure exceeds the oracle is RCIT at $n = 1000$, with a deviation of $0.054$ that is comparable in magnitude to the positive gaps observed at adjacent sample sizes and does not indicate a systematic advantage.

The gap narrows as $n$ grows. At $n=2{,}000$ the two procedures differ by at most $0.020$ across the four tests, and PCM yields identical rejection rates under the two sampling schemes. This behavior is the empirical counterpart to the vanishing $n - \tilde{n}$ rate of \main. Algorithm~\ref{Algorithm: Converting CIT into C2ST} therefore preserves the power of the base CIT with a finite-sample gap that closes with $n$.

\subsection{A Practitioner's Guide} 
\label{Appendix: practitioners_guide}

\begin{figure}[t]
\centering
\includegraphics[width=\textwidth]{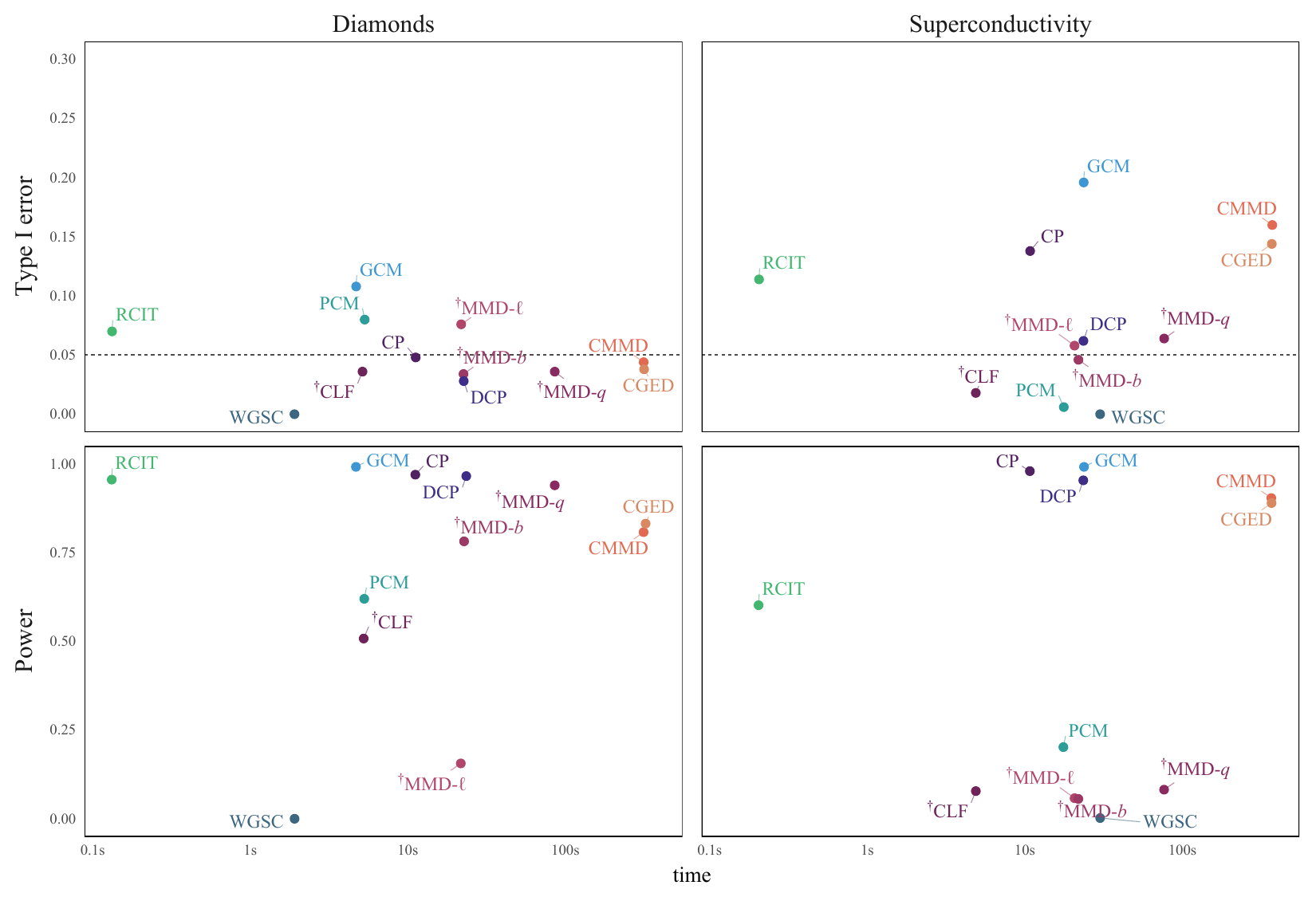}
\caption{Computation cost (seconds) versus performance at $n=2{,}000$ on the diamonds and superconductivity datasets, shown for the cross-fitted DRT variants ($^\dagger$MMD-$\ell$, $^\dagger$MMD-$b$, $^\dagger$MMD-$q$, $^\dagger$CLF), CP, DCP, the four CITs, and the conditional kernel methods (CMMD, CGED). The top row reports empirical type~I error under the null and the bottom row reports empirical power under the alternative. All entries are averaged over $500$ repetitions at $\alpha = 0.05$.}
\label{fig:computational-costs}
\end{figure}

{The performance of each testing method depends on the data structure and the choice of internal estimator, and our experiments confirm that no single method dominates uniformly across data-generating mechanisms or estimator choices. We therefore do not prescribe a method for each data type. Instead, this subsection reports the empirical trade-offs among computational cost, power, and type~I error control observed in our real-data experiments. Empirical result is summarized in Figure~\ref{fig:computational-costs} and Table~\ref{tab:practitioners_guide}, with all runtimes reported as end-to-end wall-clock times averaged over $500$ repetitions in \textsf{R} on a single thread.}

\mypara{Framework Selection}
The CIT-based and DRT-based frameworks rest on different identifiability conditions. The CIT-based framework, obtained by applying \Cref{Algorithm: Converting CIT into C2ST} to a conditional independence test, does not require boundedness of the marginal density ratio $r_X$ and is broadly robust across our synthetic scenarios under both the bounded and unbounded variants. The DRT-based framework relies on the importance-weighted statistics of \Cref{Section: Approach via Density Ratio Estimation}, which inherit the power properties of their unconditional counterparts when $r_X$ is bounded and accurately estimable. Its performance degrades when $r_X$ is unbounded or the estimator is misspecified, as in \Cref{Section: Synthetic Data Examples} and \Cref{Section: Real Data Analysis} of \main. The conditional kernel methods CMMD and CGED bypass both $r_X$ estimation and \Cref{Algorithm: Converting CIT into C2ST}. They attain competitive power in our experiments but incur computational costs at $n=2{,}000$ exceeding $300$ seconds on both real datasets, an order of magnitude beyond the slowest CIT variant in our experiments.

\mypara{Choices within the CIT Framework}
{Within the CIT family, our experiments reveal substantial heterogeneity across both methods and regression backbones. GCM combined with random forests attains the highest empirical power at every sample size on both real datasets at runtimes of $4.686$ and $23.760$ seconds for $n=2{,}000$, respectively. RCIT attains the lowest runtimes at $n = 2{,}000$, $0.132$ and $0.204$ seconds on the two datasets. The random-feature kernel approximation underlying RCIT also induces mild type~I error inflation at $n = 200$ in our experiments. PCM shares the regression-based construction of GCM but attains markedly lower empirical power at comparable runtime, illustrating that the within-framework choice of test matters even when the regression backbone is held fixed. WGSC paired with gradient boosting controls type~I error tightly but yields near-zero empirical power across both real datasets, suggesting that its weighting may be poorly suited to the alternatives we consider. The choice of regression backbone influences both power and runtime in the synthetic experiments of \Cref{Section: Synthetic Data Examples}, where linear models are fast but underpower nonlinear scenarios while flexible estimators recover power at moderate computational cost.}

\begin{table*}[t!]
\centering
\small
\caption{Average computation time in seconds with standard deviation across $500$ replications, on the diamonds and superconductivity datasets.}
\label{tab:practitioners_guide}
\resizebox{\textwidth}{!}{%
\begin{tabular}{@{}llrrrrrr@{}}
\toprule
& & \multicolumn{3}{c}{Diamonds ($d=6$)} & \multicolumn{3}{c}{Superconductivity ($d=81$)} \\
\cmidrule(lr){3-5} \cmidrule(lr){6-8}
Method & Estimator & $n=200$ & $n=800$ & $n=2{,}000$ & $n=200$ & $n=800$ & $n=2{,}000$ \\
\midrule
\multirow{2}{*}{${}^\dagger$MMD-$\ell$} & LLR & $0.008 \pm 0.002$ & $0.011 \pm 0.004$ & $0.017 \pm 0.004$ & $0.044 \pm 0.006$ & $0.066 \pm 0.005$ & $0.129 \pm 0.007$ \\
                                    & KLR & $0.041 \pm 0.008$ & $0.804 \pm 0.100$ & $11.200 \pm 1.240$ & $0.044 \pm 0.005$ & $0.810 \pm 0.056$ & $10.900 \pm 0.370$ \\
\cmidrule(l){1-8}
\multirow{2}{*}{${}^\dagger$MMD-$b$}    & LLR & $0.008 \pm 0.004$ & $0.013 \pm 0.003$ & $0.023 \pm 0.005$ & $0.046 \pm 0.003$ & $0.088 \pm 0.006$ & $0.183 \pm 0.014$ \\
                                    & KLR & $0.093 \pm 0.013$ & $3.243 \pm 0.360$ & $43.560 \pm 5.090$ & $0.081 \pm 0.011$ & $2.769 \pm 0.150$ & $39.700 \pm 0.670$ \\
\cmidrule(l){1-8}
\multirow{2}{*}{${}^\dagger$MMD-$q$}    & LLR & $0.010 \pm 0.003$ & $0.042 \pm 0.004$ & $0.207 \pm 0.020$ & $0.050 \pm 0.004$ & $0.142 \pm 0.007$ & $0.576 \pm 0.027$ \\
                                    & KLR & $0.097 \pm 0.021$ & $3.298 \pm 0.360$ & $43.780 \pm 5.230$ & $0.082 \pm 0.012$ & $2.841 \pm 0.150$ & $40.140 \pm 0.950$ \\
\cmidrule(l){1-8}
\multirow{2}{*}{${}^\dagger$CLF}        & LLR & $0.048 \pm 0.010$ & $0.096 \pm 0.020$ & $0.174 \pm 0.031$ & $0.125 \pm 0.012$ & $0.222 \pm 0.042$ & $0.551 \pm 0.043$ \\
                                    & KLR & $0.087 \pm 0.013$ & $0.674 \pm 0.053$ & $7.709 \pm 0.730$ & $0.115 \pm 0.014$ & $0.755 \pm 0.048$ & $7.369 \pm 0.240$ \\
\midrule
RCIT  &              & $0.024 \pm 0.003$ & $0.062 \pm 0.009$ & $0.132 \pm 0.014$ & $0.042 \pm 0.007$ & $0.106 \pm 0.017$ & $0.204 \pm 0.022$ \\
GCM   & \texttt{rf}  & $0.314 \pm 0.047$ & $1.518 \pm 0.200$ & $4.686 \pm 0.460$ & $0.828 \pm 0.110$ & $5.971 \pm 0.730$ & $23.760 \pm 2.270$ \\
PCM   & \texttt{rf}  & $0.415 \pm 0.064$ & $1.732 \pm 0.220$ & $5.293 \pm 0.530$ & $0.761 \pm 0.100$ & $4.853 \pm 0.600$ & $17.560 \pm 1.670$ \\
WGSC  & \texttt{xgb} & $0.940 \pm 0.120$ & $1.424 \pm 0.180$ & $1.912 \pm 0.180$ & $13.740 \pm 1.810$ & $23.710 \pm 2.950$ & $30.070 \pm 2.960$ \\
\midrule
CMMD  &              & $2.821 \pm 0.070$ & $45.030 \pm 0.360$ & $313.700 \pm 3.170$ & $3.725 \pm 0.130$ & $57.910 \pm 1.220$ & $366.700 \pm 12.400$ \\
CGED  &              & $2.829 \pm 0.064$ & $45.350 \pm 0.390$ & $323.000 \pm 4.670$ & $3.652 \pm 0.180$ & $58.550 \pm 1.190$ & $368.000 \pm 14.000$ \\
\bottomrule
\end{tabular}%
}
\end{table*}

\mypara{Choices within the DRT Framework}
{Within the DRT family, both the test statistic and the density ratio estimator interact with the alternative and the dimension of $X$ in our experiments. The classifier-based test ${}^{\dagger}\widehat{\mathrm{Acc}}$ of \Cref{Section: Classifier-based Approach} is targeted at alternatives that shift the Bayes decision boundary, since its asymptotic power scales with the total variation distance between the two conditional distributions. Within the MMD family in our synthetic experiments, ${}^\dagger$MMD-$q$ attains the highest empirical power, ${}^\dagger$MMD-$\ell$ admits a balanced split with lower computational cost at the expense of moderate power, and ${}^\dagger$MMD-$b$ interpolates between the two. The density ratio estimator further interacts with dimension. In the low-dimensional diamonds setting, linear logistic regression (LLR) is both fast and sufficient for type~I error control, with ${}^\dagger$MMD-$\ell$ (LLR) requiring only $0.017$ seconds at $n=2{,}000$. In the high-dimensional superconductivity setting, however, LLR alone leaves type~I error inflated, and a richer estimator such as kernel logistic regression (KLR) is needed to restore size control. The cost of KLR grows rapidly with $n$. On superconductivity at $n=2{,}000$, ${}^\dagger$MMD-$\ell$ runs in $0.129$ seconds with LLR but $10.900$ seconds with KLR, and ${}^\dagger$CLF exhibits a gap from $0.551$ to $7.369$ seconds.}

\mypara{Computational Scaling}
{Three computational patterns emerge from Figure~\ref{fig:computational-costs} and Table~\ref{tab:practitioners_guide}. First, DRT methods with LLR and CIT methods with random forests scale at most polynomially in $n$ with small leading constants in our experiments, and ${}^\dagger$MMD-$\ell$ (LLR) consistently attains sub-second runtimes up to $n=2{,}000$ on both datasets. Second, KLR introduce a super-linear cost. Across the four DRT tests, KLR runtime at $n=2{,}000$ exceeds the corresponding LLR runtime by factors of roughly $13$ to $700$, with ${}^\dagger$MMD-$b$ and ${}^\dagger$MMD-$q$ reaching $40$ seconds and ${}^\dagger$CLF requiring up to $7.369$ seconds. Third, the conditional kernel methods CMMD and CGED, while bypassing $r_X$ estimation, dominate every other method in runtime and exceed $300$ seconds at $n=2{,}000$ on both real datasets. In the low-dimensional setting, ${}^\dagger$MMD-$q$ (LLR) and RCIT occupy the low-cost, high-power region of the trade-off in Figure~\ref{fig:computational-costs}, whereas WGSC and the conditional kernel methods occupy the high-cost end. In the high-dimensional setting where size control is the binding concern, ${}^\dagger$MMD-$\ell$ or ${}^\dagger$CLF with KLR provides size-controlled inference at moderate additional cost relative to the faster LLR variants, while GCM with random forests achieves the highest empirical power at runtimes of up to $23.760$ seconds at $n=2{,}000$. We caution that these observations are specific to the data configurations and estimator choices considered, and that practitioners should weigh the cost-power-validity trade-off in light of their own data structure and computational budget.}

\end{document}